\documentclass{IEEEtran}
\usepackage{fancyhdr}
\usepackage{amssymb}
\usepackage{times}
\usepackage{graphicx}
\usepackage{subfigure}
\usepackage{multirow}
\usepackage{multicol}
\usepackage{amsmath}
\usepackage{dashrule}
\usepackage{algorithm}
\usepackage{algorithmic}
\usepackage[table*]{xcolor}
\usepackage{dcolumn}
\usepackage{multirow}
\usepackage{multicol}
\usepackage{booktabs}
\usepackage{subfigure}

\usepackage{amsthm}

\usepackage{latexsym}

\usepackage{rotating}
\usepackage{textcomp}

\usepackage{colortbl}
\definecolor{light-gray}{gray}{0.75} 

\newcommand{\etal}{\mbox{\emph{et al.}}}

\usepackage{xcolor}
\usepackage[normalem]{ulem} 
\newcommand\hl{\bgroup\markoverwith
  {\textcolor{yellow}{\rule[-.5ex]{2pt}{2.5ex}}}\ULon}

\begin{document}
%
\title{A Many-Objective Evolutionary Algorithm with\\ Angle-Based Selection and Shift-Based \\ Density Estimation}

\author{Zhi-Zhong Liu, Yong Wang, \emph{Member, IEEE}, and Pei-Qiu Huang
\thanks{Z.-Z. Liu and P.-Q. Huang are with the School of Information Science and Engineering, Central South University, Changsha 410083, China (Email: zhizhongliu@csu.edu.cn; pqhuang@csu.edu.cn)}
\thanks{Y. Wang is with the School of Information Science and Engineering, Central South University, Changsha 410083, China, and also with the Centre for Computational Intelligence (CCI), School of Computer Science and Informatics, De Montfort University, Leicester LE1 9BH, UK. (Email: ywang@csu.edu.cn)}
}
\maketitle

\begin{abstract}

Evolutionary many-objective optimization has been gaining increasing attention from the evolutionary computation research community. Much effort has been devoted to addressing this issue by improving the scalability of multiobjective evolutionary algorithms, such as Pareto-based, decomposition-based, and indicator-based approaches. Different from current work, we propose a novel algorithm in this paper called AnD, which consists of an angle-based selection strategy and a shift-based density estimation strategy. These two strategies are employed in the environmental selection to delete the poor individuals one by one. Specifically, the former is devised to find a pair of individuals with the minimum vector angle, which means that these two individuals share the most similar search direction. The latter, which takes both the diversity and convergence into account, is adopted to compare these two individuals and to delete the worse one. AnD has a simple structure, few parameters, and no complicated operators. The performance of AnD is compared with that of seven state-of-the-art many-objective evolutionary algorithms on a variety of benchmark test problems with up to $15$ objectives. The experimental results suggest that AnD can achieve highly competitive performance. In addition, we also verify that AnD can be readily extended to solve constrained many-objective optimization problems.

\end{abstract}

\begin{IEEEkeywords}
Evolutionary algorithms,
many-objective optimization,
angle-based selection,
shift-based density estimation
\end{IEEEkeywords}

%
\IEEEpeerreviewmaketitle

\section{Introduction}\label{sec:Intro}
\IEEEPARstart{M}{ultiobjective} optimization problems (MOPs) refer to the optimization problems with more than one conflicting objective. Usually, an MOP can be expressed as:
\begin{equation}\label{eqn:mops}
\begin{aligned}
minimize & \ \textbf{F}(\textbf{x})=(f_{1}(\textbf{x}),f_{2}(\textbf{x}),...,f_{m}(\textbf{x})) \\
subject \ to & \ \textbf{x}\ \in \ \Omega
\end{aligned}
\end{equation}
where $\textbf{x} = (x_{1}, x_{2},..., x_{n})$ is the decision vector, $\textbf{F}(\textbf{x})$ is the objective vector, $m$ is the number of objectives, and ${\Omega}$ is the decision space. The ultimate goal of multiobjective optimization is to obtain a set of well-distributed and well-converged nondominated solutions to approximate the Pareto front (PF). To achieve this goal, numerous multiobjective evolutionary algorithms (MOEAs) have been proposed over the last few decades. According to their selection mechanisms, MOEAs can be roughly classified into three categories: Pareto-based methods, decomposition-based methods, and indicator-based methods~\cite{zhou2011multiobjective}. MOEAs have shown great potential to solve MOPs with two or three objectives. However, for MOPs with more than three objectives, often known as many-objective optimization problems (MaOPs), they encounter substantial difficulties~\cite{ishibuchi2011behavior,ishibuchi2015behavior}.

For Pareto-based methods, such as NSGA-II~\cite{deb2002fast} and SPEA2~\cite{zitzler2001spea2}, the selection criteria (i.e., the Pareto-based selection and the diversity-based selection) may lose their effectiveness to push the population toward the PF. It is because with the increase of the number of objectives, the proportion of nondominated solutions will increase drastically. As a result, the Pareto-based (primary) selection fails to distinguish the individuals in the population. Under this condition, the diversity-based (secondary) selection will play a major role in the selection process. The secondary selection may make the population well-distributed over the objective space; however, the population tends to be far away from the desired PF due to the neglect of convergence performance. With respect to decomposition-based~\cite{zhang2007moea,zhang2008multiobjective} and indicator-based~\cite{bader2011hype} methods, they do not suffer from the selection pressure issues since they do not rely on Pareto dominance to evolve the population. However, they face their own challenges. Regarding decomposition-based methods, it is not a trivial task to assign the weight vectors or reference points in the high-dimensional objective space. In addition, indicator-based methods always result in high computational time complexity~\cite{Wagner2013A}.

To enhance the scalability of MOEAs for MaOPs, a considerable number of attempts~\cite{li2017multi,bhattacharjee2017bridging} have been made to improve the performance of Pareto-based, decomposition-based, and indicator-based methods, which are briefly introduced below.
\begin{itemize}
  \item \emph{Pareto-based Methods}: Recognizing the drawback of the Pareto-dominance relation for MaOPS, this kind of method intends to modify/relax the definition of Pareto dominance. Along this line, several rules have been proposed such as $\epsilon$-dominance~\cite{laumanns2002combining}, L-dominance~\cite{zou2008new}, fuzzy dominance~\cite{wang2007fuzzy}, and grid dominance~\cite{yang2013grid}. Addtionally, another avenue is to develop customized diversity mechanisms, with the purpose of alleviating the loss of selection pressure. In~\cite{adra2011diversity}, a diversity management mechanism is introduced, which can determine whether or not to activate diversity promotion based on the distribution of population. In~\cite{li2014shift}, a shift-based density estimation strategy is proposed, which shifts the poorly converged individuals into crowded regions and assigns them high density values. As a result, these individuals are very likely to be removed from the population. Inspired by the idea that the knee points are naturally most preferred among nondominated solutions, a knee point-driven evolutionary algorithm is proposed~\cite{zhang2015knee}, in which knee points are explicitly used to enhance the diversity.
  \item {\emph{Decomposition-based Methods}~\cite{trivedi2017survey}}: This kind of method contains two different types. The first type decomposes an MaOP into a series of single-objective optimization problems. MOEA/D~\cite{zhang2007moea} is the most famous one. In MOEA/D, a set of weight vectors are predefined to specify multiple search directions toward the PF. Since the search directions spread out widely, it is expected that the obtained solutions cover the PF well. MOEA/D is original designed for solving MOPs. Recent advances have successfully adapted MOEA/D to solve MaOPs, such as adaptively allocating search effort in MOEA/D-AM2M~\cite{Liu2017Adaptively}, exploiting the perpendicular distance from the solution to the weight vector in MOEA/D-DU~\cite{Yuan2016Balancing}, and using Pareto adaptive scalarizing methods in MOEA/D-PaS~\cite{Wang2016Decomposition}. The second type divides an MaOP into a group of sub-MaOPs~\cite{liu2014decomposition}. One representative is NSGA-III~\cite{deb2014evolutionary}, which makes use of a set of predefined well-distributed reference points to manage nondominated solutions. That is, the nondominated solutions close to the reference points are prioritized. For these two types, to achieve better performance, a crucial issue is how to assign the appropriate weight vectors or reference points~\cite{ishibuchi2017performance}. To this end, an automatic weight vector generation system is devised in~\cite{hughes2007msops}, and a two-layered generation strategy for reference points is proposed in~\cite{deb2014evolutionary}.
  \item \emph{Indicator-based Methods}: In this kind of method, the indicator values are used to guide the search process. Among all the indicators, the hypervolume indicator~\cite{zitzler1998multiobjective} is the most commonly used, which is originally an quality indicator to compare different MOEAs, while subsequently integrated into the evolutionary process. The hypervolume indicator has an attractive property, i.e., it is strictly monotonic with regard to Pareto dominance~\cite{bader2011hype}. Note, however, that the burden for the calculation of hypervolume is very high, and will increase exponentially as the number of objectives increases. To overcome this shortcoming, Monte Carlo simulation is employed in~\cite{bader2011hype} to approximate the exact hypervolume values, with the aim of striking a trade-off between accuracy and computational time. Besides the hypervolume indicator, there are also some cheaper indicators, such as $I_{(\epsilon)^{+}}$ indicator in IBEA~\cite{zitzler2004indicator} and $R2$ indicator in $R2$-EMOA~\cite{trautmann2013r2}. Moreover, the collaboration of different cheap indicators seems to be a promising direction for solving MaOPs~\cite{li2016stochastic}.
\end{itemize}

Apart from the above three categories, several preference-based many-objective evolutionary algorithms (MaOEAs) have been proposed recently~\cite{wang2013preference,wang2015ipicea}, which focus on a subset of the PF based on the user's preference. There are also some dimensionality reduction approaches~\cite{singh2011pareto,bandyopadhyay2015algorithm,yuan2017objective}, aiming to deal with MaOPs with redundant objectives. Addtionally, researchers have tried to take advantage of the merits offered by different categories. Two representatives are MOEA/DD and Two\_Arch2. MOEA/DD~\cite{li2015evolutionary} is based on Pareto dominance and decomposition, and Two\_Arch2~\cite{wang2015two_arch2} is based on Pareto dominance and indicator. Both MOEA/DD and Two-Arch2 show excellent performance on MaOPs. For more information about MaOEAs, interested readers are referred to two survey papers~\cite{ishibuchi2008evolutionary,li2015many}.

Unlike current work, we propose an alternative MaOEA in this paper, called AnD. In evolutionary many-objective optimization, the task of environmental selection is to choose some promising individuals from the union population, which is composed of the parent and offspring populations, for the next generation. AnD tackles this task by two strategies: angle-based selection and shift-based density estimation. First of all, the angle-based selection finds out a pair of individuals with the minimum vector angle. Intuitively, it is necessary to delete one of these two individuals since they search in the most similar direction and it will definitely waste a lot of computational resource if they coexist. In order to make the deletion wiser, we need to take both convergence and diversity of them into account since achieving balance between convergence and diversity is the most important concern in many-objective optimization. Fortunately, the shift-based density estimation has the capability to cover both the distribution and convergence information of individuals~\cite{li2014shift}. Therefore, it is utilized to compare these two individuals and to delete the worse one. By repeating this process, AnD provides a quite natural way for solving MaOPs\----the individuals with poor diversity and convergence will be eliminated from the union population one by one.

The main contributions of this paper are summarized as follows:
\begin{itemize}
\item We degign AnD as an alternative MaOEA, which has a simple structure, few parameters, and no complicated operators. More importantly, AnD is different from existing methods\----it does not use dominance rules, weights vectors/reference points, and indicators. As a consequence, it has the following advantages for solving MaOPs: exempting from insufficient selection pressure in Pareto-based methods, avoiding assigning weight vectors/reference points in decomposition-based methods, and no need to consume a high computational cost as in indicator-based methods.
\item The vector angle~\cite{cheng2016reference,xiang2017vector} and the shift-based density estimation~\cite{li2014shift,wang2016cooperative} have been extensively investigated in the design of MaOEAs, respectively. However, to the best of our knowledge, it is the first attempt to effectively combine them together for solving MaOPs, by making use of their complementary properties. Moreover, AnD provides a straightforward way to achieve both diversity and convergence\----identifying the two individuals with the minimum vector angle via the angle-based selection and removing the worse one with poor diversity and convergence via the shift-based density estimation in an iterative way.
\item Systematic experiments have been conducted on both DTLZ and WFG test suites to demonstrate the effectiveness of AnD. The performance of AnD is compared  with that of seven state-of-the-art MaOEAs. The experimental results suggest that, overall, AnD can achieve better performance in terms of two widely used performance metrics, i.e., IGD~\cite{coello2007evolutionary} and HV~\cite{zitzler1998multiobjective}.
\item AnD has been further extended to solve constrained MaOPs with promising performance.
\end{itemize}

The rest of this paper is organized as follows. Section~\ref{sec:preworks} introduces the preliminary knowledge. The details of AnD are presented in Section~\ref{sec:AnD}. Subsequently, the experimental setup is described in Section~\ref{sec:experimentalsetup}. The empirical results on both unconstrained and constrained MaOPs are given in Section~\ref{sec:results}. Finally, Section~\ref{conclusion} concludes this paper.

\begin{figure} [!t]
    \begin{center}
      \subfigure{\includegraphics[width=5cm]{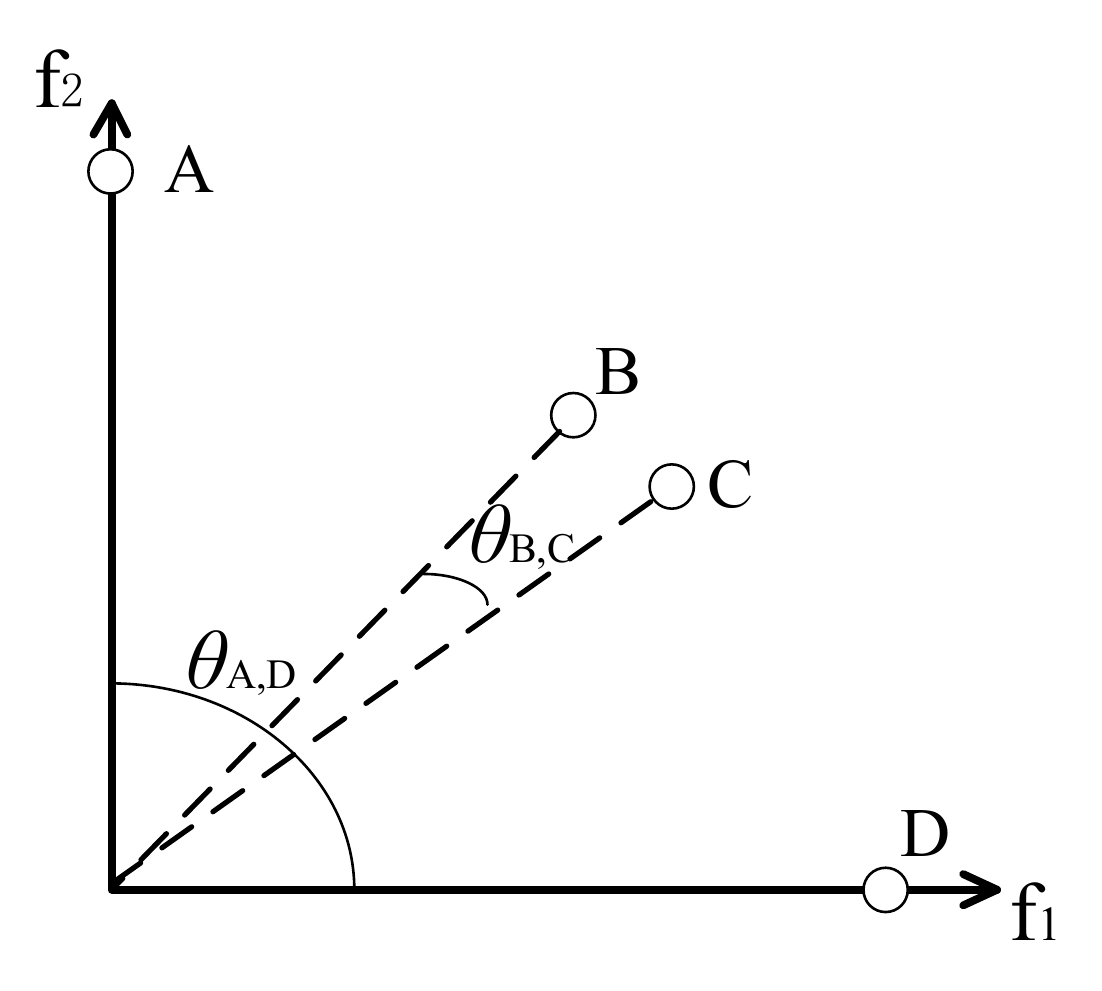}}
       \caption{Illustration of the vector angle in a bi-objective minimization scenario.}\label{fig:angle}
    \end{center}
\end{figure}

\section{Preliminary Knowledge}\label{sec:preworks}

\subsection{Vector Angle} \label{sec:vec}

In this paper, the vector angle denotes the included angle between two individuals in the normalized objective space. The normalized objective vector of an individual is computed as follows. First of all, we find the ideal point $Z^{min}=(z_{1}^{min},z_{2}^{min},...,z_{m}^{min})$ and the nadir point $Z^{max}=(z_{1}^{max},z_{2}^{max},...,z_{m}^{max})$, where $z_{i}^{min}$ and $z_{i}^{max}$ are the minimum and maximum values of the $i$th objective for all individuals, respectively. Afterward, for the $j$th individual $\textbf{x}_{j}$, its objective vector $\textbf{F}(\textbf{x}_{j})$ is normalized as $\textbf{F}^{'}(\textbf{x}_{j})=(f_{1}^{'}(\textbf{x}_{j}),f_{2}^{'}(\textbf{x}_{j}),...,f_{m}^{'}(\textbf{x}_{j}))$ according to
\begin{equation}\label{eqn:mops}
f_{i}^{'}(\textbf{x}_{j})= \frac{f_{i}(\textbf{x}_{j})- z_{i}^{min}}{z_{i}^{max}-z_{i}^{min}}, i= 1,2,...,m
\end{equation}
After the normalization, the vector angle between two individuals $\textbf{x}_{j}$ and $\textbf{x}_{k}$, referred to as $\theta_{\textbf{x}_{j},\textbf{x}_{k}}$, is computed as
\begin{equation}\label{eqn:angle}
\theta_{\textbf{x}_{j},\textbf{x}_{k}} = arccos \left | \frac{\textbf{F}^{'}(\textbf{x}_{j}) \bullet \textbf{F}^{'}(\textbf{x}_{k})}{\parallel\textbf{F}^{'}(\textbf{x}_{j})\parallel\times\parallel\textbf{F}^{'}(\textbf{x}_{k})\parallel} \right |
\end{equation}
where $\textbf{F}^{'}(\textbf{x}_{j}) \bullet \textbf{F}^{'}(\textbf{x}_{k})$ denotes the inner product of $\textbf{F}^{'}(\textbf{x}_{j})$ and $\textbf{F}^{'}(\textbf{x}_{k})$, and $\| \cdot \|$ calculates the norm of a vector. It is clear that $\theta_{\textbf{x}_{j},\textbf{x}_{k}} \in \left [ 0,\frac{\pi}{2} \right ]$.

In principle, the vector angle reflects the similarity of search directions between two individuals. To be specific, if two individuals search in quite different directions, the vector angle between them is large; otherwise, the vector angle is small. Fig.~\ref{fig:angle} gives an example. From Fig.~\ref{fig:angle}, we can observe that: 1) individuals $\textbf{A}$ and $\textbf{D}$ search in quite different directions, then $\theta_{\textbf{A},\textbf{D}}$ is relatively larger, and 2) individuals $\textbf{B}$ and $\textbf{C}$ share the similar search directions, then $\theta_{\textbf{B},\textbf{C}}$ is relatively smaller.

During the recent two years, the vector angle has attracted a high level of interest in evolutionary many-objective optimization. For instance, it has been incorporated into decomposition-based approaches. In~\cite{cheng2016reference}, a reference vector guided evolutionary algorithm (RVEA) for many-objective optimization is proposed. In RVEA, the angle-penalized distance is used to balance convergence and diversity of individuals in the high-dimensional objective space. In~\cite{wang2016localized}, a novel decomposition based MaOEA called MOEA/D-LWS is proposed. In MOEA/D-LWS, for each search direction, the optimal solution is selected only among its neighboring solutions. Note that the neighborhood is defined by a hypercone, whose apex angle is determined automatically in $a$ $priori$. Very recently, a new variant of MOEA/D with sorting-and-selection (MOEA/D-SAS) is presented in~\cite{cai2017decomposition}. In MOEA/D-SAS, the balance between convergence and diversity is achieved by two distinctive components, i.e., decomposition-based-sorting and angle-based-selection. In the latter, the angle information between two individuals in the objective space is used to maintain the diversity.

In addition, the vector angle also has the potential to improve the performance of Pareto-based approaches. In~\cite{xiang2017vector}, a vector angle-based evolutionary algorithm (VaEA) for unconstrained many-objective optimization is developed. VaEA implements the nondominated sorting procedure to obtain different layers, and deals with the last layer through the vector angle. Specifically, the maximum-vector-angle-fist principle is used to guarantee the wideness and uniformity of the solution set. Thereafter, the worse-elimination principle enables worse individuals in terms of convergence to be conditionally replaced with other individuals. Very recently, VaEA is further generalized to solve constrained MaOPs~\cite{Xiang2017An}.

Other kinds of attempts have also been made to solve MaOPs with the usage of vector angle. For example, He and Yen~\cite{he2017many} suggested an MaOEA based on coordinate selection strategy (MaOEA-CSS), in which a new diversity measure based on vector angle is designed in the mating and environmental selection. 

\subsection{Shift-based Density Estimation} \label{sec:sde}

The shift-based density estimation is an advanced density estimation strategy proposed by Li \etal~\cite{li2014shift}. Compared with the traditional density estimation, it shifts the positions of other individuals when estimating the density of an individual (e.g., $\textbf{x}_{j}$) in the population $\mathcal{P}$. This shift process is simple and it is based on the convergence comparison between other individuals and $\textbf{x}_{j}$ on each objective. To be specific, if $\textbf{x}_{k}$ (suppose that $\textbf{x}_{k}$ is another individual in $\mathcal{P}$) outperforms $\textbf{x}_{j}$ on one objective, its objective value on this objective will be shifted to the same position of $\textbf{x}_{j}$ on this objective; otherwise, its objective value keeps unchanged. This process can be described as
\begin{equation}\label{eqn:shift}
f_{i}^{s}(\textbf{x}_{k})  = \left \{
\begin{aligned}
& f_{i}^{'}(\textbf{x}_{j}), \  \text{if} f_{i}^{'}(\textbf{x}_{k}) < f_{i}^{'}(\textbf{x}_{j})\\
& f_{i}^{'}(\textbf{x}_{k}), \  \text{otherwise}
\end{aligned}
\right.
\end{equation}
where $f_{i}^{s}(\textbf{x}_{k})$ is the shifted objective value of $f_{i}^{'}(\textbf{x}_{k})$, and $\textbf{F}^{s}(\textbf{x}_{k})= (f_{1}^{s}(\textbf{x}_{k}),f_{2}^{s}(\textbf{x}_{k}),...,f_{m}^{s}(\textbf{x}_{k})$ is the shifted objective vector of $\textbf{F}^{'}(\textbf{x}_{k})$. Note that before shifting, the objective vector of each individual is normalized via Eq.~\eqref{eqn:mops}.

\begin{figure} [!t]
    \begin{center}
      \subfigure{\includegraphics[width=6cm]{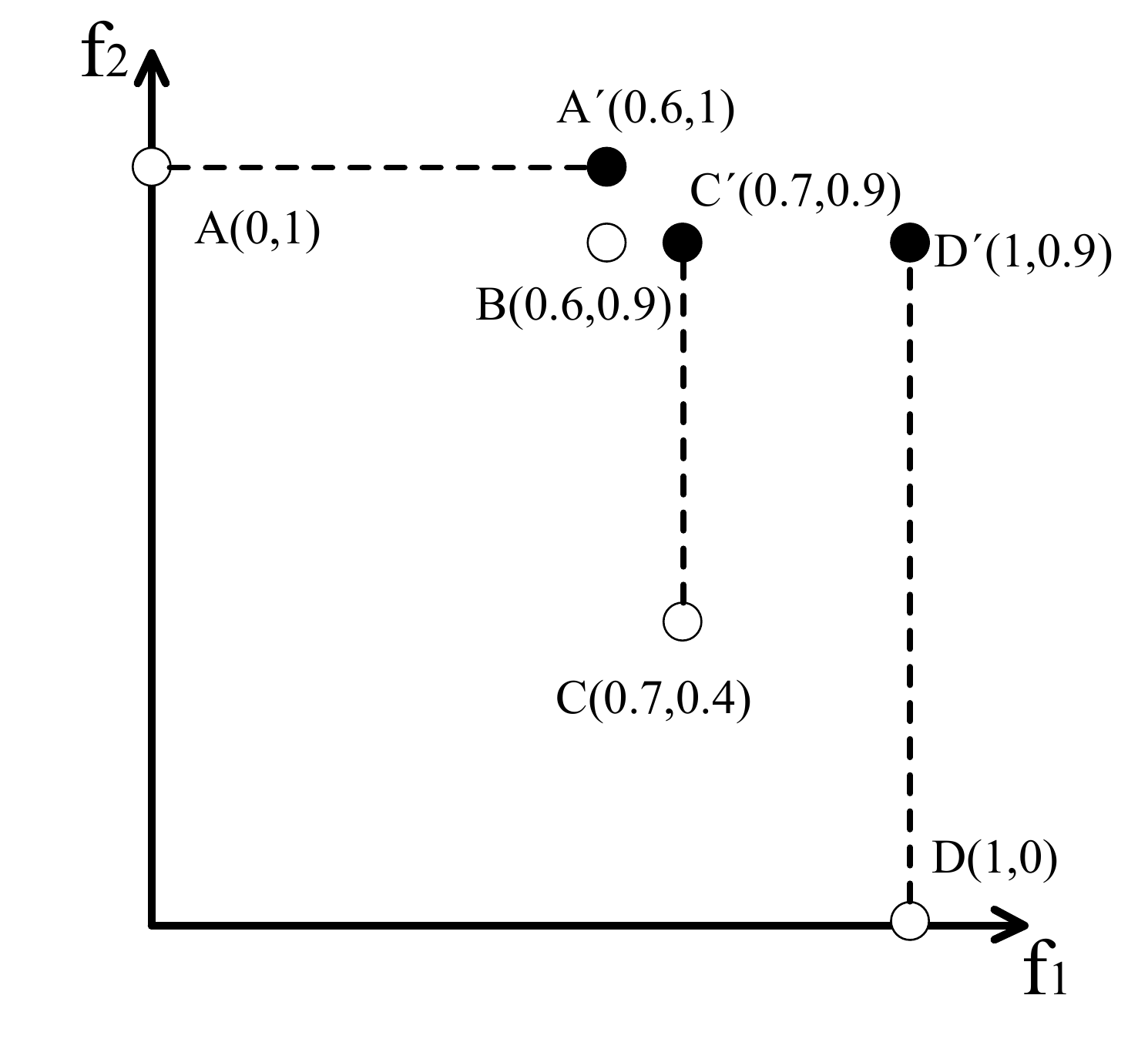}}
       \caption{Illustration of the shift-based density estimation in a two-dimensional normalized objective space. To estimate the density of individual $\textbf{B}$, the other individuals $\textbf{A}$, $\textbf{C}$, and $\textbf{D}$ are shifted to $\textbf{A}'$, $\textbf{C}'$, and $\textbf{D}'$, respectively.}\label{fig:shiftposition}
    \end{center}
\end{figure}

To understand the shift process more clearly, we take the shift-based density estimation of individual $\textbf{B}(0.6,0.9)$ in Fig.~\ref{fig:shiftposition} as an example. First of all, individuals $\textbf{A}(0,1)$, $\textbf{C}(0.7,0.4)$, and $\textbf{D}(1,0)$ in Fig.~\ref{fig:shiftposition} are shifted to individuals $\textbf{A}{'}(0.6,1)$, $\textbf{C}{'}(0.7,0.9)$, and $\textbf{D}{'}(1,0.9)$, respectively, due to the fact that $\textbf{A}_{1}=0<\textbf{B}_{1}=0.6$, $\textbf{C}_{2} = 0.4 < \textbf{B}_{2}=0.9$, and $\textbf{D}_{2}=0<\textbf{B}_{2}=0.9$. Subsequently, it can be observed that the poorly converged individual $\textbf{B}$ is located in a crowded region. Thus, $\textbf{B}$ will be assigned a high density value and is very likely to be removed from the population. It is noteworthy that in order to obtain the density of an individual, the shift process should be combined with a density estimator, such as the crowding distance in NSGA-II~\cite{deb2002fast}, the $k$th nearest neighbor in SPEA2~\cite{zitzler2001spea2}, or the grid crowding degree in PESA-II~\cite{PESA}. Actually, as pointed out in~\cite{li2014shift}, only the individual with both good diversity and good convergence will have a low density value, which means that both diversity and convergence are elaborately considered in the shift-based density estimation.

In this paper, we integrate the shift-based density estimation with the $k$th nearest neighbor to estimate the density of individual $\textbf{x}_{j}$ in $\mathcal{P}$, denoted as $SD(\textbf{x}_{j})$. The implementation is the following:
 \begin{enumerate}
   \item Shift the normalized objective vectors of the other individuals in $\mathcal{P}$ via Eq.~\eqref{eqn:shift};
   \item Calculate the Euclidian distances between the other shifted normalized objective vectors and $\textbf{F}^{'}(\textbf{x}_{j})$ according to:
\begin{equation}\label{eqn:ED}
d(\textbf{x}_{j},\textbf{x}_{k}) = \| \textbf{F}^{s}(\textbf{x}_{k})- \textbf{F}^{'}(\textbf{x}_{j})\|, \textbf{x}_{k} \in \mathcal{P} \cap \textbf{x}_{k} \neq \textbf{x}_{j}
\end{equation}
   \item Find the $k$th  minimum value $\ell (\textbf{x}_{j})$ in the set of $ \left \{ d(\textbf{x}_{j},\textbf{x}_{k}) , \textbf{x}_{k} \in \mathcal{P} \cap \textbf{x}_{k} \neq \textbf{x}_{j} \right \}$, where $k$ is set to $\sqrt{N}$ and $N$ is the size of $\mathcal{P}$;
   \item Compute $SD(\textbf{x}_{j})$ according to Eq.~\eqref{eqn:desity}:
\begin{equation}\label{eqn:desity}
SD(\textbf{x}_{j}) = \frac{1}{\ell(\textbf{x}_{j})+2}
\end{equation}
 \end{enumerate}
Note that the higher the density value, the worse the performance of an individual.

The shift-based density estimation has become an important technique in evolutionary many-objective optimization. From~\cite{li2014shift}, it can significantly enhance the scalability of NSGA-II~\cite{deb2002fast}, SPEA2~\cite{zitzler2001spea2}, and PESA2~\cite{PESA} for solving MaOPs. Moreover, SPEA2 achieves better performance than NSGA-II and PESA2, after these three algorithms are integrated with the shift-based density estimation. Recently, Wang \etal~\cite{wang2016cooperative} presented a cooperative differential evolution with multiple populations (CMODE) for multi- and many-objective optimization. From the experimental results, the combination of CMODE and the shift-based density estimation reaches outstanding performance when solving MaOPs. Very recently, Li \etal~\cite{li2016stochastic} presented a stochastic ranking-based multi-indicator algorithm (SRA). SRA adopts the stochastic ranking technique to balance the search biases of different indicators. Among these indicators, one is designed based on the shift-based density estimation.

\begin{algorithm}[!t]   
   \renewcommand{\algorithmicrequire}{\textbf{Input:}}
  \renewcommand{\algorithmicensure}{\textbf{Output:}}
  \caption{The framework of AnD}\label{alg:AnD}
   \begin{algorithmic}[1]
   \REQUIRE  an MaOP and the population size $N$
   \ENSURE $\mathcal{P}_{t+1}$
   \STATE \emph{Initialization}($\mathcal{P}_{0}$);
   \STATE $t \leftarrow 0$
   \WHILE {$the$ $stopping$ $criterion$ $is$ $not$ $met$}
   \STATE $\mathcal{Q}_{t} \leftarrow \emph{Mating}(\mathcal{P}_{t})$ ;
   \STATE $\mathcal{U}_{t} \leftarrow \mathcal{P}_{t} \bigcup \mathcal{Q}_{t}$;
   \STATE $\mathcal{P}_{t+1} \leftarrow \emph{Enviromental-Selection}(\mathcal{U}_{t})$
   \STATE $t \leftarrow t + 1$;
   \ENDWHILE
\end{algorithmic}
\end{algorithm}

\section{Proposed Approach}\label{sec:AnD}


\subsection{AnD} \label{sec:general}
This paper proposes a new MaOEA with \emph{an}gle-based selection and shift-based \emph{d}ensity estimation, named AnD. The framework of AnD is given in \textbf{Algorithm 1}. First of all, a population $\mathcal{P}_{0}$ with $N$ individuals is randomly initialized in the decision space ${\Omega}$. During the evolution, an offspring population $\mathcal{Q}_{t}$ is generated from $\mathcal{P}_{t}$ through mating. Afterward, an union population $\mathcal{U}_{t}$ is obtained by combining $\mathcal{Q}_{t}$ with $\mathcal{P}_{t}$. Finally, the environmental selection is performed on $\mathcal{U}_{t}$ to produce the next population $\mathcal{P}_{t+1}$. The above procedure repeats until the stopping criterion is met.

From the above introduction, it can be seen that similar to most MaOEAs, AnD involves two main components: mating and environmental selection. The aim of mating is to generate a number of offspring (i.e., $\mathcal{Q}_{t}$) from the parents (i.e., $\mathcal{P}_{t}$) by making use of evolutionary operators, such as selection, crossover, and mutation. AnD does not apply any explicit selection to choose parents from $\mathcal{P}_{t}$ or employ any special crossover and mutation to generate offspring. Instead, the parents are randomly chosen from $\mathcal{P}_{t}$, and the simulated binary crossover (SBX) and the polynomial mutation are utilized to generate $\mathcal{Q}_{t}$. The reasons are twofold: 1) the random selection, SBX, and the polynomial mutation have been widely used in the community of evolutionary many-objective optimization; and 2) we would like to ensure a fair comparison with other algorithms. The unique characteristic of AnD lies in its environmental selection, which is described in the sequel.

\begin{algorithm}[!t]   
   \renewcommand{\algorithmicrequire}{\textbf{Input:}}
  \renewcommand{\algorithmicensure}{\textbf{Output:}}
  \caption{\emph{Environmental-Selection}($\mathcal{U}_{t}$)}\label{alg:selection}
   \begin{algorithmic}[1]
   \REQUIRE $\mathcal{U}_{t}$ which is the union of $\mathcal{P}_{t}$ and $\mathcal{Q}_{t}$
   \ENSURE $\mathcal{P}_{t+1}$
   \STATE Calculate the vector angles between any two individuals in $\mathcal{U}_{t}$ based on Section \ref{sec:vec};
   \WHILE {$|\mathcal{U}_{t}|> N $}
   \STATE Find out two individuals (denoted as $\textbf{u}_{j}$ and $\textbf{u}_{k}$) with the smallest vector angle $\theta_{\textbf{u}_{j},\textbf{u}_{k}}$ in  $ \mathcal{U}_{t} $;
   \STATE Calculate the shift-based density estimation $SD(\textbf{u}_{j})$ and $SD(\textbf{u}_{k})$ according to Section \ref{sec:sde};
   \IF {$SD(\textbf{u}_{j}) < SD(\textbf{u}_{k})$}
   \STATE $\mathcal{U}_{t} \leftarrow \mathcal{U}_{t} / \textbf{u}_{k}$;   \qquad  // delete $\textbf{u}_{k}$ from $\mathcal{U}_{t}$
   \ELSE
   \STATE $\mathcal{U}_{t} \leftarrow \mathcal{U}_{t} / \textbf{u}_{j}$;   \qquad  // delete $\textbf{u}_{j}$ from $\mathcal{U}_{t}$
   \ENDIF
   \ENDWHILE
   \STATE $\mathcal{P}_{t+1} \leftarrow \mathcal{U}_{t}$;
\end{algorithmic}
\end{algorithm}

\subsection{Environmental Selection}\label{enviromental}

The environmental selection aims at choosing $N$ individuals with the most potential from the union population $\mathcal{U}_{t}$ for the next generation. AnD accomplishes this by two strategies: angle-based selection and shift-based density estimation. \textbf{Algorithm 2} describes the environmental selection of AnD. Firstly, the vector angles between any two individuals in $\mathcal{U}_{t}$ are calculated. Thereafter, the angle-based selection is conducted to identify two individuals (denoted as $\textbf{x}_{j}$ and $\textbf{x}_{k}$) with the minimum vector angle $\theta_{\textbf{x}_{j},\textbf{x}_{k}}$ in $\mathcal{U}_{t}$. Subsequently, the shift-based density estimation is employed to compare $\textbf{x}_{j}$ and $\textbf{x}_{k}$, and the one with a higher density value is removed from $\mathcal{U}_{t}$. This process proceeds until the size of $\mathcal{U}_{t}$ is less than or equal to $N$. Next, we explain the importance of the combination of these two strategies in AnD.

As mentioned in Section ~\ref{sec:vec}, the vector angle reflects the similarity between two individuals in their search directions. If two individuals share the minimum vector angle, they absolutely have the most similar search direction. To improve the diversity of search directions, one of them should be discarded. If we repeatedly delete one of two individuals with the most similar search direction in the remaining population, the diversity of search directions will be well maintained in the final population. In fact, the main idea behind the angle-based selection is to approximate the PF from diverse search directions. Decomposition-based approaches, which have shown great success in solving MaOPs, also occupy the similar idea. However, compared with decomposition-based approaches, the angle-based selection only exploits the information provide by vector angle, rather than weight vectors or reference points.

Although the angle-based selection is able to find out a pair of individuals with the most similar search direction, it cannot distinguish them. In many-objective optimization, when comparing two individuals, it has been widely accepted that both the diversity and convergence should be considered. Fortunately, the shift-based density estimation provides an effective way to measure the quality of two individuals, because it focuses on both the diversity and convergence of individuals as introduced in Section~\ref{sec:sde}. As a result, the shift-based density estimation has the capability to judge which individual is worse and should be deleted under this condition.

Overall, the environmental selection of AnD combines the angle-based selection and the shift-based density estimation in a quite natural way: the former identifies the two most similar individuals in terms of search direction, and the latter eliminates the worse one in terms of both diversity and convergence. Therefore, by iteratively implementing both of them, the individuals with poor diversity and convergence will be deleted one by one from $\mathcal{U}_{t}$; thus the population will continuously approach the PF with good diversity.

In principle, AnD can be regarded as a ``diversity-first-and-convergence-second'' MaOEA. It is because the diversity of search directions is considered first in the angle-based selection. Note that the rationality of ``diversity-first-and-convergence-second`` for many-objective optimization has already been verified in~\cite{Jiang2016A}.


\begin{figure*}[!htp]
    \begin{center}
        \subfigure[NSGA-III and SPEA2+SDE]{\label{fig:tnsgaIII}\includegraphics[width=0.55\columnwidth]{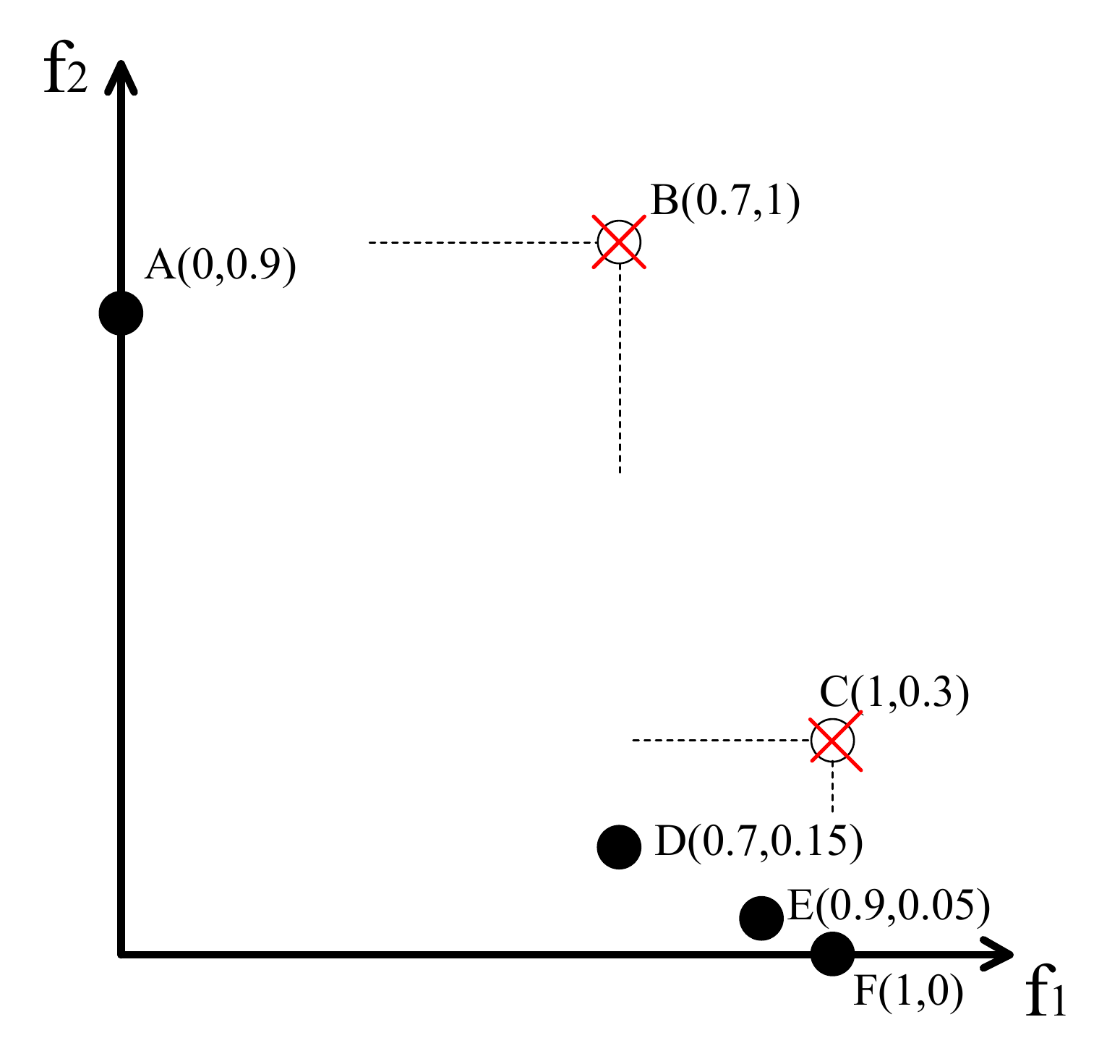}}
        \subfigure[MOEA/D]{\label{fig:tmoead}\includegraphics[width=0.55\columnwidth]{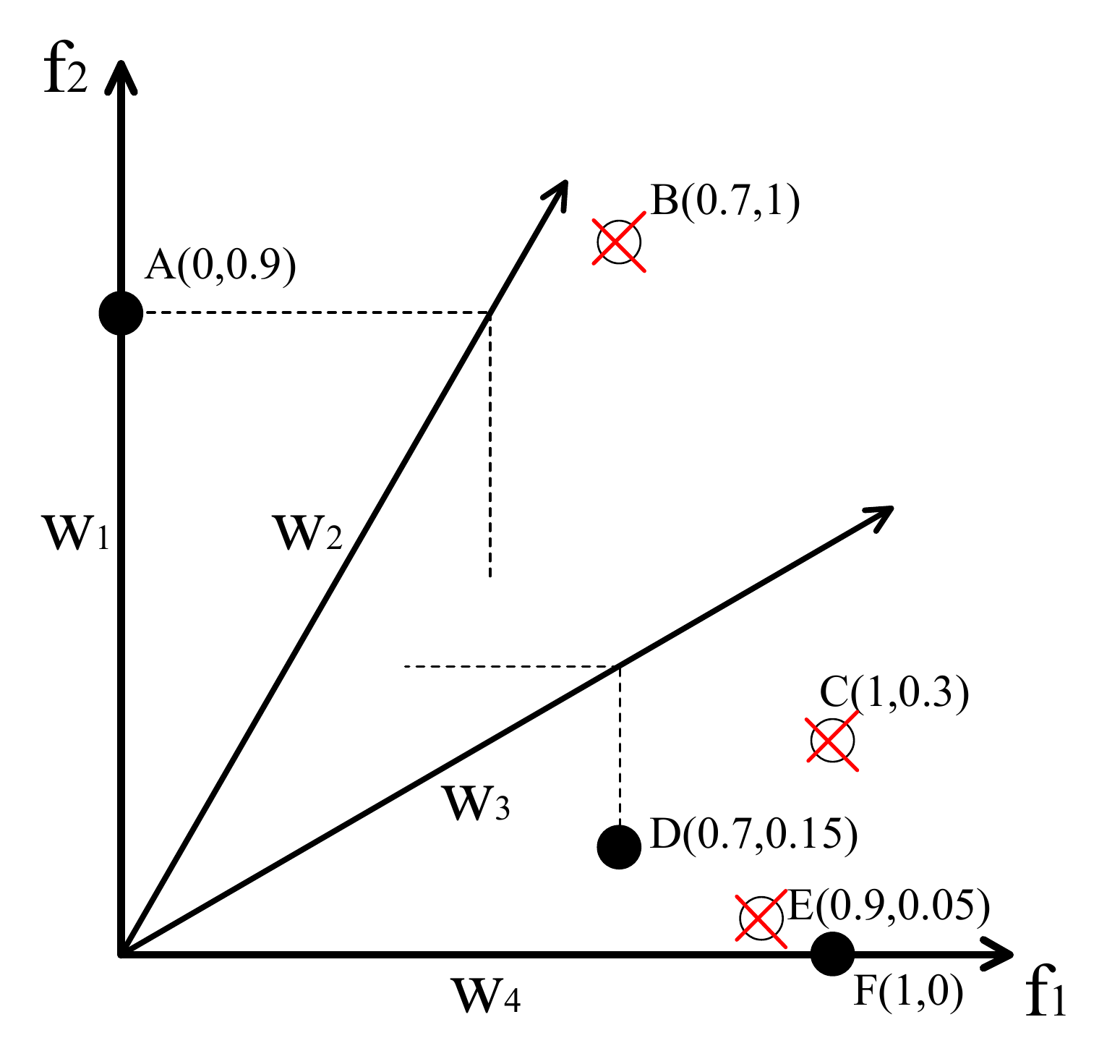}}
        \subfigure[MOEA/DD]{\label{fig:tmoeadd}\includegraphics[width=0.55\columnwidth]{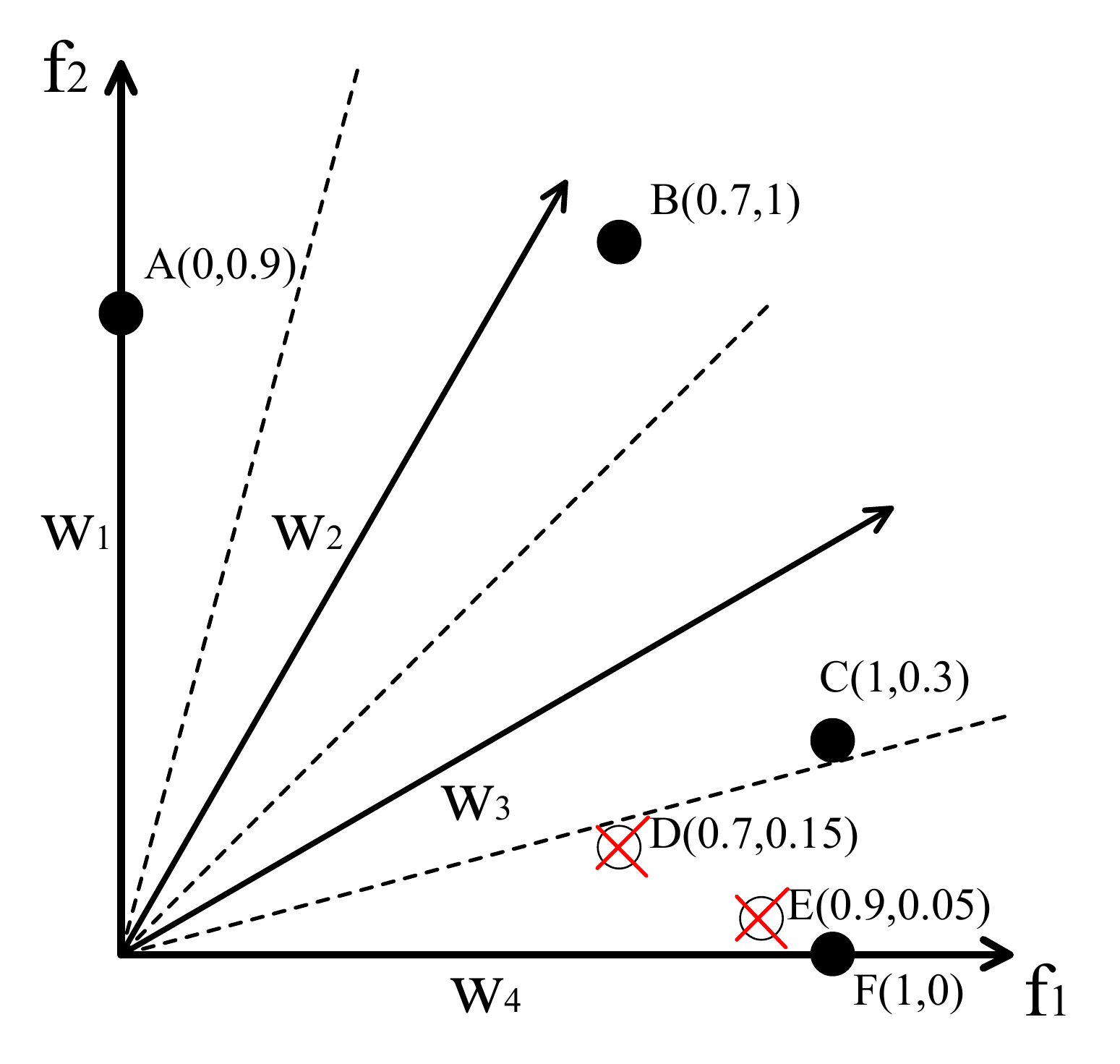}}\\
        \subfigure[HypE]{\label{fig:thype}\includegraphics[width=0.55\columnwidth]{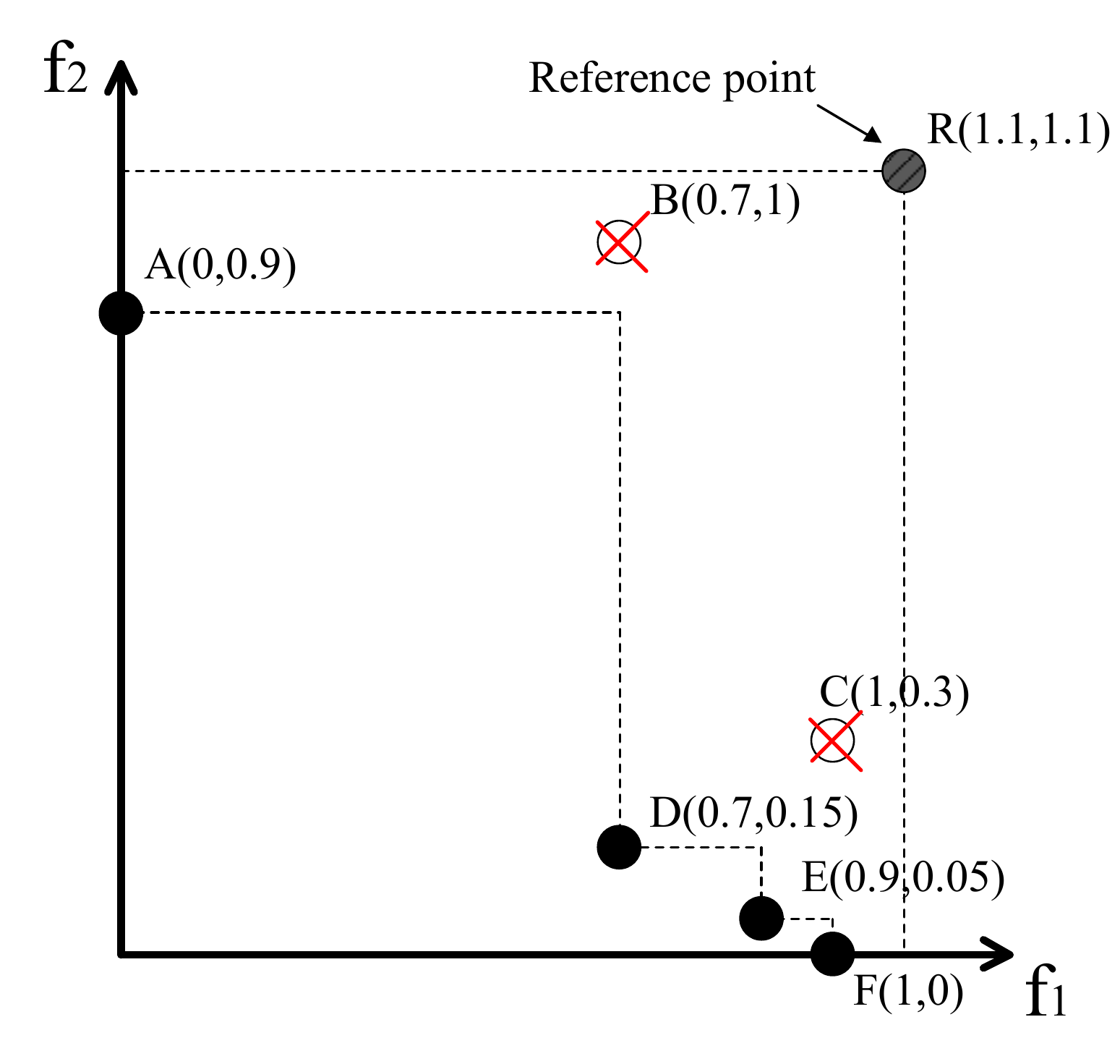}}
        \subfigure[AnD]{\label{fig:tAnD}\includegraphics[width=0.55\columnwidth]{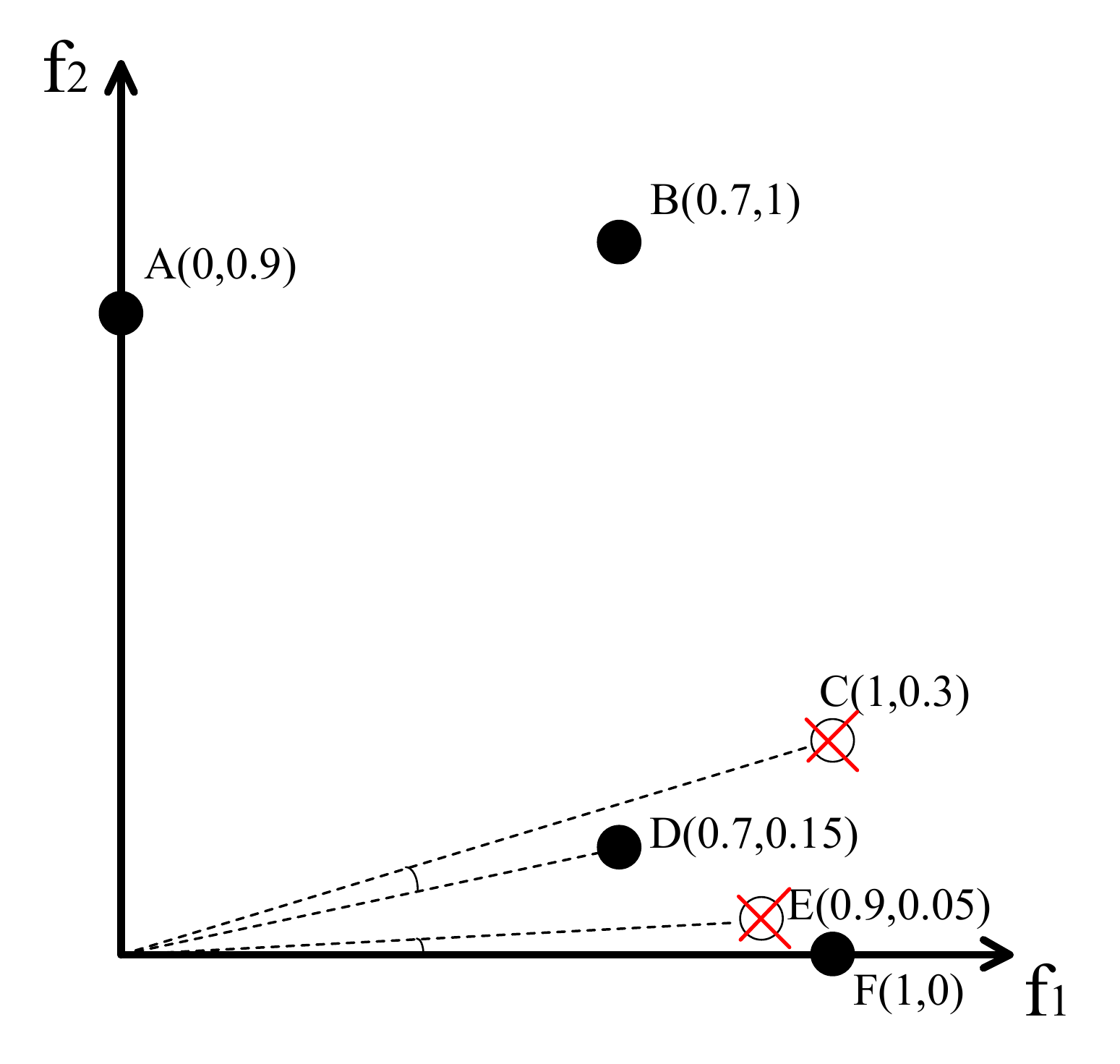}}
       \caption{Illustration of the working principles of six MaOEAs. There are six individuals in the population, i.e., $\textbf{A}$, $\textbf{B}$, $\textbf{C}$, $\textbf{D}$, $\textbf{E}$, and $\textbf{F}$ and the task is to select four promising individuals into the next generation.}\label{fig:difference}
    \end{center}
\end{figure*}

\subsection{Analysis of the Principle}
An example in a two-dimensional objective space is used to illustrate the working principles of six different MaOEAs, i.e., NSGA-III~\cite{deb2014evolutionary}, SPEA2+SDE (a combination of SPEA2 and the shift-based density estimation)~\cite{li2014shift}, MOEA/D~\cite{zhang2007moea}, MOEA/DD~\cite{li2015evolutionary}, HypE~\cite{bader2011hype}, and AnD. Suppose that there are six individuals (i.e., \textbf{A}(0,0.9), \textbf{B}(0.7,1), \textbf{C}(1,0.3), \textbf{D}(0.7,0.15), \textbf{E}(0.9,0.05), and \textbf{F}(0,1)) in the population, and that our task is to select four promising individuals into the next generation. Fig.~\ref{fig:difference} depicts what happens to the six compared methods.

From Fig.~\ref{fig:difference}, we can give the following comments:
\begin{itemize}
\item In both NSGA-III and SPEA2+SDE, $\textbf{A}$, $\textbf{D}$, $\textbf{E}$, and $\textbf{F}$ are selected into the next generation. The reason is that these two algorithms prefer nondominated individuals. From Fig.~\ref{fig:3nsga3}, it can be observed that $\textbf{B}$ is Pareto dominated by $\textbf{A}$ and $\textbf{D}$, and $\textbf{C}$ is Pareto dominated by $\textbf{D}$, $\textbf{E}$, and $\textbf{F}$. Thus, $\textbf{A}$, $\textbf{D}$, $\textbf{E}$, and $\textbf{F}$ are the nondominated individuals, while $\textbf{B}$ and $\textbf{C}$ are the dominated individuals.
\item With respect to MOEA/D, suppose that there are four weight vectors (i.e., $\textbf{w}_{1}$, $\textbf{w}_{2}$, $\textbf{w}_{3}$, and $\textbf{w}_{4}$) as shown in Fig.~\ref{fig:3moead}, and that the Tchebycheff approach is used. According to the principle of MOEA/D with the Tchebycheff approach, each weight vector will be associated with an individual. From Fig.~\ref{fig:3moead}, it is clear that $\textbf{w}_{1}$ and $\textbf{w}_{2}$ are associated with $\textbf{A}$, $\textbf{w}_{3}$ is associated with $\textbf{D}$, and $\textbf{w}_{4}$ is associated with $\textbf{F}$. Therefore, $\textbf{A}$, $\textbf{D}$ and $\textbf{F}$ will survive into the next generation. In particular, $\textbf{A}$ will be duplicated. For the other individuals (i.e., $\textbf{B}$, $\textbf{C}$, and $\textbf{F}$), they will be eliminated.
\item For MOEA/DD, $\textbf{A}$, $\textbf{B}$, $\textbf{C}$, and $\textbf{F}$ will survive into the next generation. It is because the weight vectors in MOEA/DD is used to divide the objective space into a series of subregions. From Fig.~\ref{fig:3moeadd}, we can find that $\textbf{A}$, $\textbf{B}$, and $\textbf{C}$ are in their own isolated subregions and all of them will be selected into the next generation. For $\textbf{D}$, $\textbf{E}$, and $\textbf{F}$, they are in the same subregion and only the boundary individual $\textbf{F}$ will be chosen into the next generation.
\item In terms of HypE, like NSGA-III and SPEA2+SDE, $\textbf{A}$, $\textbf{D}$, $\textbf{E}$, and $\textbf{F}$ are remained and the others are deleted. The reason is the following: $\textbf{B}$ and $\textbf{C}$ have no contribution to the whole population's hypervolume value, while the other individuals (i.e., $\textbf{A}$, $\textbf{D}$, $\textbf{E}$, and $\textbf{F}$) have. As a result, $\textbf{B}$ and $\textbf{C}$ are eliminated from the population.
\item To implement AnD, firstly, it is necessary to calculate the vector angles between any two individuals in the population. Subsequently, we need to identify two individuals with the minimum vector angle and employ the shift-based density estimation to differentiate them. It is easy to find that $\theta_{\textbf{E},\textbf{F}}$ is the minimum vector angle in the population. Then, according to Section~\ref{sec:preworks}, we can obtain the density values of $\textbf{E}$ and $\textbf{F}$ as $SD(\textbf{E})\approx 0.4762$ and $SD(\textbf{F})\approx 0.4651$, respectively. Thus, $\textbf{E}$ will be removed from the population since it has the higher density value. After $\textbf{E}$ has been eliminated, $\theta_{\textbf{C},\textbf{D}}$ becomes the minimum vector angle and the density values of $\textbf{C}$ and $\textbf{D}$ are computed as $SD(\textbf{C}) =0.5$ and $SD(\textbf{D})\approx 0.4348$, accordingly. Thereafter, $\textbf{C}$ will be removed from the population due to its higher density value. In summary, $\textbf{C}$ and $\textbf{E}$ will be eliminated from the population, while $\textbf{A}$, $\textbf{B}$, $\textbf{D}$, and $\textbf{F}$ will be chosen into the next generation.
\end{itemize}

From the above discussions, we can observe that:
\begin{itemize}
\item The principle of AnD is different from that of the other five state-of-the-art MaOEAs. AnD employs two simple strategies, namely the angle-based selection and the shift-based density estimation, to delete the inferior individuals one by one from the population.
\item AnD can obtain more suitable results compared with the five competitors. In NSGA-III, SPEA2+SDE, and HypE, \textbf{A}, \textbf{D}, \textbf{E}, and \textbf{F} survive into the next generation. In terms of MOEA/D, \textbf{A}, \textbf{D}, and \textbf{F} are remained. In particular, \textbf{A} is copied. With respect to MOEA/DD, \textbf{A}, \textbf{B}, \textbf{C}, and \textbf{F} are chosen for survival. However, only in AnD, \textbf{A}, \textbf{B}, \textbf{D}, and \textbf{F} are selected for the next generation simultaneously. Note that \textbf{B} plays an important role in maintaining the diversity of search directions since it is in a sparse region. Nevertheless, only in AnD and MOEA/DD, it survives. In contrast to MOEA/DD, AnD keeps \textbf{D} and deletes \textbf{C}. This seems more reasonable since \textbf{C} and \textbf{D} share the similar search direction but \textbf{D} has better convergence performance. Consequently, we can conclude that AnD can strike a better balance between convergence and diversity than the five competitors.
\end{itemize}

\subsection{Computational Time Complexity}

The computational time complexity of AnD is dependent mainly on its environmental selection. In \textbf{Algorithm 2}, since the vector angles should be computed between any two individual in the union population $\mathcal{U}_{t}$ of size 2$N$, the computational time complexity is thus $O(mN^{2})$. In addition, the computational time complexity of sorting these vector angles is $O(N^{2}log_{2}N)$. The implementation of the shift-based density estimation has a time complexity of $O(mN^{2})$. Therefore, the overall computational time complexity of AnD at one generation is $max\{O(mN^{2}),O(N^{2}log_{2}N)\}$.

\subsection{Discussion}

AnD is an alternative MaOEA, since it abandons the usage of dominance rules, weight vectors or reference points, and indicators. As a result, AnD alleviates the disadvantages of other MaOEAs to some extent when solving MaOPs, such as the loss of selection pressure in Pareto-based approaches, the requirement of specifying weight vectors or reference points in decomposition-based methods, and the high computational time complexity in hypervolume-based approaches. AnD also has some other good properties. For instance, it has a simple structure, few parameters, and no complicated operators. Actually, it is an effective algorithm for solving both unconstrained and constrained MaOPs as demonstrated in Section~\ref{sec:results}.

As introduced in Section~\ref{sec:preworks}, RVEA~\cite{cheng2016reference}, LWS~\cite{wang2016localized}, MOEA/D-SAS~\cite{cai2017decomposition}, VaEA~\cite{xiang2017vector}, MaOEA-CSS~\cite{he2017many}, and MOEA/VAN~\cite{denysiuk2017multiobjective} utilize the information of vector angle, while  SPEA2+SDE~\cite{li2014shift}, CMODE+SDE~\cite{wang2016cooperative}, and SRA~\cite{li2015many} employ the shift-based density estimation. To the best of our knowledge, AnD is the first attempt to combine both the vector angle and the shift-based density estimation together, by taking advantage of their complementary features.

\section{Experimental Setup}\label{sec:experimentalsetup}

\subsection{Benchmark Test Problems}

To evaluate the performance of the proposed AnD, we applied it to solve two well-known benchmark test suites, namely, DTLZ~\cite{deb2005scalable} and WFG~\cite{huband2006review} test suites. DTLZ1-DTLZ4 and WFG1-WFG9 with five, 10, and 15 objectives were chosen for our empirical studies. Following the suggestion in~\cite{deb2005scalable}, the number of decision variables $n$ was set to $n = m + k - 1$ for DTLZ test suite, where $m$ denotes the number of objectives, $k = 5$ for DTLZ1, and $k = 10$ for DTLZ2-DTLZ4. As recommended in~\cite{huband2006review}, $n$ was set to $n = k +l$ for WFG test suite, where the position-related variable $k = 2\times(m - 1)$, and the distance related variable $l = 20$.

As pointed out in~\cite{deb2005scalable} and~\cite{huband2006review}, the PFs of DTLZ and WFG test suites have various characteristics (i.e., linear, convex, concave, mixed, and multi-modal), which pose a grant challenge for an MaOEA to find a well-converged and well-distributed solution set.

\subsection{Performance Metrics}

Two widely used performance metrics, i.e., the inverted generational distance (IGD)~\cite{coello2007evolutionary} and hypervolume(HV)~\cite{zitzler1998multiobjective}, were employed to compare AnD with other MaOEAs.

\begin{itemize}
\item \emph{IGD}: Suppose that $\mathcal{P}$ is an approximation set and $\mathcal{P}^{*}$ is a set of nondominated solutions uniformly distributed on the true PF. The IGD metric is calculated as:
\begin{equation}
IGD(\mathcal{P})=\frac{1}{|\mathcal{P}^{*}|}\sum_{z^{*} \in \mathcal{P}^{*}}distance(z^{*},\mathcal{P})
\end{equation}
where $distance(z^{*},\mathcal{P})$ is the minimum Euclidean distance between $z^{*}$ and all members in $\mathcal{P}$, and $|\mathcal{P}^{*}|$ is the cardinality of $\mathcal{P}^{*}$. The IGD metric has some advantages such as computational efficiency, good flexibility, and generality. It is believed that the smaller the IGD value, the better the performance of an MaOEA.

\item \emph{HV}: HV measures the volume enclosed by $\mathcal{P}$ and a specified reference point in the objective space~\cite{emmerich2005emo}. It assesses both convergence and diversity of $\mathcal{P}$, and is the only indicator which is Pareto-compliant~\cite{bader2011hype}. For an MaOEA, a larger HV value is desirable. In our experiments, firstly, the objective vectors of $\mathcal{P}$ are normalized. Thereafter, the HV value is calculated by using the reference point which is set to 1.1 times of the upper bounds of the true PF. To approximate the exact HV value, usually the Monte Carlo sampling~\cite{bader2011hype} is adopted.
\end{itemize}

\begin{table}[!t]
\caption{Population Size of Three Algorithms}
\label{TablePopulationSize}
\centering
\begin{tabular}{cccc}
\toprule
$m$ &No. of Vectors & NSGA-III & MOEA/D and MOEA/DD\\
\midrule
5 & 210 & 212 & 210\\
10& 275 & 276 & 275\\
15& 135 & 136 & 135\\
\bottomrule
\end{tabular}
\end{table}

\subsection{Algorithms for Comparison}\label{sec:peera}

The following seven state-of-the-art MaOEAs are under our consideration for performance comparison.
\begin{itemize}
\item \emph{RVEA~\cite{cheng2016reference}:} RVEA is a reference vector guided evolutionary algorithm for many-objective optimization. In RVEA, the angle between the reference vector and the objective vector is utilized to compute the angle-penalized distance.
\item \emph{SPEA2+SDE~\cite{li2014shift}:} SPEA2+SDE incorporates the shift-based density estimation into SPEA2 for MaOPs. The effectiveness of SPEA2+SDE has been verified in numerous studies.
\item \emph{MOEA/D~\cite{zhang2007moea}:} Herein, MOEA/D with the penalty-based boundary intersection (PBI) function is used in our experiments, which has been found to be very effective for solving MaOPs.
\item \emph{NSGA-III~\cite{deb2014evolutionary}:} NSGA-III is a reference-point-based MaOEA following NSGA-II framework.
\item \emph{MOMBI-II~\cite{hernandez2015improved}:} MOMBI-II is a recently proposed indicator-based MaOEA which adopts $R2$ indicator as an the selection criterion.
\item \emph{MOEA/DD~\cite{li2015evolutionary}:} MOEA/DD is a popular MaOEA which is based on both Pareto dominance and decomposition.
\item \emph{Two\_Arch2~\cite{wang2015two_arch2}:} Two\_Arch2 is a well-known MaOEA which assigns different selection principles (indicator-based and Pareto-based) to the two archives for convergence and diversity, respectively.
\end{itemize}
\begin{table}[!t]
\caption{Performance Comparison of AnD, AnD-WoA, and AnD-WoD in Terms of the Average IGD Value on DTLZ and WFG Test Suites. The Best Average IGD Value among All the Algorithms on Each Test Problem is Highlighted in Gray.}
\label{table:IGDvariants}
\centering
\begin{scriptsize}
\begin{tabular}{ccccc}
\toprule
Problem&$m$  & AnD-WoA     & AnD-WoD   & AnD\\
\midrule
DTLZ1 &          5 & $7.1041e-2$   & $7.3325e-2$   & \cellcolor{gray}$6.0100e-2$  \\
DTLZ1 &         10 &\cellcolor{gray} $1.2443e-1$   & $1.6631e-1$   & $1.2504e-1 $ \\
DTLZ1 &         15 & \cellcolor{gray}$1.7912e-1$   & $2.3414e-1$   & $1.9064e-1 $ \\
\multicolumn{5}{ c }{\hdashrule[0.5ex]{7.5cm}{0.5pt}{0.8mm} } \\
DTLZ2 &          5 & $2.4507e-1$   & $1.6893e-1$   & \cellcolor{gray}$1.6826e-1 $  \\
DTLZ2 &         10 & $4.8020e-1$   & $4.5503e-1$   &\cellcolor{gray} $3.7456e-1 $ \\
DTLZ2 &         15 & $7.4842e-1$   & $7.3188e-1$   & \cellcolor{gray}$5.4876e-1 $ \\
\multicolumn{5}{ c }{\hdashrule[0.5ex]{7.5cm}{0.5pt}{0.8mm} } \\
DTLZ3 &          5 & $2.5380e-1$   & $1.9300e-1$   &\cellcolor{gray} $1.8791e-1 $ \\
DTLZ3 &         10 & \cellcolor{gray}$4.8583e-1$   & $4.9184e-1$   & $1.1336e+0 $ \\
DTLZ3 &         15 & \cellcolor{gray}$7.6376e-1$   & $9.1091e-1$   & $1.8929e+0 $ \\
\multicolumn{5}{ c }{\hdashrule[0.5ex]{7.5cm}{0.5pt}{0.8mm} } \\
DTLZ4 &          5 & $2.4661e-1$   & \cellcolor{gray}$1.6692e-1$   & $1.6868e-1 $ \\
DTLZ4 &         10 & $4.5454e-1$   & $4.0055e-1$   & \cellcolor{gray}$3.7863e-1 $ \\
DTLZ4 &         15 & $6.8081e-1$   & $5.9132e-1$   & \cellcolor{gray}$5.5382e-1 $ \\
\multicolumn{5}{ c }{\hdashrule[0.5ex]{7.5cm}{0.5pt}{0.8mm} } \\
WFG1 &          5 & $9.3495e-1 $   & $1.0391e+0$   & \cellcolor{gray}$8.2485e-1$  \\
WFG1 &         10 & $1.9120e+0 $   & $2.3222e+0$   & \cellcolor{gray}$1.8344e+0$  \\
WFG1 &         15 & $4.2708e+0 $   & $4.4120e+0$   & \cellcolor{gray}$2.5826e+0$  \\
\multicolumn{5}{ c }{\hdashrule[0.5ex]{7.5cm}{0.5pt}{0.8mm} } \\
WFG2 &          5 & $1.2924e+0 $   & $1.2482e+0$   & \cellcolor{gray}$7.4199e-1$  \\
WFG2 &         10 & $4.8891e+0$    & $4.6090e+0$   & \cellcolor{gray}$3.7346e+0$   \\
WFG2 &         15 & $1.4048e+1 $   & $1.4005e+1$   & \cellcolor{gray}$1.2394e+1$  \\
\multicolumn{5}{ c }{\hdashrule[0.5ex]{7.5cm}{0.5pt}{0.8mm} } \\
WFG3 &          5 & $1.0499e+0$    & $8.2442e-1$   & \cellcolor{gray}$5.0305e-1$  \\
WFG3 &         10 & \cellcolor{gray}$1.5811e+0$    & $2.1080e+0$   & $1.7025e+0$   \\
WFG3 &         15 & $8.6690e+0 $   &\cellcolor{gray} $1.2946e+0$   & $2.6156e+0$  \\
\multicolumn{5}{ c }{\hdashrule[0.5ex]{7.5cm}{0.5pt}{0.8mm} } \\
WFG4 &          5 &$ 1.2669e+0 $   & \cellcolor{gray}$9.4798e-1$   & $9.5061e-1$   \\
WFG4 &         10 &$ 4.6742e+0 $   & $4.3908e+0$   & \cellcolor{gray}$3.6441e+0 $ \\
WFG4 &         15 &$ 1.0007e+1 $   & $1.0596e+1$   & \cellcolor{gray}$7.6264e+0$  \\
\multicolumn{5}{ c }{\hdashrule[0.5ex]{7.5cm}{0.5pt}{0.8mm} } \\
WFG5 &          5 & $1.2722e+0 $   & $9.5546e-1$   & \cellcolor{gray}$9.3925e-1$  \\
WFG5 &         10 & $4.7402e+0 $   & $4.2061e+0$   & \cellcolor{gray}$3.5788e+0 $ \\
WFG5 &         15 & $1.0252e+1 $   & $9.4419e+0$   &\cellcolor{gray} $7.5925e+0 $ \\
\multicolumn{5}{ c }{\hdashrule[0.5ex]{7.5cm}{0.5pt}{0.8mm} } \\
WFG6 &          5 & $1.3676e+0 $   &\cellcolor{gray} $9.3884e-1$    & $9.5995e-1$   \\
WFG6 &         10 & $4.7496e+0 $   & $4.0846e+0$    & \cellcolor{gray}$3.5574e+0 $ \\
WFG6 &         15 & $1.1015e+1 $   & $1.0492e+1$    & \cellcolor{gray}$7.5193e+0 $ \\
\multicolumn{5}{ c }{\hdashrule[0.5ex]{7.5cm}{0.5pt}{0.8mm} } \\
WFG7 &          5 & $1.3386e+0 $   & $9.5717e-1$   & \cellcolor{gray}$9.5631e-1 $ \\
WFG7 &         10 & $4.6176e+0$    & $4.1528e+0$    & \cellcolor{gray}$3.4909e+0 $ \\
WFG7 &         15 & $1.0715e+1 $   & $9.7741e+0$   & \cellcolor{gray}$7.5817e+0 $  \\
\multicolumn{5}{ c }{\hdashrule[0.5ex]{7.5cm}{0.5pt}{0.8mm} } \\
WFG8 &          5 & $1.3795e+0 $   & $1.0943e+0$    &\cellcolor{gray} $1.0138e+0$  \\
WFG8 &         10 & $5.4254e+0 $   & $5.0135e+0$   & \cellcolor{gray}$3.8497e+0 $ \\
WFG8 &         15 & $1.1483e+1 $   & $1.2461e+1$    & \cellcolor{gray}$8.8015e+0 $ \\
\multicolumn{5}{ c }{\hdashrule[0.5ex]{7.5cm}{0.5pt}{0.8mm} } \\
WFG9 &          5 & $1.2789e+0$    & $1.0351e+0$    & \cellcolor{gray}$9.4961e-1 $ \\
WFG9 &         10 & $4.9261e+0$    & $4.6810e+0$    & \cellcolor{gray}$3.9489e+0 $ \\
WFG9 &         15 & $1.0603e+1 $   & $9.8005e+0$    & \cellcolor{gray}$8.0252e+0 $ \\

\bottomrule
\end{tabular}
\end{scriptsize}
\end{table}

\begin{table}[!t]
\caption{Performance Comparison of of AnD, AnD-WoA, and AnD-WoD in Terms of the Average HV Value on DTLZ and WFG Test Suites. The Best Average HV Value among All the Algorithms on Each Test Problem is Highlighted in Gray.}
\label{table:HVvariants}
\centering
\begin{scriptsize}
\begin{tabular}{ccccc}
\toprule
Problem&$m$  & AnD-WoA     & AnD-WoD   & AnD\\
\midrule
DTLZ1 &          5 & $9.6282e-1$   & $9.1737e-1$   &\cellcolor{gray} $9.7330e-1$  \\
DTLZ1 &         10 & $9.9827e-1$   & $9.5804e-1 $  & \cellcolor{gray}$9.9959e-1$ \\
DTLZ1 &         15 & $9.9371e-1$   & $8.5799e-1$   & \cellcolor{gray}$9.9759e-1$  \\
\multicolumn{5}{ c }{\hdashrule[0.5ex]{7.5cm}{0.5pt}{0.8mm} } \\
DTLZ2 &          5 & $7.3680e-1$   & $7.8779e-1 $  & \cellcolor{gray}$8.0057e-1$  \\
DTLZ2 &         10 & $8.8263e-1$   & $8.9729e-1 $  & \cellcolor{gray}$9.6438e-1$  \\
DTLZ2 &         15 & $7.5375e-1$   & $8.3318e-1$  &\cellcolor{gray}$ 9.8283e-1$ \\
\multicolumn{5}{ c }{\hdashrule[0.5ex]{7.5cm}{0.5pt}{0.8mm} } \\
DTLZ3 &          5 & $7.1446e-1$   & $7.4188e-1 $  & \cellcolor{gray}$7.7597e-1$  \\
DTLZ3 &         10 &\cellcolor{gray} $8.7058e-1$   & $8.1980e-1 $ & $5.1899e-1$  \\
DTLZ3 &         15 & \cellcolor{gray}$6.9805e-1$  & $5.4163e-1$  & $4.5641e-1 $ \\
\multicolumn{5}{ c }{\hdashrule[0.5ex]{7.5cm}{0.5pt}{0.8mm} } \\
DTLZ4 &          5 & $7.4610e-1$   & $7.9689e-1 $  & \cellcolor{gray}$8.0242e-1$  \\
DTLZ4 &         10 & $9.2569e-1$   & $9.4718e-1 $ & \cellcolor{gray}$9.6371e-1$  \\
DTLZ4 &         15 & $8.9579e-1$   & $9.5041e-1 $  &\cellcolor{gray} $9.8315e-1$  \\
\multicolumn{5}{ c }{\hdashrule[0.5ex]{7.5cm}{0.5pt}{0.8mm} } \\
WFG1 &          5 &$ 6.2555e-1$   & $5.9549e-1 $  & \cellcolor{gray}$6.7931e-1$  \\
WFG1 &         10 & $4.7609e-1$   & $4.2742e-1 $  &\cellcolor{gray}$ 5.1846e-1$  \\
WFG1 &         15 & $1.9723e-1$   & $1.9179e-1 $  &\cellcolor{gray} $8.1532e-1$  \\
\multicolumn{5}{ c }{\hdashrule[0.5ex]{7.5cm}{0.5pt}{0.8mm} } \\
WFG2 &          5 & $9.3432e-1$   & $9.5884e-1 $   &\cellcolor{gray} $9.8630e-1 $ \\
WFG2 &         10 & $9.7120e-1$   & $9.7291e-1 $   & \cellcolor{gray}$9.8332e-1$  \\
WFG2 &         15 & $9.4314e-1$   & $9.6010e-1$   &\cellcolor{gray} $9.7782e-1$  \\
\multicolumn{5}{ c }{\hdashrule[0.5ex]{7.5cm}{0.5pt}{0.8mm} } \\
WFG3 &          5 & $6.3336e-1$  & $6.1818e-1 $  &\cellcolor{gray} $6.9653e-1$  \\
WFG3 &         10 & $3.8900e-1$   &\cellcolor{gray} $7.1603e-1 $  & $6.2796e-1$  \\
WFG3 &         15 & $6.7204e-1$  & $6.6477e-1 $ & \cellcolor{gray}$7.0778e-1 $  \\
\multicolumn{5}{ c }{\hdashrule[0.5ex]{7.5cm}{0.5pt}{0.8mm} } \\
WFG4 &          5 & $6.7342e-1$  & $7.5084e-1 $  &\cellcolor{gray} $7.5858e-1 $ \\
WFG4 &         10 & $7.9018e-1$  & $7.9607e-1 $  &\cellcolor{gray} $8.6804e-1$  \\
WFG4 &         15 & $6.5414e-1$  & $7.3419e-1 $  &\cellcolor{gray} $9.0693e-1$  \\
\multicolumn{5}{ c }{\hdashrule[0.5ex]{7.5cm}{0.5pt}{0.8mm} } \\
WFG5 &          5 & $6.5241e-1$   & $6.7651e-1 $  &\cellcolor{gray} $7.2568e-1 $ \\
WFG5 &         10 & $7.3629e-1$   & $7.3755e-1 $  & \cellcolor{gray}$8.3440e-1 $ \\
WFG5 &         15 & $6.1065e-1$   & $6.6653e-1 $  & \cellcolor{gray}$8.5625e-1 $ \\
\multicolumn{5}{ c }{\hdashrule[0.5ex]{7.5cm}{0.5pt}{0.8mm} } \\
WFG6 &          5 & $6.2895e-1$   & $7.2332e-1 $  & \cellcolor{gray}$7.3791e-1$  \\
WFG6 &         10 & $7.6931e-1$   & $8.1213e-1 $  &\cellcolor{gray} $8.5307e-1$  \\
WFG6 &         15 & $6.2529e-1$   & $7.1294e-1 $  & \cellcolor{gray}$8.9315e-1$  \\
\multicolumn{5}{ c }{\hdashrule[0.5ex]{7.5cm}{0.5pt}{0.8mm} } \\
WFG7 &          5 & $6.9996e-1$   & $7.5865e-1 $  &\cellcolor{gray}$7.9304e-1 $ \\
WFG7 &         10 & $8.2436e-1$   & $8.5946e-1 $  &\cellcolor{gray} $9.2941e-1$  \\
WFG7 &         15 & $6.5830e-1$   & $7.8871e-1 $  & \cellcolor{gray}$9.7162e-1 $ \\
\multicolumn{5}{ c }{\hdashrule[0.5ex]{7.5cm}{0.5pt}{0.8mm} } \\
WFG8 &          5 & $5.4790e-1$  & $6.0129e-1 $  &\cellcolor{gray}$ 6.6159e-1 $ \\
WFG8 &         10 & $6.5845e-1$   & $6.8902e-1 $  & \cellcolor{gray}$7.4075e-1 $ \\
WFG8 &         15 & $5.4006e-1$   & $7.1689e-1 $  &\cellcolor{gray} $8.9873e-1 $ \\
\multicolumn{5}{ c }{\hdashrule[0.5ex]{7.5cm}{0.5pt}{0.8mm} } \\
WFG9 &          5 & $6.1183e-1 $ & $6.4993e-1  $   &\cellcolor{gray} $6.6130e-1 $ \\
WFG9 &         10 & $7.1796e-1$  & $ 6.7073e-1 $   & \cellcolor{gray}$7.3830e-1 $ \\
WFG9 &         15 & $6.3554e-1$  & $5.9935e-1 $   &\cellcolor{gray} $7.2111e-1 $ \\
\bottomrule
\end{tabular}
\end{scriptsize}
\end{table}

\subsection{Parameter Settings}
\begin{itemize}
  \item \emph{Population Size}: Table~\ref{TablePopulationSize} presents the population size of NSGA-III, MOEA/D, and MOEA/DD. For other algorithms, the population size kept the same with NSGA-III.
  \item \emph{Parameter Settings for Evolutionary Operators}: For all algorithms, the simulated binary crossover (SBX) and polynomial mutation were used to produce offspring. Following the suggestion in~\cite{TianAn}, the crossover probability and the mutation probability were set to 1.0 and $1/D$, respectively, and the distribution indexes of both SBX and the polynomial mutation were set to 20.
  \item \emph{Number of Independent Runs and Termination Condition}: All algorithms were independently run 20 times on each test problem, and terminated when 90,000 function evaluations (FEs) reached~\cite{wang2015two_arch2,li2015many}.
  \item \emph{Parameter Settings for Algorithms}: For RVEA~\cite{cheng2016reference}, $\alpha = 2$ and $f_{r} = 0.1$ for all test problems following the suggestion in~\cite{cheng2016reference}. For MOEA/D~\cite{zhang2007moea}, the neighborhood size was set to 20, the maximum replacement number was set to 2, and the penalty parameter $\theta$ was set to 5. For MOEA/DD~\cite{li2015evolutionary}, the neighborhood size and  $\theta$ kept the same with MOEA/D, and the probability $\delta$ was set to 0.9. According to~\cite{hernandez2015improved}, two parameters in MOMBI-II were set as $\epsilon = 0.001$ and $\alpha = 0.5$, respectively.
\end{itemize}

In this paper, all the experiments were implemented in the platform recently developed by Tian \etal~\cite{Tian2017PlatEMO}.

\begin{table*}[!t]
\caption{Performance Comparison between AnD and Seven State-of-the-Art MaOEAs in Terms of the Average IGD Value on DTLZ Test Suite. The Best and Second Best Average IGD Values among All the Algorithms on Each Test Problem are Highlighted in Gray and Light Gray, Respectively.}
\label{table:IGDstateDTLZ}
\centering
\begin{scriptsize}
\begin{tabular}{cccc cccc cc}
\toprule
Problem& $m$  & RVEA     & SPEA2$+$SDE & MOEA/D & NSGA-III & MOMBI-II & MOEA/DD & Two$\_$Arch2  & AnD\\
\midrule
 DTLZ1 &          5 & \cellcolor{lightgray}$5.2408e-2$   &  \cellcolor{gray}$5.0463e-2$  & $5.2640e-2$  & $5.2550e-2$   & $5.2675e-2$   & $5.2435e-2$   & $5.3565e-2$   & $6.0100e-2 $ \\
DTLZ1 &         10 & $1.4004e-1$   &  \cellcolor{gray}$1.1292e-1$   & \cellcolor{lightgray}$1.1831e-1$  & $1.8856e-1$   & $2.1156e-1$  & $1.2387e-1$   & $1.2356e-1$   & $1.2504e-1$  \\
 DTLZ1 &         15 & $1.8157e-1$   &\cellcolor{lightgray} $1.6530e-1$   &  \cellcolor{gray}$1.6486e-1$   & $2.6475e-1$   & $3.1515e-1$   & $1.6710e-1$  & $1.8887e-1$   & $1.9064e-1$  \\
\multicolumn{10}{ c }{\hdashrule[0.5ex]{17cm}{0.5pt}{0.8mm} } \\
 DTLZ2 &          5 & \cellcolor{lightgray}$1.6124e-1$   & $1.8868e-1$   & $1.6124e-1$   & $1.6125e-1$  & $1.6307e-1$   &  \cellcolor{gray}$1.6124e-1$   & $1.7374e-1$   & $1.6826e-1$  \\
 DTLZ2 &         10 & $4.1870e-1$   &\cellcolor{lightgray} $3.8855e-1$   & $4.2106e-1$   & $4.9906e-1$   & $4.1920e-1$   & $4.2108e-1 $  & $4.5879e-1 $  &  \cellcolor{gray}$3.7456e-1 $ \\
 DTLZ2 &         15 & $5.7768e-1$   &\cellcolor{lightgray} $5.5854e-1$   & $5.7862e-1$   & $6.4743e-1$   & $8.3078e-1$   & $5.7824e-1$   & $6.7923e-1$   &  \cellcolor{gray}$5.4876e-1$  \\
\multicolumn{10}{ c }{\hdashrule[0.5ex]{17cm}{0.5pt}{0.8mm} } \\
DTLZ3 &          5 & $1.7390e-1$   & $1.8818e-1$   & \cellcolor{lightgray}$1.6662e-1$  & $1.7486e-1$   & $1.7057e-1$  & \cellcolor{gray} $1.6305e-1$   & $2.2382e-1$  & $1.8791e-1$  \\
DTLZ3 &         10 & $6.4615e-1$   & \cellcolor{gray} $3.9359e-1$   & $8.3022e-1$  ≈ & $9.5607e+0$   & $4.8416e-1$   & \cellcolor{lightgray}$4.5511e-1$  & $1.8047e+0$   & $1.1336e+0$  \\
DTLZ3 &         15 & $7.3919e-1$   &  \cellcolor{gray}$5.7993e-1$   & $1.1593e+0$  ≈ & $3.1400e+1 $  & $1.1082e+0$  & \cellcolor{lightgray}$5.9058e-1$  & $6.6586e+0$   & $1.8929e+0$  \\
\multicolumn{10}{ c }{\hdashrule[0.5ex]{17cm}{0.5pt}{0.8mm} } \\
DTLZ4 &          5 & \cellcolor{lightgray}$1.6122e-1$   & $1.8854e-1$   & $4.6392e-1$   & $1.6128e-1$  & $1.6474e-1$   &  \cellcolor{gray}$1.6122e-1$   & $1.7605e-1 $  & $1.6868e-1 $ \\
DTLZ4 &         10 & $4.1616e-1$   & \cellcolor{lightgray}$3.9027e-1 $  & $6.2527e-1 $  & $4.4139e-1$   & $4.2156e-1$  & $4.2107e-1$   & $4.5855e-1$   &  \cellcolor{gray}$3.7863e-1 $ \\
DTLZ4 &         15 & $5.8259e-1 $  & \cellcolor{lightgray}$5.5802e-1$   & $7.7231e-1 $  & $6.2785e-1$   & $5.9901e-1$   & $5.8698e-1$   & $6.7621e-1 $ &  \cellcolor{gray}$5.5382e-1$  \\
\bottomrule
\end{tabular}
\end{scriptsize}
\end{table*}

\begin{table*}[!t]
\caption{Performance Comparison between AnD and Seven State-of-the-Art MaOEAs in Terms of the Average HV Value on DTLZ Test Suite. The Best and Second Best Average HV Values among All the Algorithms on Each Test Problem are Highlighted in Gray and Light Gray, Respectively}
\label{table:HVstateDTLZ}
\centering
\begin{scriptsize}
\begin{tabular}{cccc  cccc cc}
\toprule
Problem&$m$  & RVEA     & SPEA2$+$SDE & MOEA/D & NSGA-III & MOMBI-II & MOEA/DD & Two$\_$Arch2  & AnD\\
\midrule
DTLZ1 &    5 & \cellcolor{lightgray}$9.7971e-1$   & $9.6924e-1$   & $9.7942e-1$   & $9.7962e-1$   & $9.7942e-1$   &\cellcolor{gray} $9.7977e-1$   & $9.7582e-1$   & $9.7330e-1$  \\
DTLZ1 &   10 & $9.9868e-1$   & $9.9598e-1$   & $9.9826e-1$   & $9.1305e-1$   &$ 9.6719e-1$   & \cellcolor{gray}$9.9962e-1$   & $9.9507e-1$   & \cellcolor{lightgray}$9.9894e-1$  \\
DTLZ1 &   15 & \cellcolor{gray}$9.9950e-1$   & $9.9122e-1$   & $9.5880e-1$   & $9.0076e-1$   & $8.2655e-1$   & $9.9707e-1$   & $9.8287e-1$   & \cellcolor{lightgray}$9.9759e-1$ \\
\multicolumn{10}{ c }{\hdashrule[0.5ex]{17cm}{0.5pt}{0.8mm} } \\
DTLZ2 &    5 & $8.1234e-1 $  & $8.1125e-1$   & \cellcolor{gray}$8.1254e-1$   & $8.1233e-1$   & $8.1174e-1$   & \cellcolor{lightgray}$8.1247e-1$   & $7.6937e-1$   & $ 8.0057e-1$  \\
DTLZ2 &   10 & $9.7440e-1 $  & $9.7222e-1$   & \cellcolor{lightgray}$9.7529e-1$   & $9.2862e-1$   & $9.7439e-1$   & \cellcolor{gray}$9.7531e-1$   & $7.7695e-1$   & $9.6438e-1$  \\
DTLZ2 &   15 & \cellcolor{lightgray}$9.8964e-1 $  & $9.8403e-1$   & $9.8841e-1$   & $9.3018e-1$   & $8.4042e-1$   &\cellcolor{gray} $9.9015e-1$   & $6.3825e-1$   & $9.8283e-1$  \\
\multicolumn{10}{ c }{\hdashrule[0.5ex]{17cm}{0.5pt}{0.8mm} } \\
DTLZ3 &    5 & $7.7880e-1 $  & \cellcolor{gray}$8.0874e-1$   & $7.8499e-1$   & $7.7721e-1$   & \cellcolor{lightgray}$8.0386e-1$   & $7.9779e-1$   & $7.2701e-1$   & $7.7597e-1$  \\
DTLZ3 &   10 & $7.0475e-1 $  &\cellcolor{gray} $9.6885e-1$   & $4.9755e-1$   & $0.0000e+0$   & \cellcolor{lightgray}$9.3247e-1$   & $9.2789e-1$   & $1.6481e-1$   & $5.1899e-1$  \\
DTLZ3 &   15 & $7.6387e-1 $  & \cellcolor{lightgray}$9.7532e-1$   & $2.3293e-1$   & $0.0000e+0$   & $4.3883e-1$   & \cellcolor{gray}$9.7556e-1$   & $0.0000e+0$   & $4.5641e-1$ \\
\multicolumn{10}{ c }{\hdashrule[0.5ex]{17cm}{0.5pt}{0.8mm} } \\
DTLZ4 &    5 &\cellcolor{lightgray} $8.1251e-1 $  & $8.1225e-1$   & $6.6181e-1$   & $8.1186e-1$   & $8.1128e-1$   &\cellcolor{gray}$ 8.1256e-1$   & $7.5774e-1$   & $8.0242e-1 $ \\
DTLZ4 &   10 & \cellcolor{lightgray}$9.7383e-1 $  & $9.7124e-1$   & $8.7615e-1$   & $9.6258e-1$   & $9.7453e-1$   & \cellcolor{gray}$9.7536e-1$   & $7.8132e-1$   & $9.6371e-1$  \\
DTLZ4 &   15 & \cellcolor{lightgray}$9.8789e-1 $  & $9.8745e-1$   & $8.8487e-1$   & $9.5702e-1$   & $9.8474e-1$   &\cellcolor{gray} $9.8837e-1$   & $6.3605e-1$   & $9.8315e-1 $ \\
\bottomrule
\end{tabular}
\end{scriptsize}
\end{table*}

\section{Results and Discussions}\label{sec:results}

\subsection{Benefit of Two Strategies}

Firstly, we are interested in identifying the benefit of two crucial strategies of AnD: angel-based selection and shift-based density estimation. To this end, two variants of AnD were devised named as AnD-WoA and AnD-WoD, respectively. In AnD-WoA, the angle-based selection was eliminated. Instead, the individuals in the union population $\mathcal{U}_{t}$ were sorted based on their shift-based density values~\cite{li2016stochastic} and $N$ individuals with the highest density values were removed from $\mathcal{U}_{t}$. In AnD-WoD, the shift-based density estimation was abandoned. As an alternative, for two individuals with the minimum vector angle, their Euclidean distances to the ideal point were computed and the one with the larger Euclidean distance was deleted~\cite{denysiuk2017multiobjective,he2017many}. The comparative experiments between AnD and its two variants were carried out on DTLZ and WFG test suites. The IGD and HV values are shown in Tables~\ref{table:IGDvariants} and~\ref{table:HVvariants}, respectively.

From Tables~\ref{table:IGDvariants} and~\ref{table:HVvariants}, it is evident that AnD outperforms its two variants on a vast majority of test problems. In terms of the IGD metric, AnD obtains the best performance on 30 out of 39 test problems, while AnD-WoA and AnD-WoD achieve the best performance on only five and four test problems, respectively. With respect to the HV metric, AnD performs the best on 36 test problems. Nevertheless, AnD-WoA and AnD-WoD perform the best on no more than two test problems. The reason for the above results seems obvious. Compared with AnD, AnD-WoA discards the angle-based selection, therefore it is unable to maintain the diversity of search directions. Regarding AnD-WoD, it replaces the shift-based density estimation with the Euclidean distance. However, the Euclidean distance only considers an individual's convergence property; thus, it is less reasonable than the shift-based density estimation which takes both the diversity and convergence into account.

From the above discussions, we can conclude that the angel-based selection and the shift-based density estimation are two indispensable strategies in AnD.

\begin{figure*} [!t]
    \begin{center}
      \subfigure[RVEA]{\includegraphics[width=4cm]{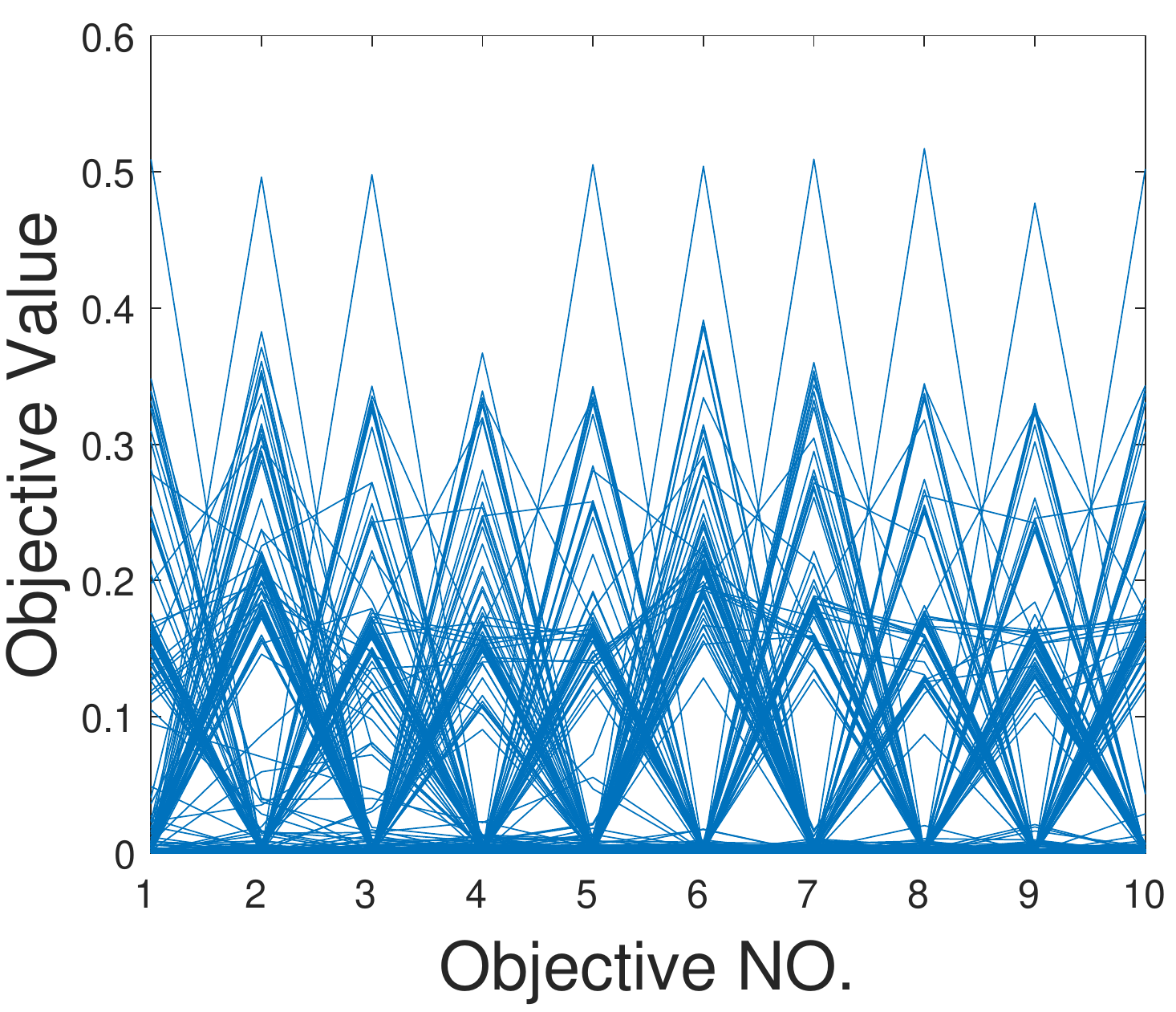}}
      \subfigure[SPEA2+SDE]{\includegraphics[width=4cm]{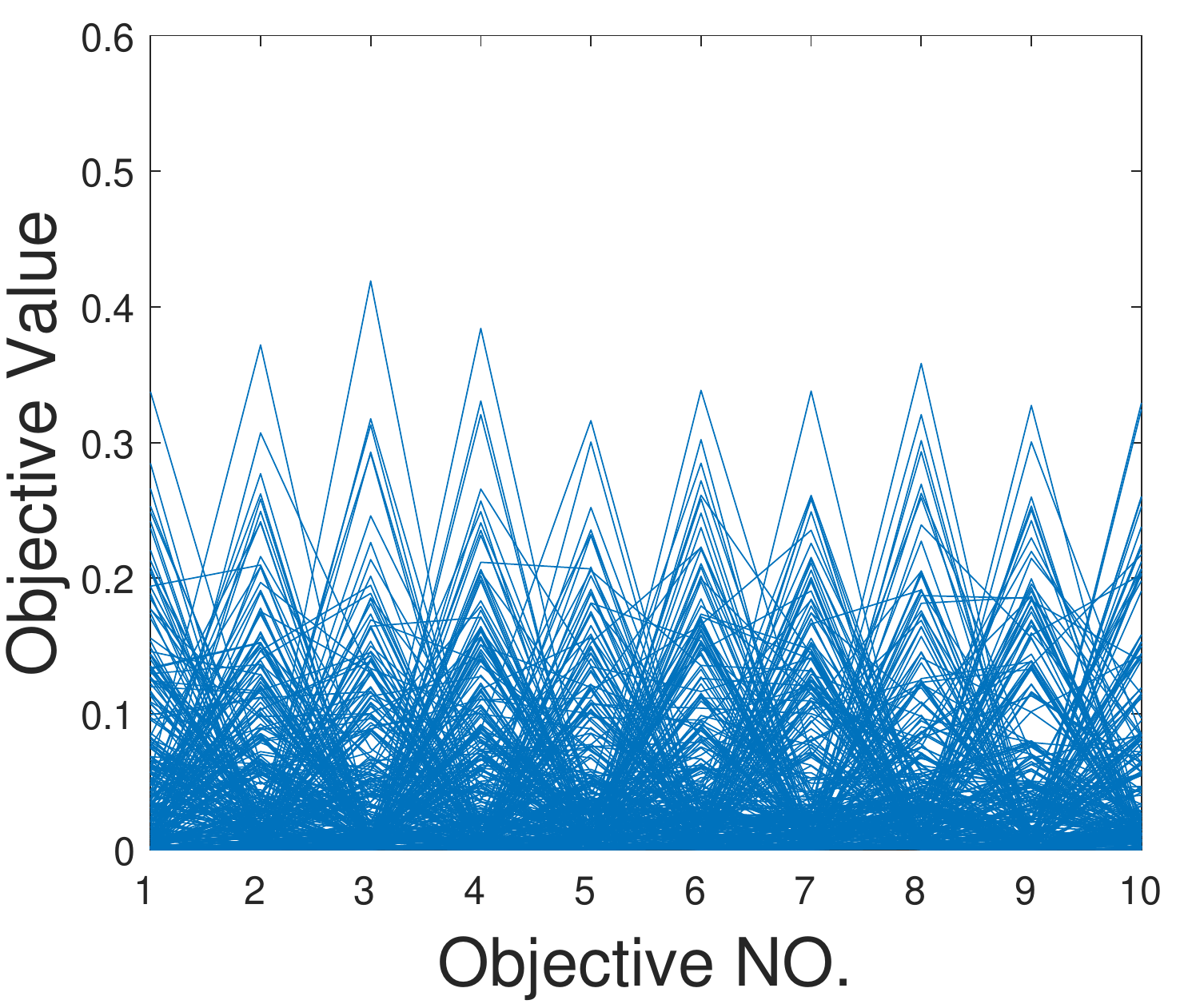}}
      \subfigure[MOEA/D]{\includegraphics[width=4cm]{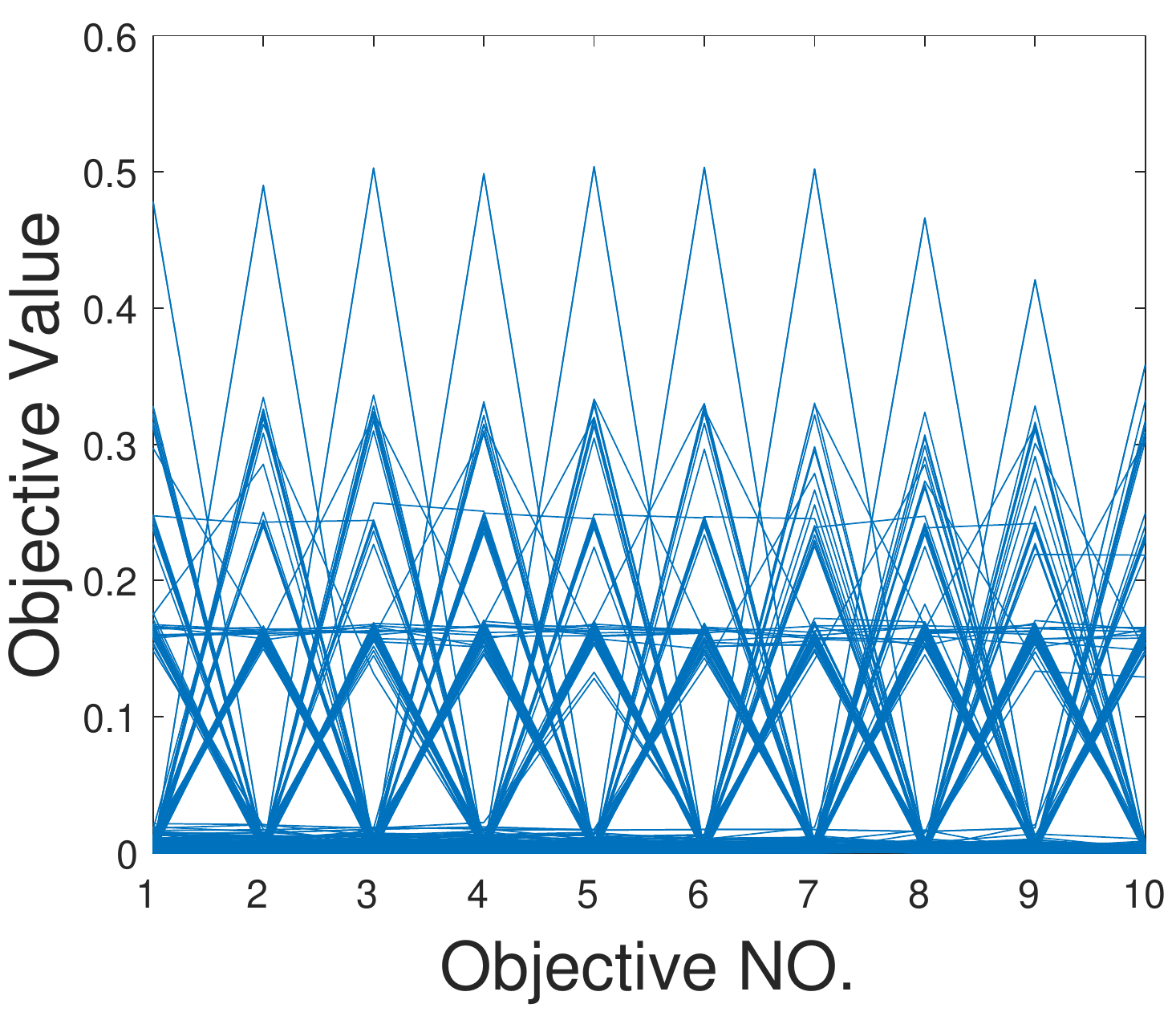}}
      \subfigure[NSGA-III]{\includegraphics[width=4cm]{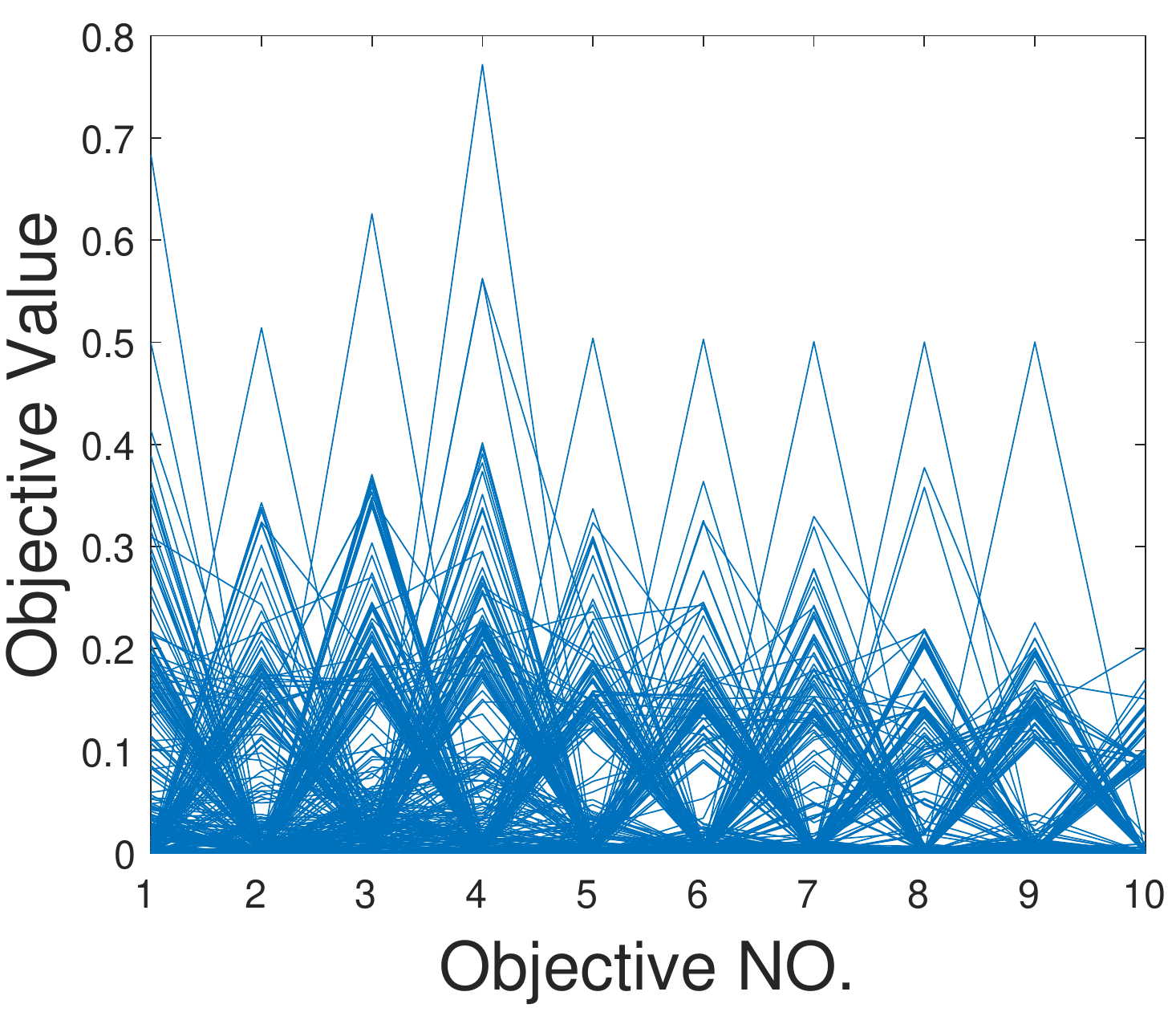}}
      \subfigure[MOMBI-II]{\includegraphics[width=4cm]{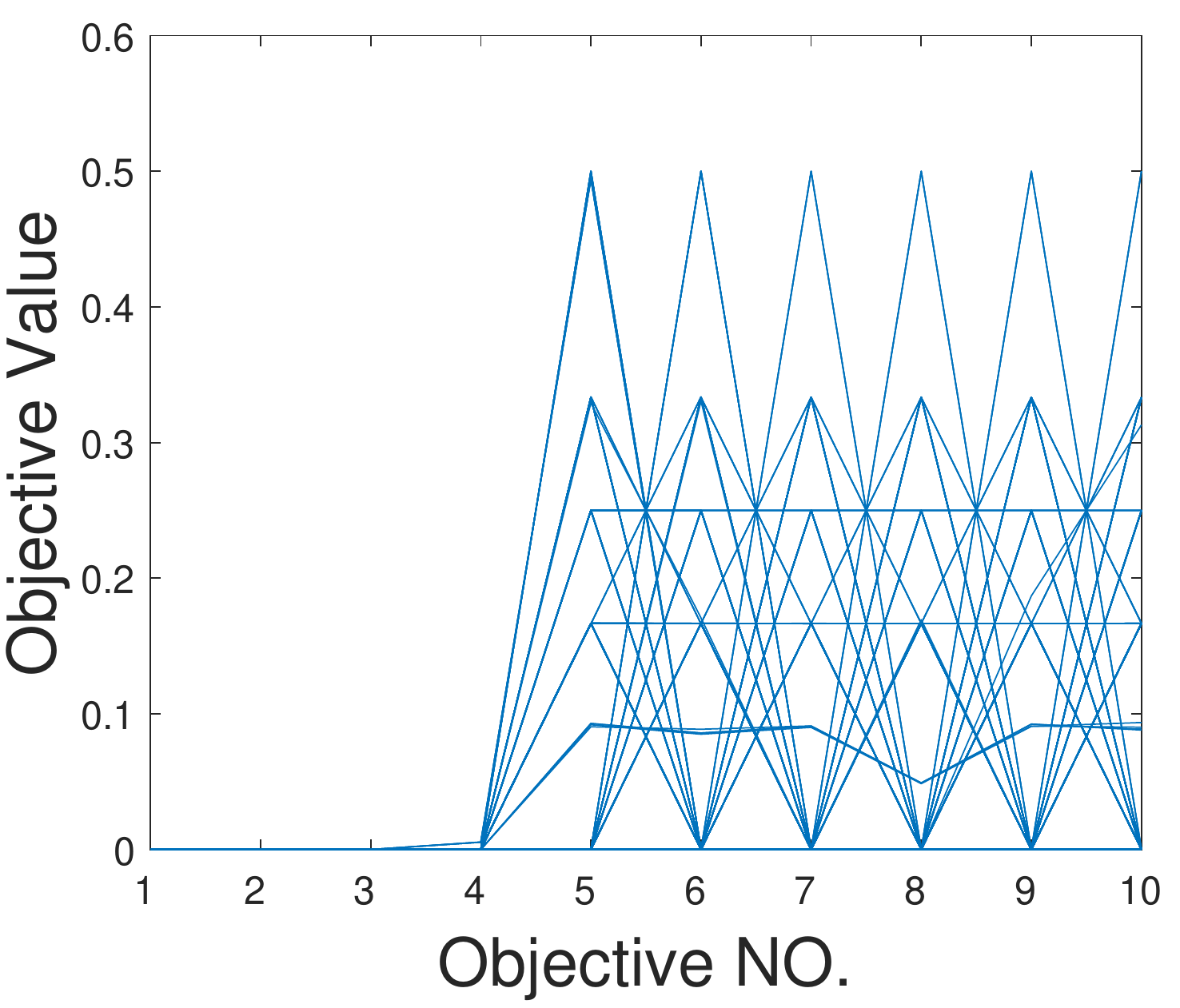}}
      \subfigure[MOEA/DD]{\includegraphics[width=4cm]{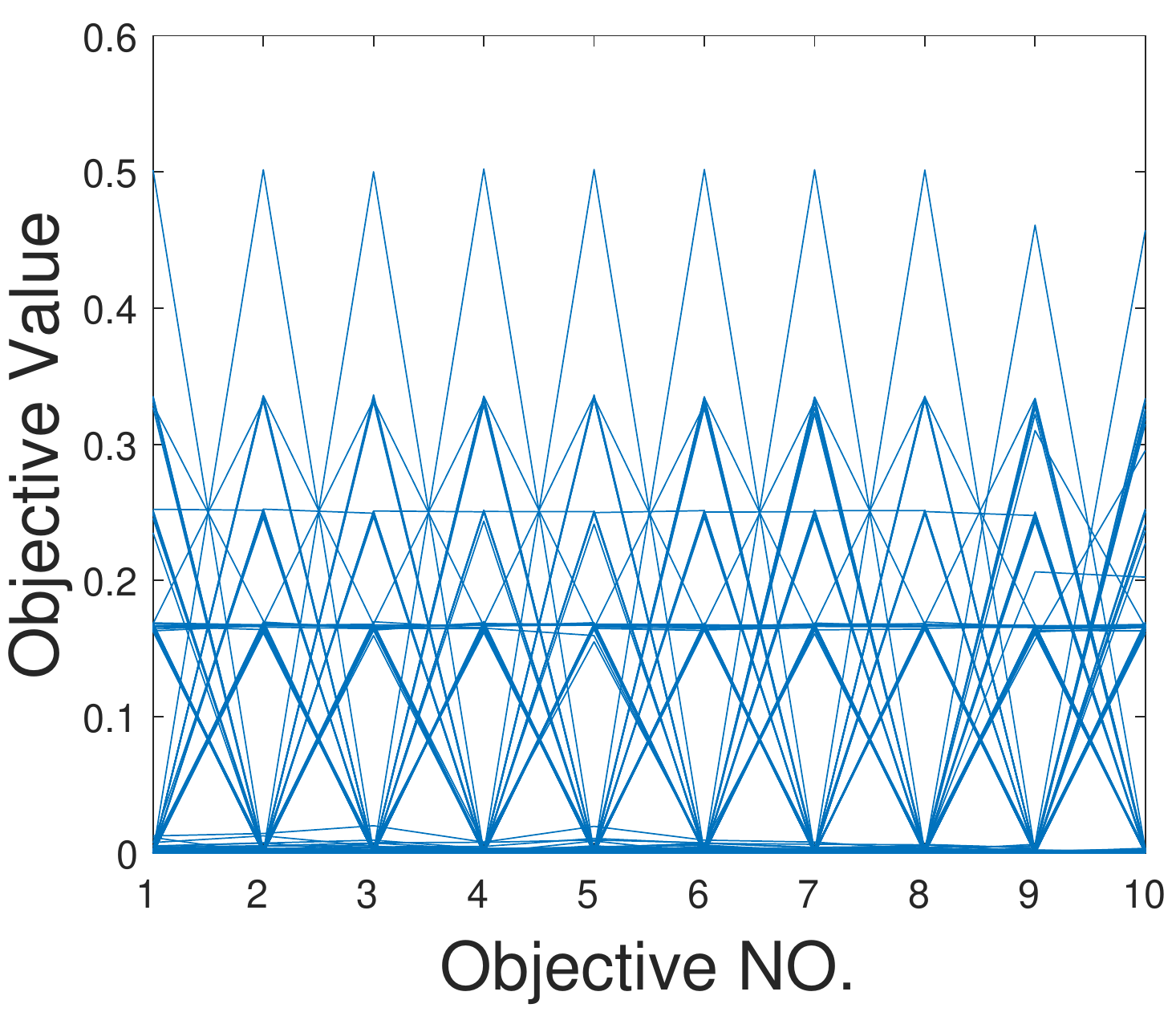}}
      \subfigure[Two\_Arch2]{\includegraphics[width=4cm]{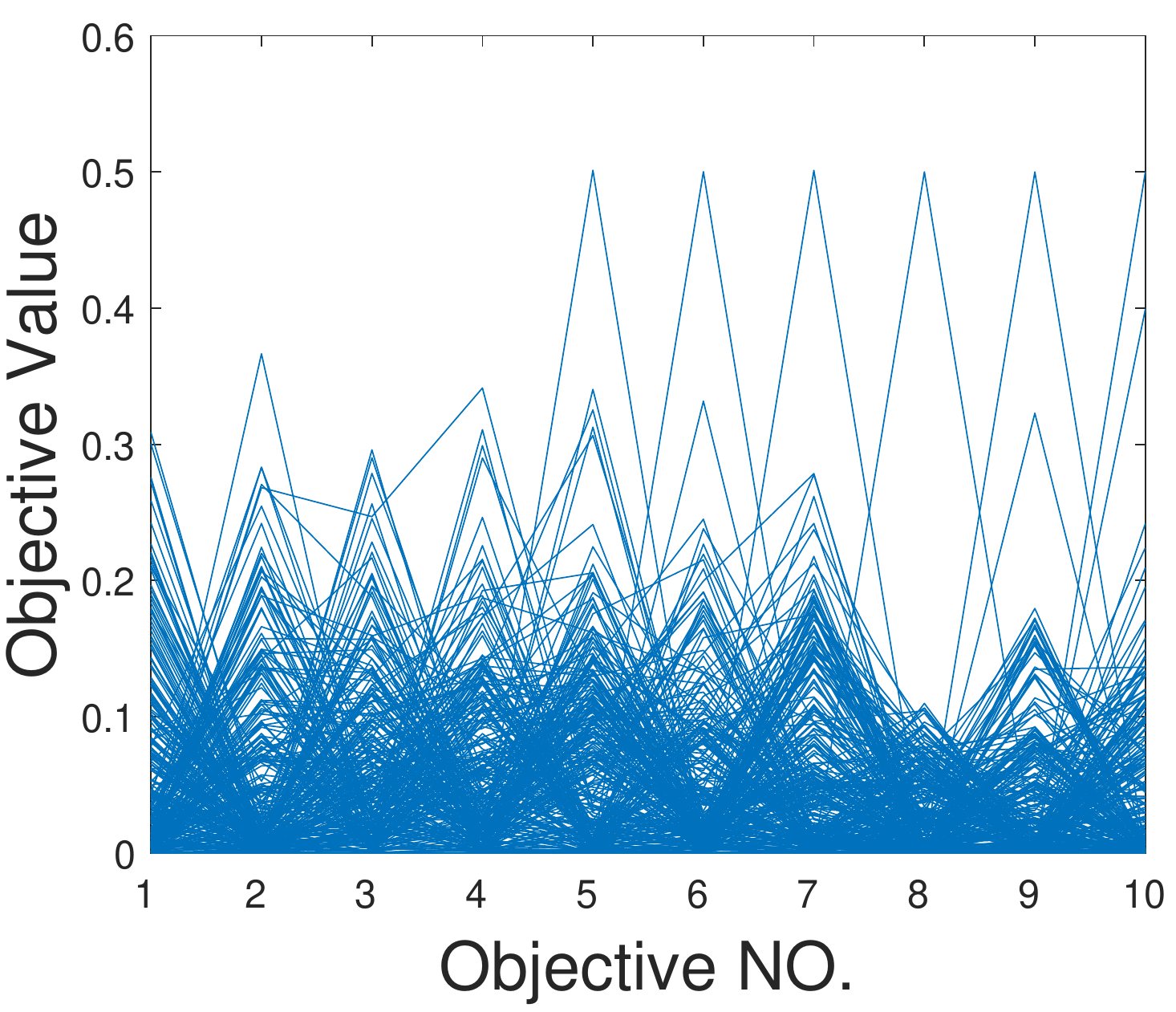}}
      \subfigure[AnD]{\includegraphics[width=4cm]{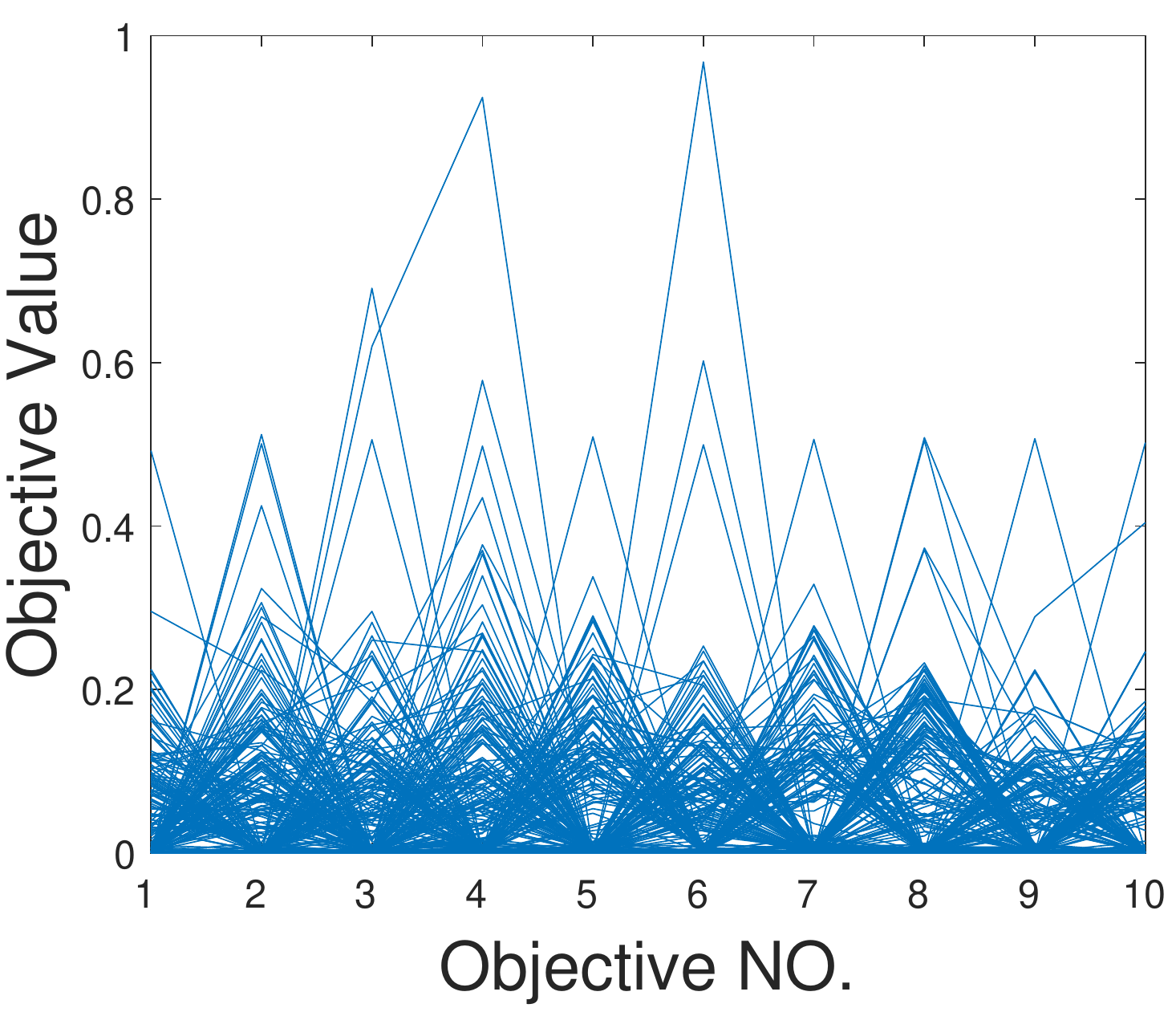}}
       \caption{The final solution sets of the eight compared algorithms on DTLZ1 with ten objectives by parallel coordinates.}\label{fig:dtlzo}
    \end{center}
\end{figure*}

\subsection{Comparison with Seven State-of-the-Art MaOEAs}

Subsequently, we compared the performance of AnD with that of the seven peer algorithms introduced in Section~\ref{sec:peera} on DTLZ and WFG test suites in terms of the IGD and HV metrics. The experimental results are summarized in Tables~\ref{table:IGDstateDTLZ}--\ref{table:HVstateWFG}. At our first glance, RVEA, SPEA2+SDE, and MOEA/DD can achieve superior performance on DTLZ test suite. The reason might be that DTLZ test suite puts more emphasis on an algorithm's convergence ability than diversity ability~\cite{ishibuchi2014review} and AnD is a ``diversity-first-and-convergence-second'' MaOEA. Note, however, that AnD can obtain the best overall performance on WFG test suite. To visualize the experimental results, we plotted the final populations resulting from the eight compared algorithms in a typical run by parallel coordinates on four representative test problems in Figs.~\ref{fig:dtlz1}--\ref{fig:wfg7}. Note that a typical run means a run producing the closest result to the median IGD value among all runs. The detailed discussions are given below.

\subsubsection{DTLZ Test Suite}

\begin{figure*} [!t]
    \begin{center}
      \subfigure[RVEA]{\includegraphics[width=4cm]{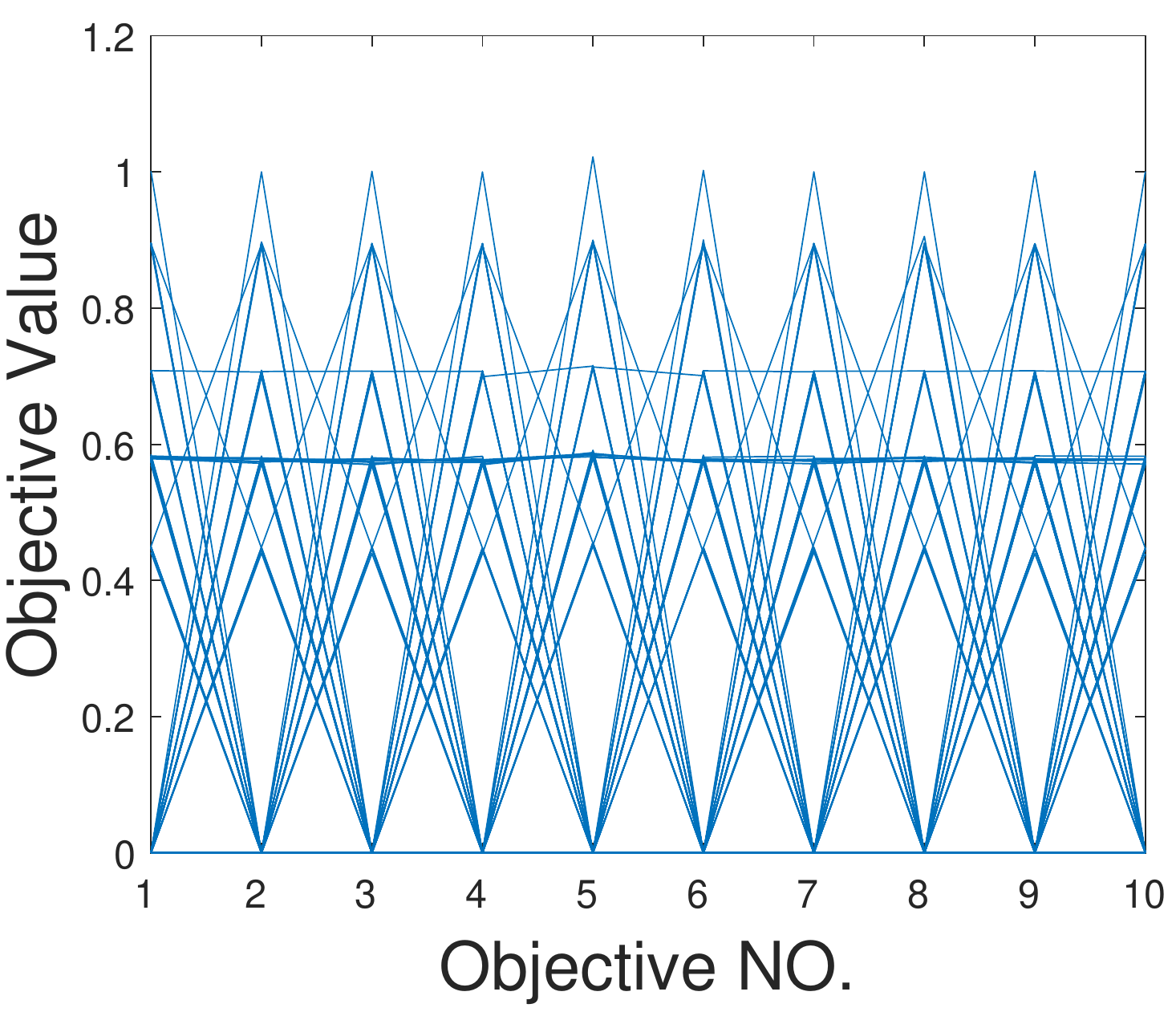}}
      \subfigure[SPEA2+SDE]{\includegraphics[width=4cm]{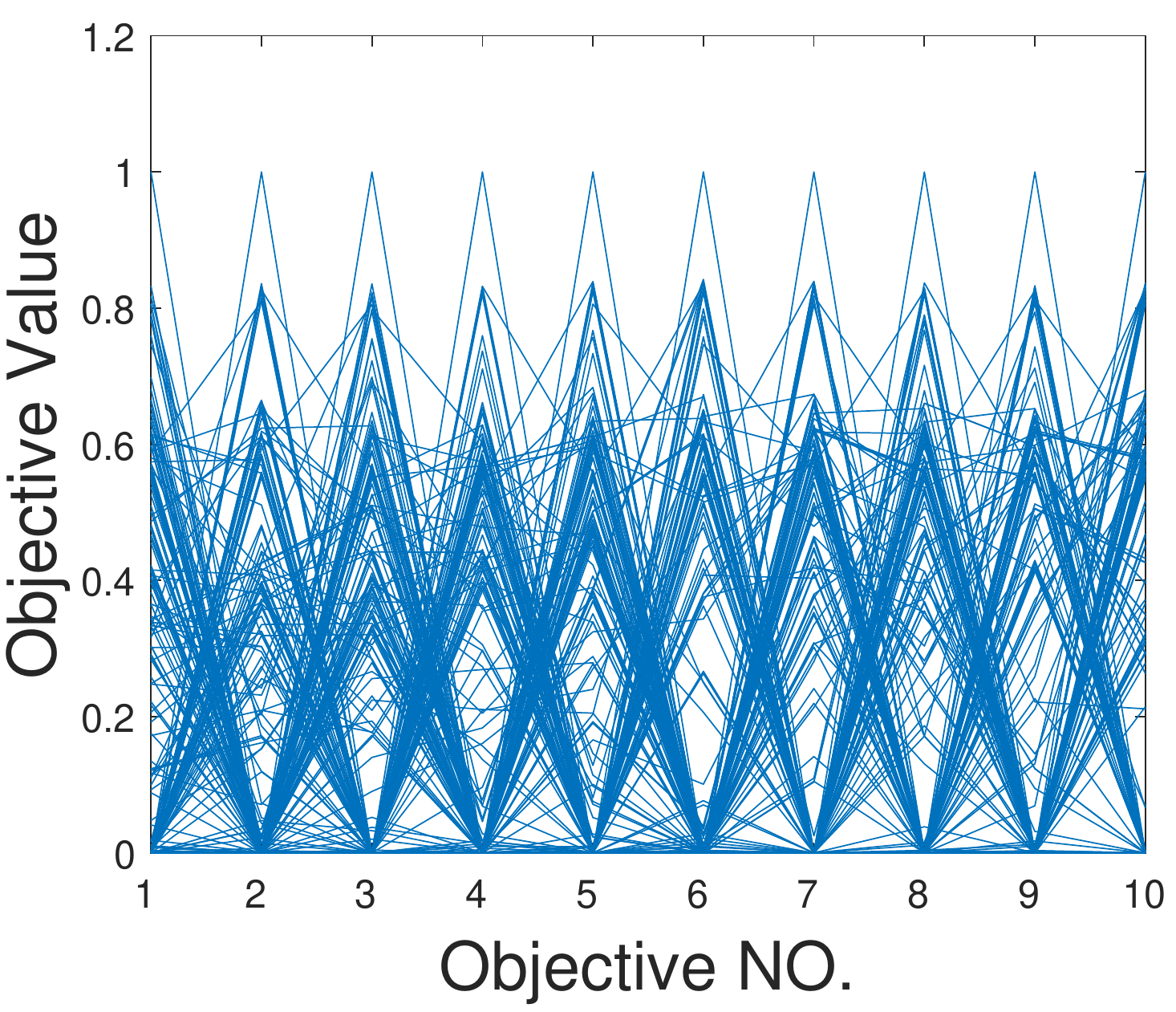}}
      \subfigure[MOEA/D]{\includegraphics[width=4cm]{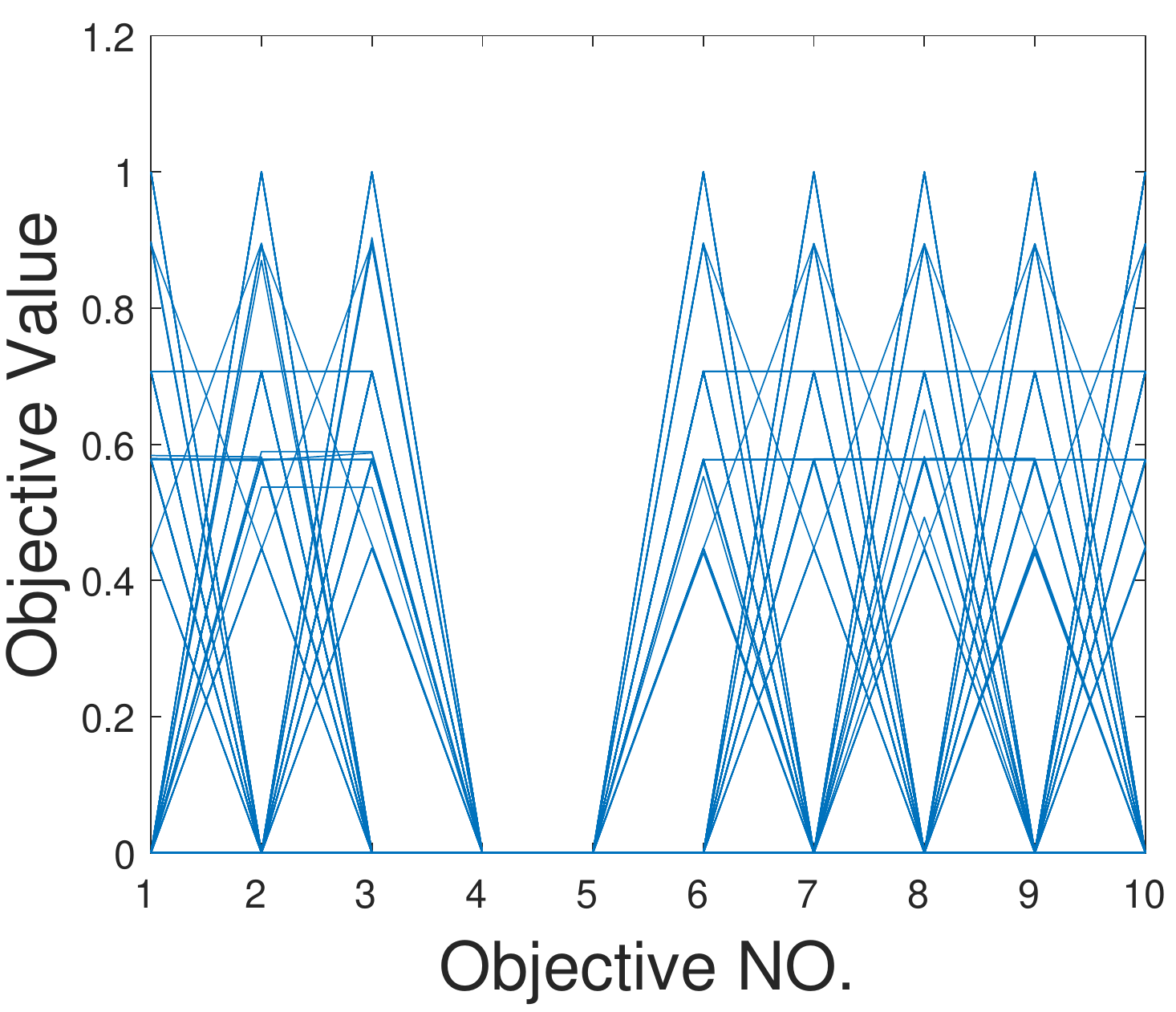}}
      \subfigure[NSGA-III]{\includegraphics[width=4cm]{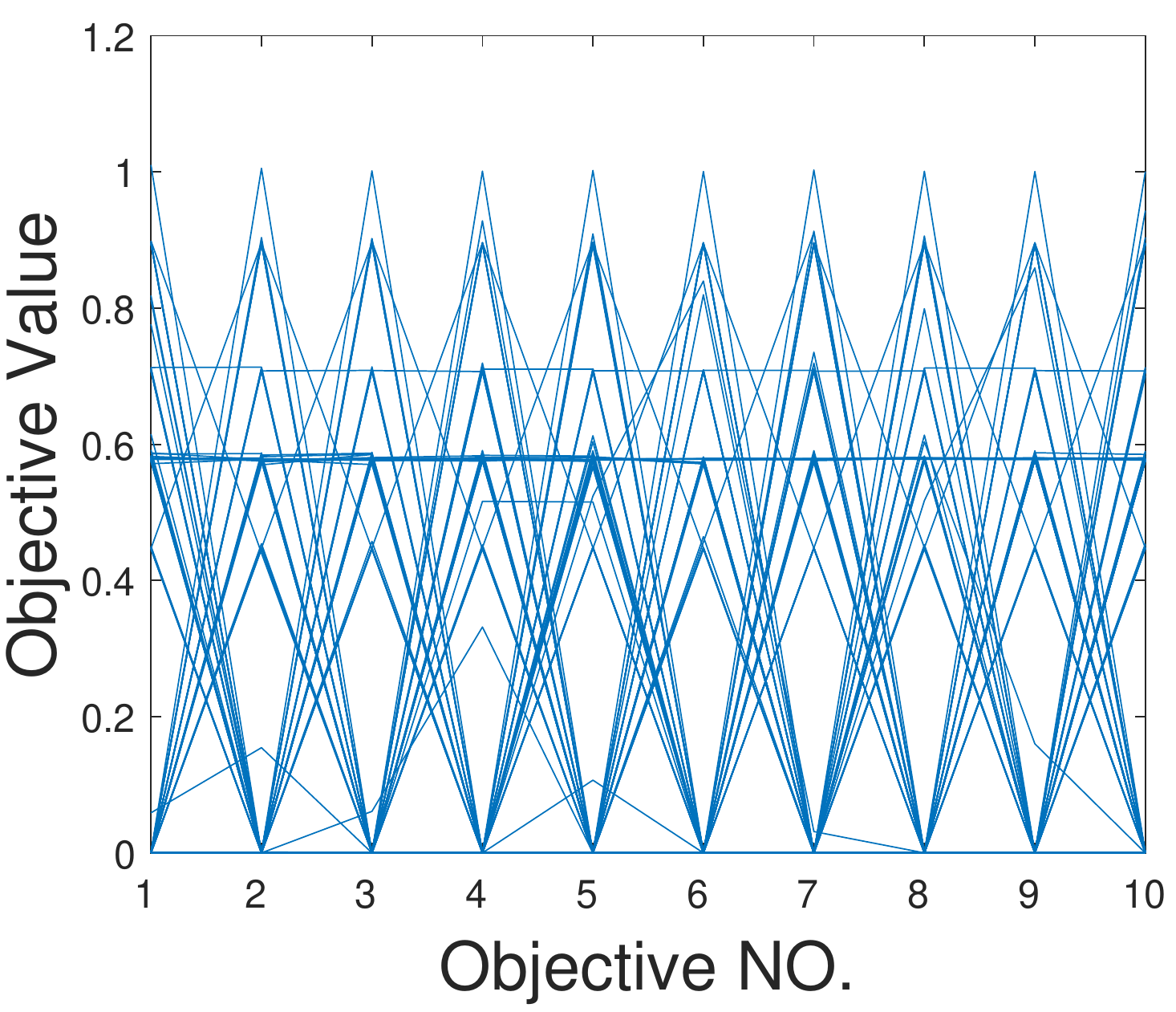}}
      \subfigure[MOMBI-II]{\includegraphics[width=4cm]{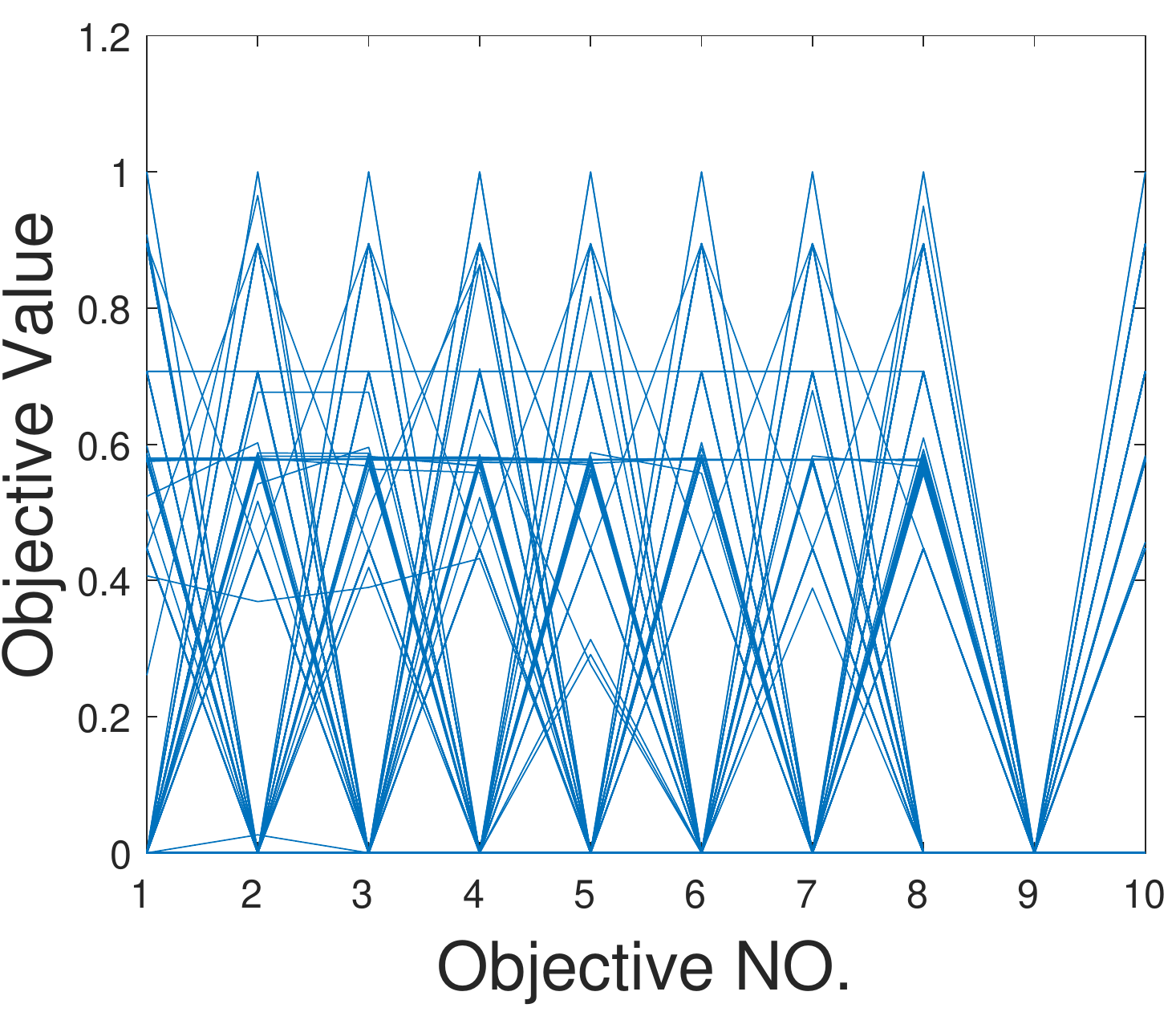}}
      \subfigure[MOEA/DD]{\includegraphics[width=4cm]{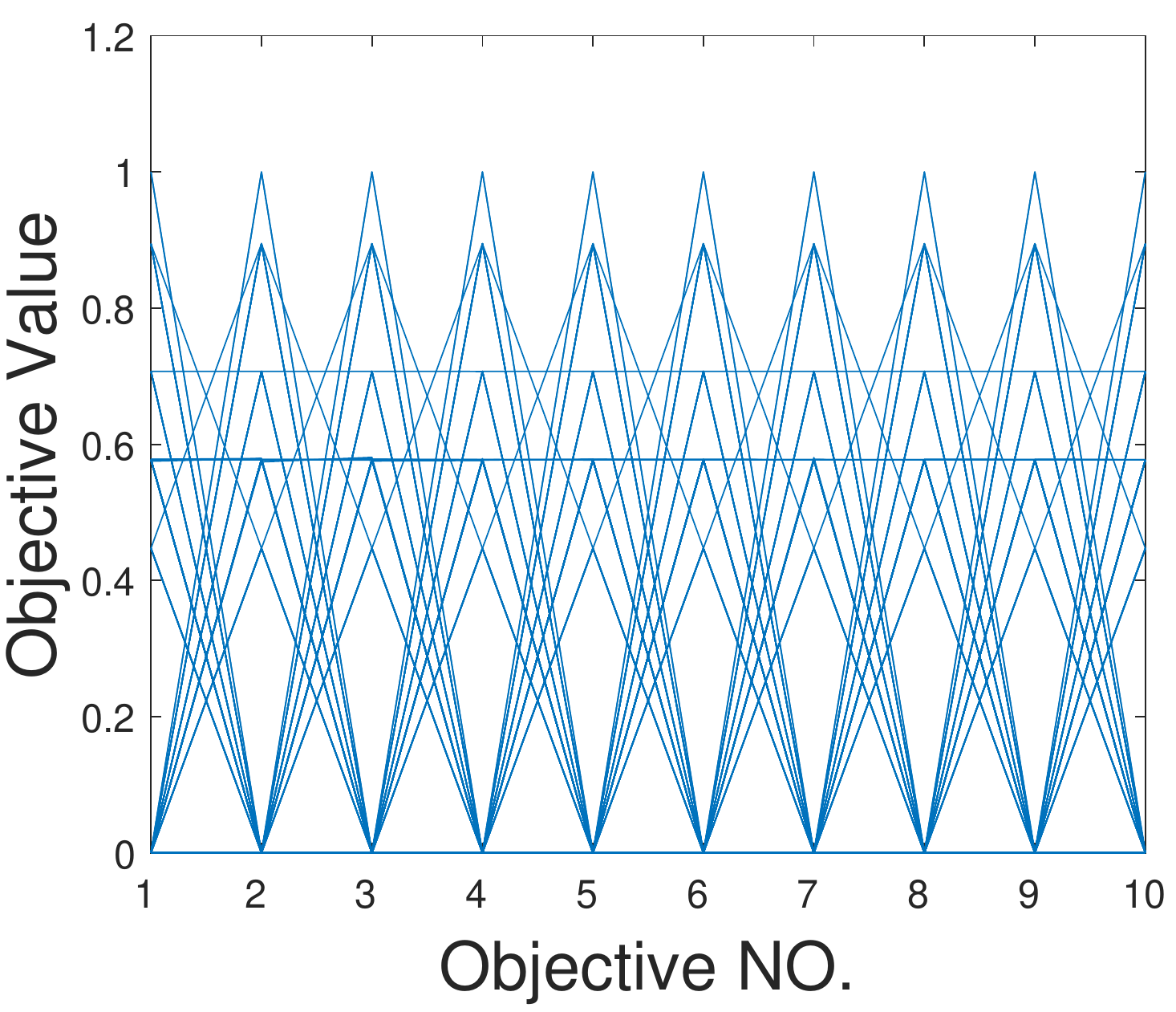}}
      \subfigure[Two\_Arch2]{\includegraphics[width=4cm]{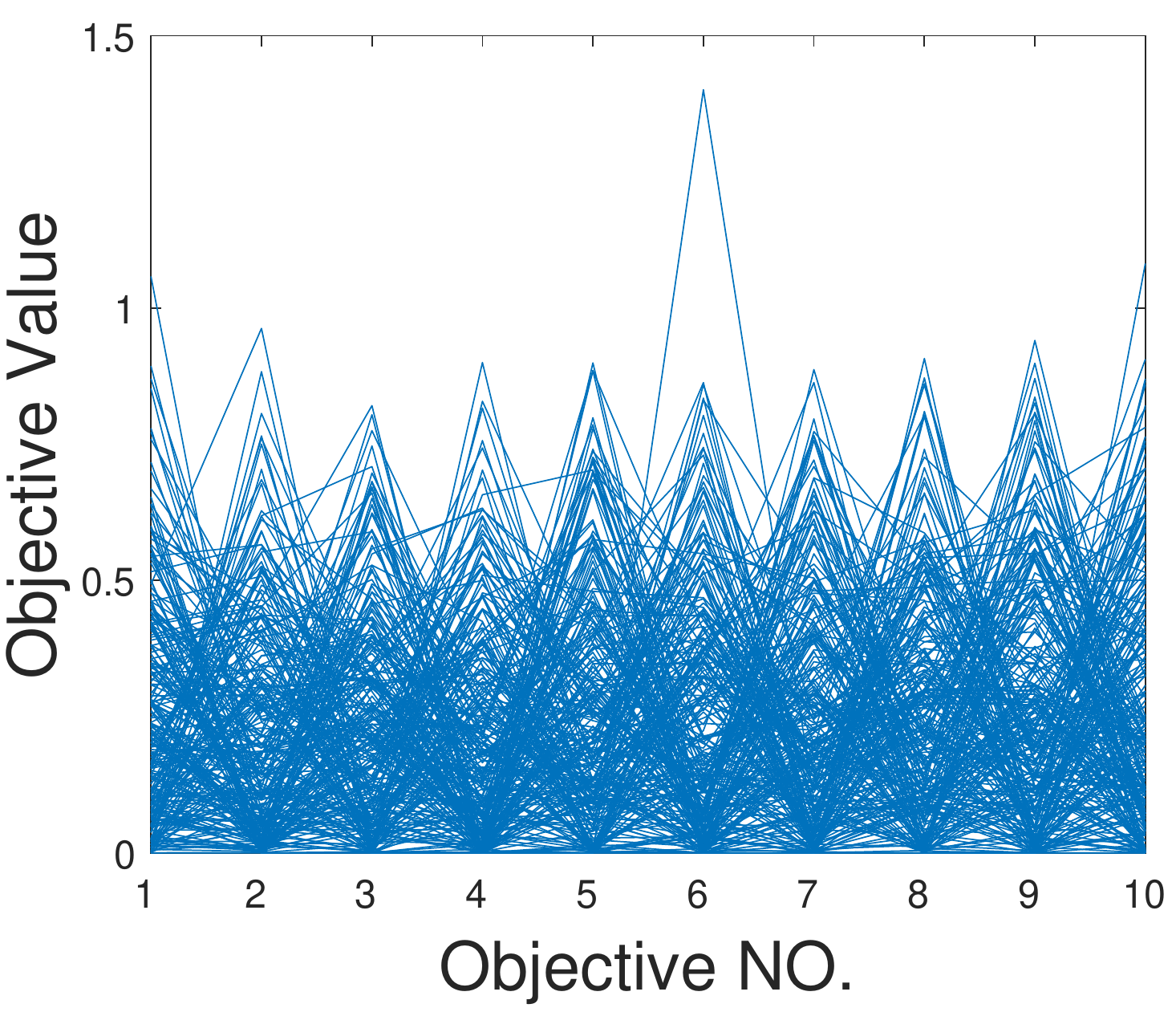}}
      \subfigure[AnD]{\includegraphics[width=4cm]{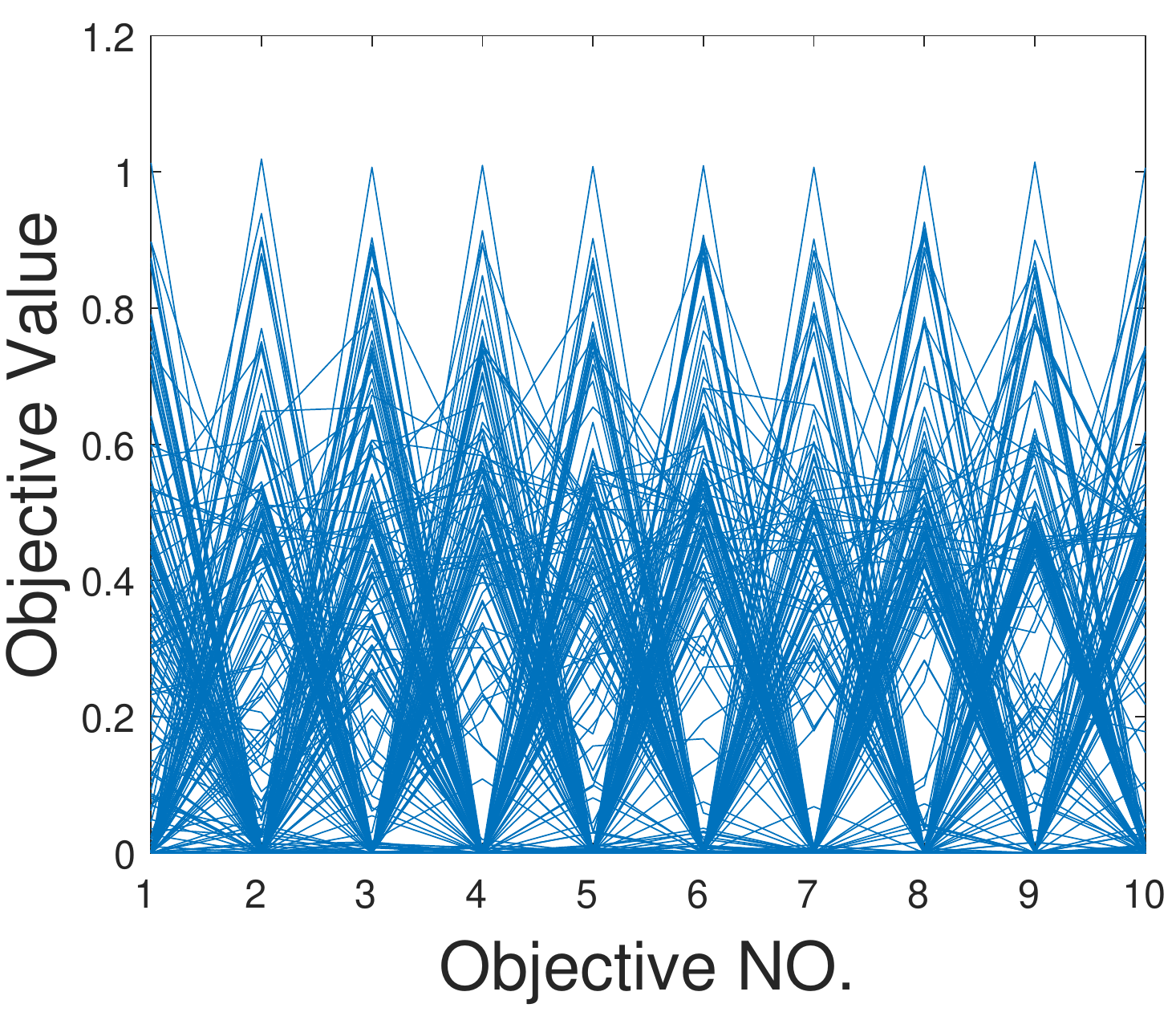}}
       \caption{The final solution sets of the eight compared algorithms on DTLZ4 with ten objectives by parallel coordinates. }\label{fig:dtlzf}
    \end{center}
\end{figure*}

\begin{table*}[!t]
\caption{Performance Comparison between AnD and Seven State-of-the-Art MaOEAs in Terms of the Average IGD Value on WFG Test Suite. The Best and Second Best Average IGD Values among All the Algorithms on Each Test Problem are Highlighted in Gray and Light Gray, Respectively.}
\label{table:IGDstateWFG}
\centering
\begin{scriptsize}
\begin{tabular}{cccc  cccc ccc}
\toprule
Problem&$m$  & RVEA     & SPEA2$+$SDE & MOEA/D & NSGA-III & MOMBI-II & MOEA/DD & Two$\_$Arch2  & AnS\\
\midrule
WFG1 &          5 & $1.2380e+0$   & \cellcolor{gray}$4.5051e-1$   & $1.4719e+0$   & $1.3003e+0$   & \cellcolor{lightgray}$7.3247e-1$   & $1.7863e+0$   & $9.5226e-1$   & $8.2485e-1$  \\
WFG1 &         10 & $2.4159e+0$   &\cellcolor{gray} $1.5483e+0$   & $3.5093e+0$   & $2.3319e+0$   & $2.9849e+0$   & $2.7158e+0$   & $2.2678e+0$   & \cellcolor{lightgray}$1.8344e+0$  \\
WFG1 &         15 & $3.4128e+0$   &\cellcolor{lightgray} $3.0000e+0$   & $5.4507e+0$   & $3.6069e+0$   & $6.0947e+0$   & $4.2376e+0$   & $3.1632e+0$   & \cellcolor{gray}$2.5826e+0$  \\
\multicolumn{10}{ c }{\hdashrule[0.5ex]{17cm}{0.5pt}{0.8mm} } \\
WFG2 &          5 & $7.3061e-1$   & $9.9429e-1$   & $2.9028e+0$   & \cellcolor{lightgray}$6.2450e-1$   & $1.2746e+0$   & $1.9358e+0$   &\cellcolor{gray} $4.7796e-1$   & $7.4199e-1$ \\
WFG2 &         10 & \cellcolor{lightgray}$3.6478e+0$   & $5.4046e+0$   & $1.1021e+1$   & $4.0066e+0$   & $3.9907e+0$   & $1.1009e+1$   &\cellcolor{gray} $1.9916e+0$   & $3.7346e+0$  \\
WFG2 &         15 & $1.2716e+1$   & $1.4549e+1$   & $1.6855e+1$   & $1.2405e+1$   & $1.9653e+1$   & $1.5737e+1$   &\cellcolor{gray} $5.7470e+0$   & \cellcolor{lightgray}$1.2394e+1$  \\
\multicolumn{10}{ c }{\hdashrule[0.5ex]{17cm}{0.5pt}{0.8mm} } \\
WFG3 &          5 & $6.4564e-1$   & $5.8114e-1$   & $1.0504e+0$   & \cellcolor{lightgray}$4.7212e-1$   & $9.5483e-1$   & $6.5805e-1$   & \cellcolor{gray}$3.6891e-1$   & $5.0305e-1$ \\
WFG3 &         10 & $3.0551e+0$   & $1.7028e+0$   & $8.6850e+0$   & \cellcolor{gray}$8.7259e-1$   & $9.3633e+0$   & $2.8279e+0$   & $1.2887e+0$   & \cellcolor{lightgray}$1.7025e+0$ \\
WFG3 &         15 & $6.2924e+0$   & $4.5129e+0$   & $1.6375e+1$   & $3.2111e+0$   & $1.6002e+1$   & $1.4214e+1$   & \cellcolor{gray}$2.5444e+0$   & \cellcolor{lightgray}$2.6156e+0$  \\
\multicolumn{10}{ c }{\hdashrule[0.5ex]{17cm}{0.5pt}{0.8mm} } \\
WFG4 &          5 & \cellcolor{gray}$9.4475e-1$   & $1.1148e+0$   & $1.6793e+0$   & $9.5187e-1$   & $1.7483e+0$   & $1.0314e+0$   & $9.7618e-1$   &\cellcolor{lightgray}$9.5061e-1$  \\
WFG4 &         10 & \cellcolor{lightgray}$3.8034e+0$   & $4.1323e+0$   & $8.9303e+0$   & $4.0807e+0$   & $8.3168e+0$   & $5.2137e+0$   & $4.5821e+0$   &\cellcolor{gray} $3.6441e+0$  \\
WFG4 &         15 & $8.7523e+0$   &\cellcolor{lightgray} $8.2858e+0$   & $1.6011e+1$   & $8.9661e+0$   & $2.3009e+1$   & $9.1948e+0$   & $9.7013e+0$   & \cellcolor{gray}$7.6264e+0$  \\
\multicolumn{10}{ c }{\hdashrule[0.5ex]{17cm}{0.5pt}{0.8mm} } \\
WFG5 &          5 &\cellcolor{gray} $9.3668e-1$   & $1.1233e+0$   & $1.5112e+0$   &\cellcolor{lightgray} $9.3688e-1$   & $1.6929e+0$   & $1.0146e+0$   & $9.7161e-1$   & $9.3925e-1$  \\
WFG5 &         10 & \cellcolor{lightgray}$3.8534e+0$   & $4.0242e+0$   & $8.6325e+0$   & $3.8863e+0$   & $7.2474e+0$   & $5.9057e+0$   & $4.5803e+0$   &\cellcolor{gray} $3.5788e+0$  \\
WFG5 &         15 & \cellcolor{lightgray}$8.2300e+0$   & $1.0363e+1$   & $1.6180e+1$   & $8.6808e+0$   & $2.7050e+1$   & $1.3822e+1$   & $9.6726e+0$   & \cellcolor{gray}$7.5925e+0$  \\
\multicolumn{10}{ c }{\hdashrule[0.5ex]{17cm}{0.5pt}{0.8mm} } \\
WFG6 &          5 &\cellcolor{gray} $9.4927e-1$   & $1.1683e+04$  & $1.9082e+0$   & \cellcolor{lightgray}$9.5167e-1$   & $1.6161e+0$   & $1.0294e+0$   & $9.8289e-1$   & $9.5995e-1$\\
WFG6 &         10 &\cellcolor{lightgray} $3.8553e+0$   & $4.1209e+0$   & $9.7139e+0$   & $3.9288e+0$   & $7.5219e+0$   & $5.4913e+0$   & $4.5845e+0$   &\cellcolor{gray} $3.5574e+0$  \\
WFG6 &         15 & \cellcolor{lightgray}$8.5768e+0$   & $9.3893e+0$   & $1.6228e+1$   & $9.0585e+0$   & $2.3092e+1$   & $1.3704e+1$   & $9.7084e+0$   & \cellcolor{gray}$7.5193e+0$  \\
\multicolumn{10}{ c }{\hdashrule[0.5ex]{17cm}{0.5pt}{0.8mm} } \\
WFG7 &          5 & \cellcolor{gray}$9.5376e-1$   & $1.1481e+0$   & $1.7551e+0$   &\cellcolor{lightgray} $9.5434e-1$   & $1.4084e+0$   & $1.0384e+0$   & $9.5596e-1$   & $9.5631e-1$  \\
WFG7 &         10 &\cellcolor{lightgray} $3.8105e+0$   & $3.9180e+0$   & $9.6364e+0$   & $4.1474e+0$   & $7.0575e+0$   & $4.8005e+0$   & $4.5546e+0$   &\cellcolor{gray} $3.4909e+0$  \\
WFG7 &         15 & $8.4477e+0$   & $8.1509e+0$   & $1.6400e+1$   & $9.2815e+0$   & $2.2087e+1$   &\cellcolor{lightgray} $8.0829e+0$   & $9.9669e+0$   & \cellcolor{gray}$7.5817e+0$ \\
\multicolumn{10}{ c }{\hdashrule[0.5ex]{17cm}{0.5pt}{0.8mm} } \\
WFG8 &          5 & \cellcolor{gray}$9.8533e-1$   & $1.1560e+0$   & $1.2965e+0$   & $1.0084e+0$   & $2.3995e+0$   & $1.0543e+0$   & $1.1102e+0$   & \cellcolor{lightgray}$1.0138e+0$  \\
WFG8 &         10 & \cellcolor{lightgray}$3.9040e+0$   & $4.3343e+0$   & $8.1439e+0$   & $5.1141e+0$   & $8.8767e+0$   & $4.1808e+0$   & $5.4467e+0$   & \cellcolor{gray}$3.8497e+0$  \\
WFG8 &         15 & $9.5526e+0$   &\cellcolor{gray} $8.5588e+0$   & $1.2660e+1$   & $1.0520e+1$   & $2.3945e+1$   & $9.0171e+0$   & $1.0866e+1$   & \cellcolor{lightgray}$8.8015e+0$  \\
\multicolumn{10}{ c }{\hdashrule[0.5ex]{17cm}{0.5pt}{0.8mm} } \\
WFG9 &          5 &\cellcolor{gray} $9.1032e-1$   & $1.0663e+0$   & $1.3802e+0$   &\cellcolor{lightgray} $9.2301e-1$   & $2.2015e+0$   & $1.0173e+0$   & $1.0066e+0$   & $9.4961e-1$  \\
WFG9 &         10 & $4.1319e+0$   & $4.2572e+0$   & $9.0024e+0$   &\cellcolor{lightgray} $4.1250e+0$   & $6.7789e+0$   & $6.0132e+0$   & $5.1443e+0$   &\cellcolor{gray} $3.9489e+0$  \\
WFG9 &         15 & \cellcolor{lightgray}$8.2901e+0$   & $9.1482e+0$   & $1.5257e+1$   & $8.6989e+0$   & $2.7200e+1$   & $1.2606e+1$   & $1.0647e+1$   &\cellcolor{gray} $8.0252e+0$ \\
\bottomrule
\end{tabular}
\end{scriptsize}
\end{table*}

From Tables~\ref{table:IGDstateDTLZ} and~\ref{table:HVstateDTLZ}, we can observe that the eight compared algorithms exhibit mixed performance. More specifically, SPEA2+SDE performs the best in terms of the IGD metric, followed by  MOEA/DD and AnD, as shown in Table~\ref{table:IGDstateDTLZ}. Nevertheless, MOEA/DD and RVEA obtain the best and second best performance with respect to the HV metric, respectively, as shown in Table~\ref{table:HVstateDTLZ}.

For DTLZ1, it is a multimodal problem, in which the PF is degenerate and the decision variables are non-separable. It challenges the convergence performance of an algorithm. We can find that SPEA2+SDE achieves the best IGD values on five and 10 objectives, while MOEA/D obtains the best IGD value on 15 objectives. As far as HV is concerned, MOEA/DD outperforms others on five and 10 objectives, and RVEA beats its competitors on 15 objectives. As shown in Fig.~\ref{fig:dtlz1}, the results provided by some methods using weight vectors or reference points (i.e., MOEA/D and MOEA/DD) have better distributions. With respect to SPEA2+SDE and Two\_Arch2, the scales of some objective values are smaller than the true PF, which suggests that they have a preference on the solutions located in central areas. In terms of AnD, some extreme values for the third, fourth, and sixth objectives occur, which means that AnD has relatively poor convergence performance on DTLZ1.

DTLZ2 is a relatively simple test problem compared with DTLZ1, which is mainly used to test an algorithm's diversity. It is observed that AnD can achieve the best overall IGD performance, while MOEA/DD can obtain the best overall HV performance. It is necessary to note that in AnD the diversity of population is considered in both the angle-based selection and the shift-based density estimation; therefore, AnD provides promising results on DTLZ2.

For DTLZ3, which is a highly multimodal problem, SPEA2+SDE can achieve the best overall performance in terms of both IGD and HV. From Table~\ref{table:HVstateDTLZ}, it can be seen that the results provided by NSGA-III and Two\_Arch2 are distant from the PFs on 10 and 15 objectives. For AnD, it can obtain medium performance in terms of both IGD and HV.

Regarding DTLZ4, the density of points on its PF is strongly biased. Therefore, the main challenge for solving this test problem is to maintain the diversity of population. Similar to DTLZ2, AnD and MOEA/DD obtain the best overall IGD and HV performance, respectively. From Fig.~\ref{fig:dtlz4}, it can be observed that Two\_Arch2 produces some extreme values on the sixth objective, which suggests the unstable convergence performance. The results obtained by RVEA, NSGA-III, and MOEA/DD distribute similarly and concentrate mainly on the boundary or the middle parts of the PF. As for MOEA/D, its obtained results fail to cover the forth and fifth objectives well. Similarly, the results obtained by MOMBI-II are unable to cover the ninth objective well. With respect to SPEA2+SDE and AnD, they are capable of covering the whole PF. The difference between them is that the objective values derived from SPEA2+SDE mainly lie in [0, 0.8], while in AnD, they are well-distributed within [0, 1].

\begin{table*}[!t]
\caption{Performance Comparison  between AnD and Seven State-of-the-Art MaOEAs in Terms of the Average HV Value on WFG Test Suite. The Best and Second Best Average HV Values among All the Algorithms on Each Test Problem are Highlighted in Gray and Light Gray, Respectively.}
\label{table:HVstateWFG}
\centering
\begin{scriptsize}
\begin{tabular}{cccc  cccc cc}
\toprule
Problem&$m$  & RVEA     & SPEA2$+$SDE & MOEA/D & NSGA-III & MOMBI-II & MOEA/DD & Two$\_$Arch2  & AnS\\
\midrule
WFG1 &          5 & $5.2614e-1$  & \cellcolor{lightgray}$8.5804e-1$   & $6.9318e-1$   & $5.0964e-1$ &\cellcolor{gray} $9.8249e-1$   & $3.7722e-1$  & $6.3042e-1$  & $6.7931e-1$ \\
WFG1 &         10 & $3.3523e-1$  & \cellcolor{lightgray}$6.1873e-1$   & $4.5187e-1$ & $4.2400e-1$  & \cellcolor{gray}$9.9177e-1$   & $2.8866e-1$  & $3.9733e-1$  & $5.1846e-1$ \\
WFG1 &         15 & $6.4838e-1$  & $6.4360e-1$   & $2.5320e-1$ & $5.6188e-1$  &\cellcolor{lightgray} $8.3535e-1$   &\cellcolor{gray} $8.5739e-1$   & $4.8177e-1$ & $8.1532e-1$ \\
\multicolumn{10}{ c }{\hdashrule[0.5ex]{17cm}{0.5pt}{0.8mm} } \\
WFG2 &          5 & $9.5415e-1$  & $9.4272e-1$   & $7.4322e-1$  & $9.6190e-1$  & $9.7075e-1$   & $9.2248e-1$  & \cellcolor{lightgray}$9.7553e-1$ & \cellcolor{gray}$9.8630e-1$ \\
WFG2 &         10 & $9.0610e-1$  &\cellcolor{lightgray} $9.5588e-1$   & $7.3343e-1$  & $9.4637e-1$   & $7.5468e-1$  & $8.8639e-1$  & $9.4264e-1$  & \cellcolor{gray}$9.8332e-1$\\
WFG2 &         15 & $7.6722e-1$  & $9.3104e-1$   & $6.7549e-1$  & $9.1018e-1$  & $5.4736e-1$  & $8.0398e-1$  & \cellcolor{lightgray}$9.7373e-1$  & \cellcolor{gray}$9.7782e-1$ \\
\multicolumn{10}{ c }{\hdashrule[0.5ex]{17cm}{0.5pt}{0.8mm} } \\
WFG3 &          5 & $2.5974e-2$  & $6.7116e-1$   & $5.9385e-1$  & \cellcolor{lightgray}$7.1066e-1$  & $6.7204e-1$   & $6.6812e-1$  & \cellcolor{gray}$7.2408e-1$  & $6.9653e-1$  \\
WFG3 &         10 & $3.7874e-1$  & $5.6384e-1$   & $2.2239e-1$ & $6.0532e-1$   & $2.7517e-1$  & $2.5414e-1$  & \cellcolor{gray}$7.3015e-1$  & \cellcolor{lightgray}$6.2796e-1$ \\
WFG3 &         15 & $4.6421e-1$  & $6.9188e-1$   & $3.6466e-1$  &\cellcolor{lightgray} $7.3573e-1$   & $3.2567e-1$  & $6.5152e-1$  & \cellcolor{gray}$7.5401e-1$   & $7.0778e-1$   \\
\multicolumn{10}{ c }{\hdashrule[0.5ex]{17cm}{0.5pt}{0.8mm} } \\
WFG4 &          5 & $7.4868e-1$  & \cellcolor{lightgray}$7.5476e-1$   & $6.4367e-1$  & $7.4082e-1$  & $6.6036e-1$  & $7.3706e-1$  & $7.1734e-1$ & \cellcolor{gray}$7.5858e-1$ \\
WFG4 &         10 & $8.3259e-1$  & $8.2460e-1$    & $3.7213e-1$  &\cellcolor{lightgray} $8.6530e-1$   & $6.2085e-1$  & $7.8810e-1$  & $6.7389e-1$  & \cellcolor{gray}$8.6804e-1$ \\
WFG4 &         15 & $7.6663e-1$  & \cellcolor{lightgray}$8.7028e-1$   & $1.6598e-1$  & $7.8689e-1$  & $3.3596e-1$  & $8.6369e-1$ & $6.1921e-1$  &\cellcolor{gray} $9.0693e-1$ \\
\multicolumn{10}{ c }{\hdashrule[0.5ex]{17cm}{0.5pt}{0.8mm} } \\
WFG5 &          5 & \cellcolor{gray}$7.3031e-1$   & $7.1458e-1$  & $6.4534e-1$  & \cellcolor{lightgray}$7.2793e-1$   & $6.1046e-1$  & $7.0839e-1$ & $6.8516e-1$  & $7.2568e-1$ \\
WFG5 &         10 &\cellcolor{lightgray} $8.4286e-1$   & $7.9010e-1$  & $3.7997e-1$  & \cellcolor{gray}$8.4800e-1$   & $5.5075e-1$  & $6.8552e-1$  & $6.1423e-1$  & $8.3440e-1$ \\
WFG5 &         15 & \cellcolor{lightgray}$8.4490e-1$   & $7.7143e-1$  & $1.1919e-1$  & $7.8855e-1$   & $2.1399e-1$  & $3.9362e-1$  & $5.1556e-1$  &\cellcolor{gray} $8.5625e-1$ \\
\multicolumn{10}{ c }{\hdashrule[0.5ex]{17cm}{0.5pt}{0.8mm} } \\
WFG6 &          5 & \cellcolor{lightgray}$7.3194e-1$  & $7.3388e-1$   & $5.4052e-1$  & $7.2387e-1$  & $6.5647e-1$  & $7.1624e-1$  & $6.8634e-1$  &\cellcolor{gray} $7.3791e-1$ \\
WFG6 &         10 & \cellcolor{gray}$8.6056e-1$   & $8.2632e-1$  & $1.9400e-1$  &\cellcolor{lightgray} $8.5875e-1$   & $6.2132e-1$  & $7.5777e-1$  & $6.0675e-1$  & $8.5307e-1$ \\
WFG6 &         15 & \cellcolor{lightgray}$8.6801e-1$  & $8.4484e-1$  & $9.4497e-2$   & $8.5336e-1$  & $3.3801e-1$ & $4.4625e-1$ & $5.1359e-1$  &\cellcolor{gray} $8.9315e-1$ \\
\multicolumn{10}{ c }{\hdashrule[0.5ex]{17cm}{0.5pt}{0.8mm} } \\
WFG7 &          5 & \cellcolor{lightgray}$7.8637e-1$  & $7.8067e-1$  & $6.3867e-1$  & $7.7461e-1$  & $7.6659e-1$  & $7.6365e-1$  & $7.5484e-1$  & \cellcolor{gray}$7.9304e-1$ \\
WFG7 &         10 & $8.8809e-1$  & $8.9075e-1$  & $2.3503e-1$  & $\cellcolor{lightgray}9.1733e-1$  & $7.4944e-1$  & $8.5089e-1$  & $6.6219e-1$  &\cellcolor{gray} $9.2941e-1$ \\
WFG7 &         15 & $7.2147e-1$  &\cellcolor{lightgray} $9.2918e-1$  & $1.2338e-1$ & $8.3698e-1$  & $4.2995e-1$  & $9.1768e-1$  & $6.0513e-1$  &\cellcolor{gray} $9.7162e-1$ \\
\multicolumn{10}{ c }{\hdashrule[0.5ex]{17cm}{0.5pt}{0.8mm} } \\
WFG8 &          5 & $6.4019e-1$  & \cellcolor{gray}$6.7275e-1$   & $5.0839e-1$  & $6.5130e-1$  & $3.5919e-1$  & $6.3295e-1$  & $6.1111e-1$  & \cellcolor{lightgray}$6.6159e-1$ \\
WFG8 &         10 & $6.0051e-1$  &\cellcolor{gray} $8.0185e-1$   & $7.3417e-2$   & \cellcolor{lightgray}$7.8469e-1$   & $4.6721e-1$  & $7.0977e-1$   & $4.4208e-1$  & $7.4075e-1$ \\
WFG8 &         15 & $3.8537e-1$  &\cellcolor{lightgray} $8.7264e-1$  & $3.4644e-1$  & $7.3358e-1$  & $2.9260e-1$  & $8.6234e-1$  & $3.5437e-1$ &\cellcolor{gray} $8.9873e-1$ \\
\multicolumn{10}{ c }{\hdashrule[0.5ex]{17cm}{0.5pt}{0.8mm} } \\
WFG9 &          5 &\cellcolor{gray} $6.8336e-1$   & \cellcolor{lightgray}$6.7929e-1$   & $5.8874e-1$  & $6.4825e-1$  & $3.8392e-1$  & $6.3831e-1$  & $6.3829e-1$  & $6.6130e-1$ \\
WFG9 &         10 & $7.0854e-1$  &\cellcolor{gray} $7.4727e-1$   & $1.5646e-1$  & $7.3520e-1$   & $5.1312e-1$  & $5.2354e-1$  & $5.6358e-1$  & \cellcolor{lightgray}$7.3830e-1$ \\
WFG9 &         15 & $6.1502e-1$  & $7.1075e-1$   & $6.0542e-2 $  &\cellcolor{lightgray} $7.1895e-1$   & $2.0338e-1$  & $2.4844e-1$  & $4.4957e-1$  & \cellcolor{gray}$7.2111e-1$ \\
\bottomrule
\end{tabular}
\end{scriptsize}
\end{table*}

\begin{figure*} [!t]
    \begin{center}
      \subfigure[RVEA]{\includegraphics[width=4cm]{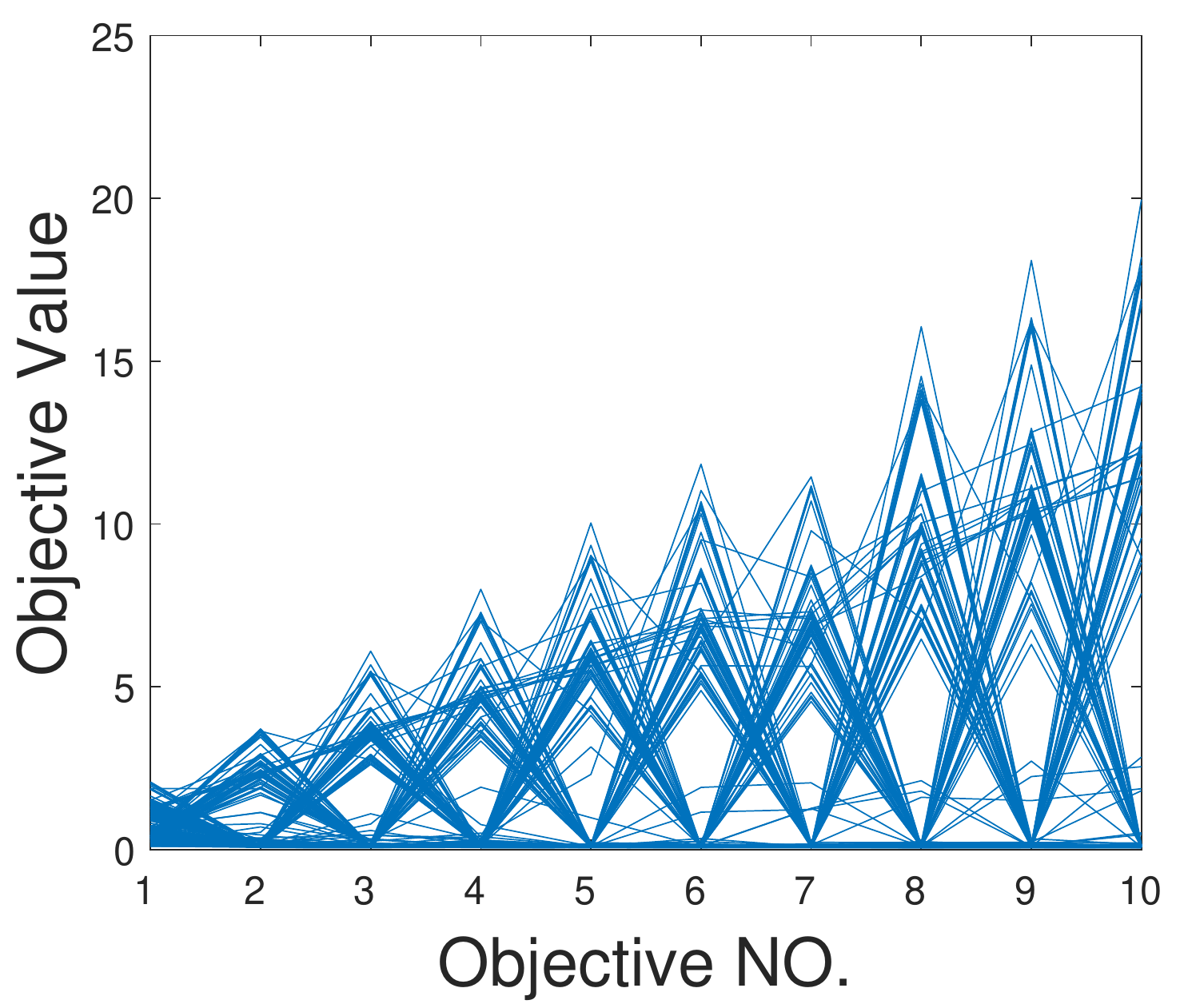}}
      \subfigure[SPEA2+SDE]{\includegraphics[width=4cm]{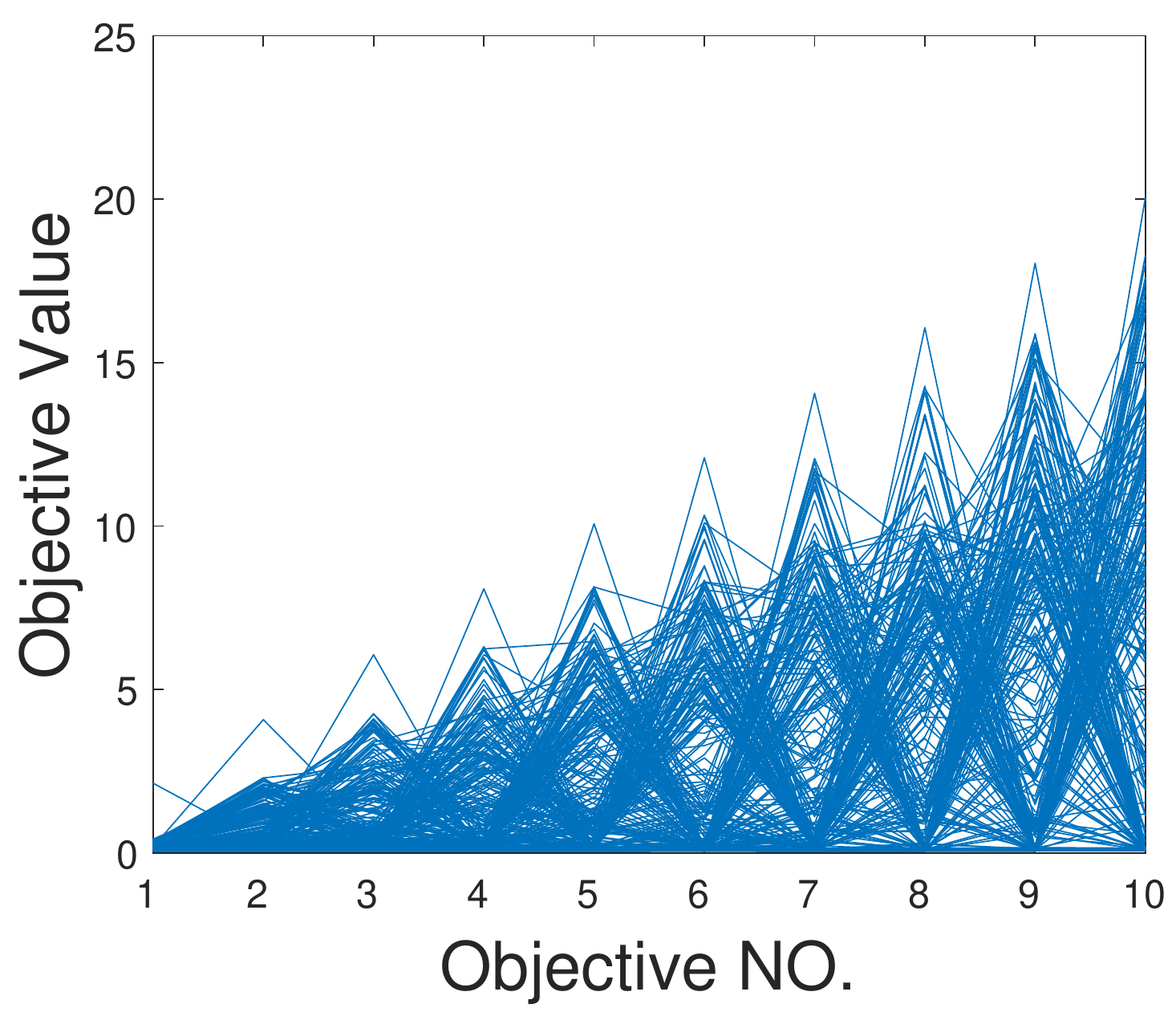}}
      \subfigure[MOEA/D]{\includegraphics[width=4cm]{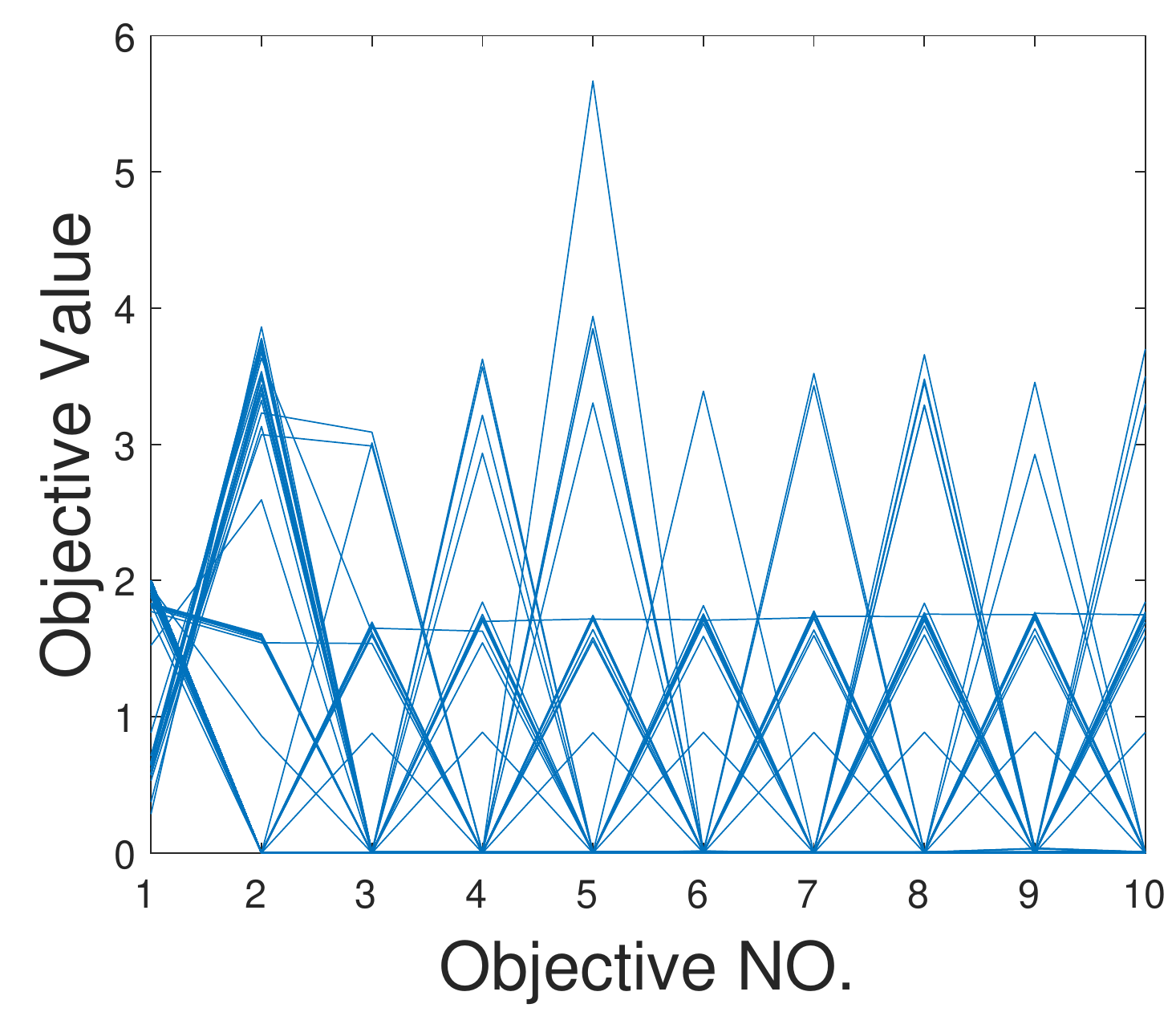}}
      \subfigure[NSGA-III]{\includegraphics[width=4cm]{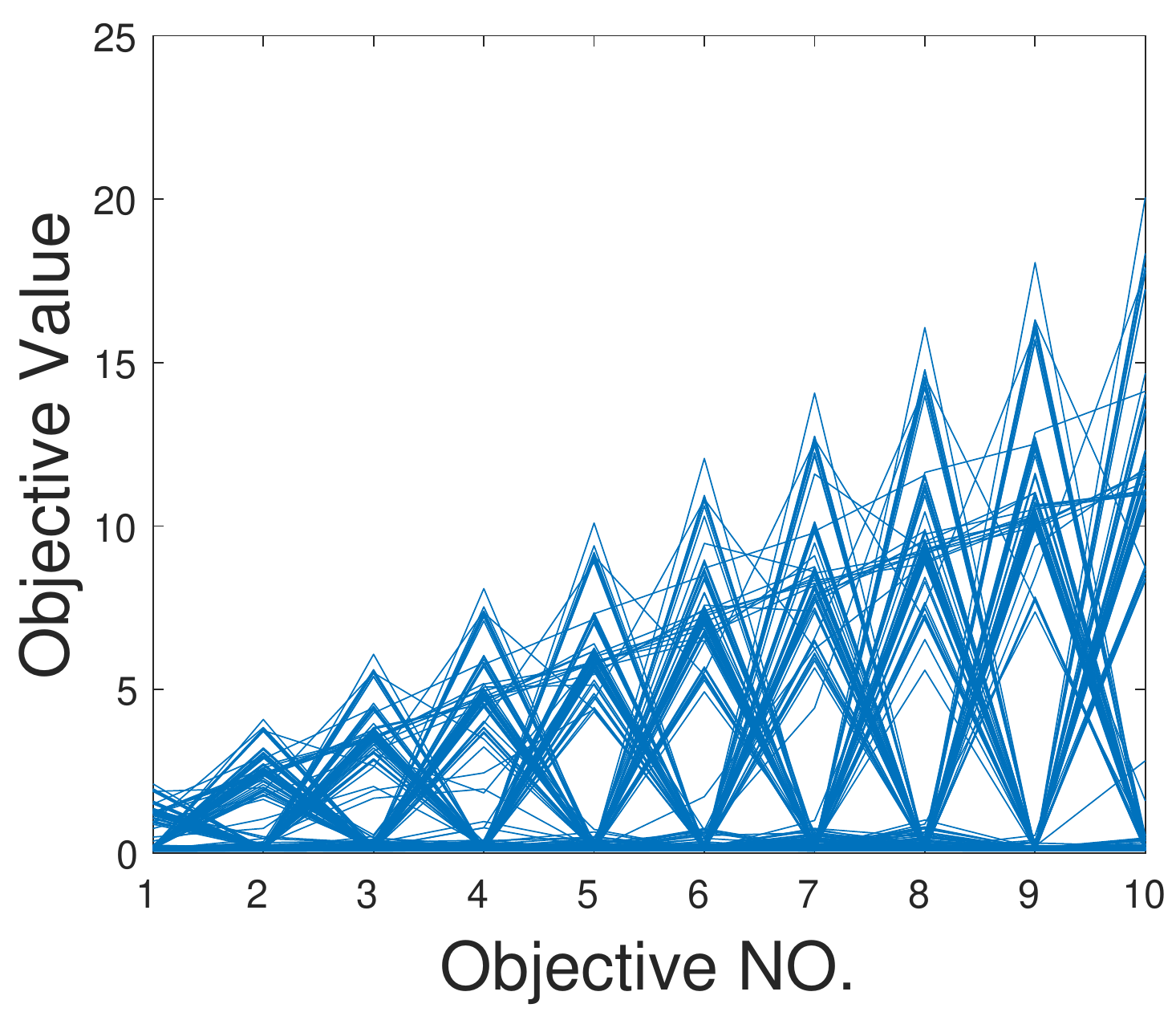}}
      \subfigure[MOMBI-II]{\includegraphics[width=4cm]{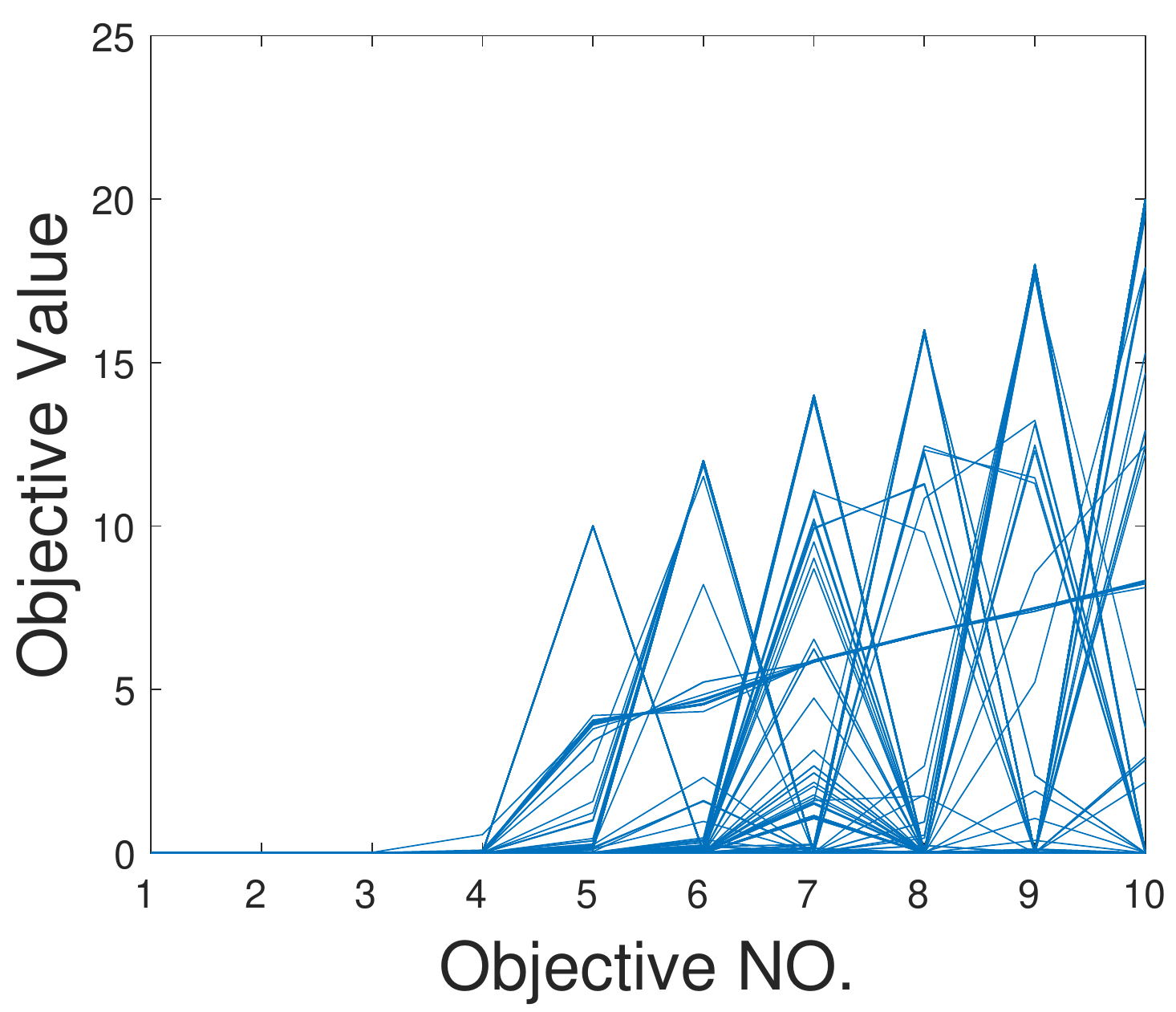}}
      \subfigure[MOEA/DD]{\includegraphics[width=4cm]{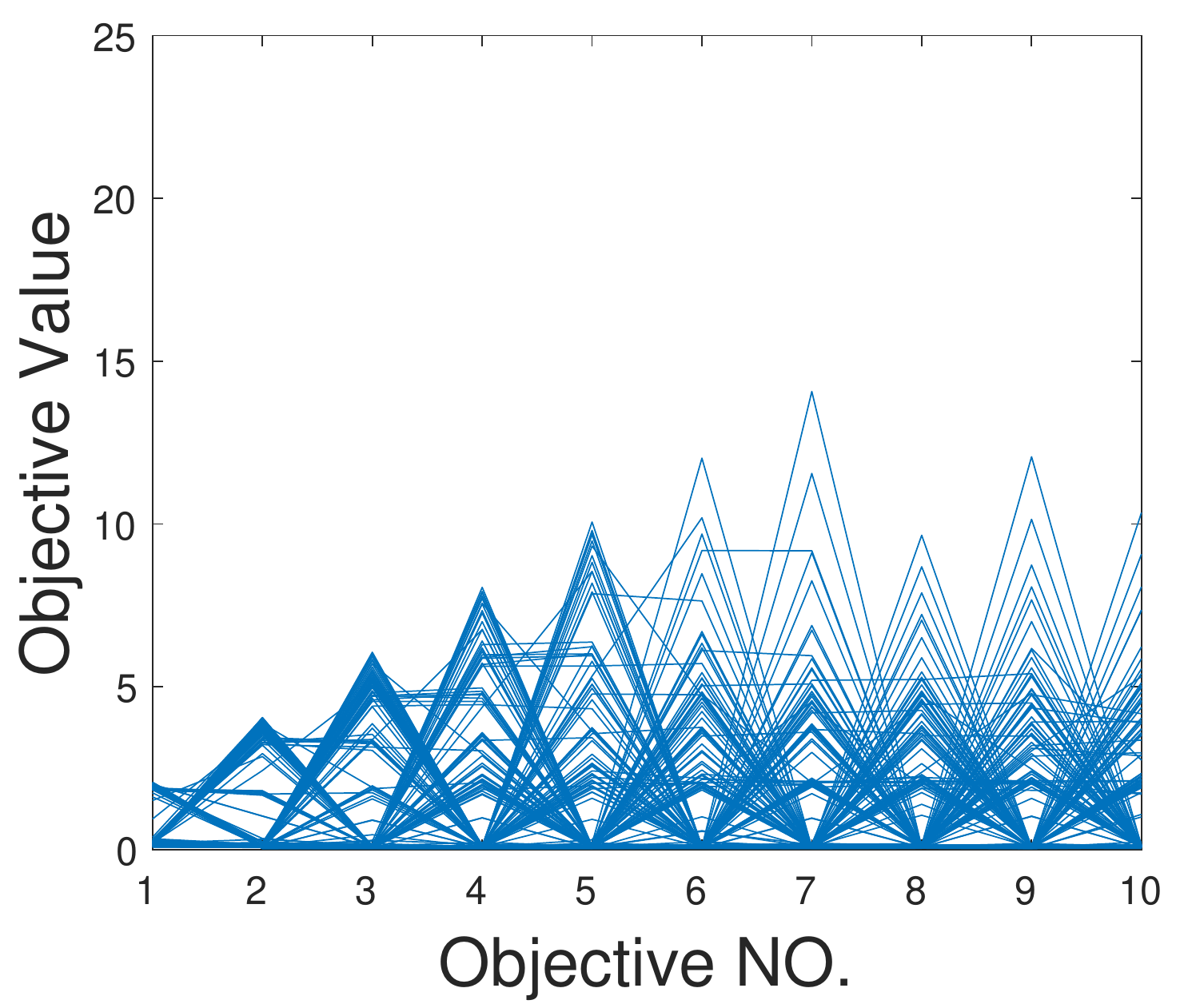}}
      \subfigure[Two\_Arch2]{\includegraphics[width=4cm]{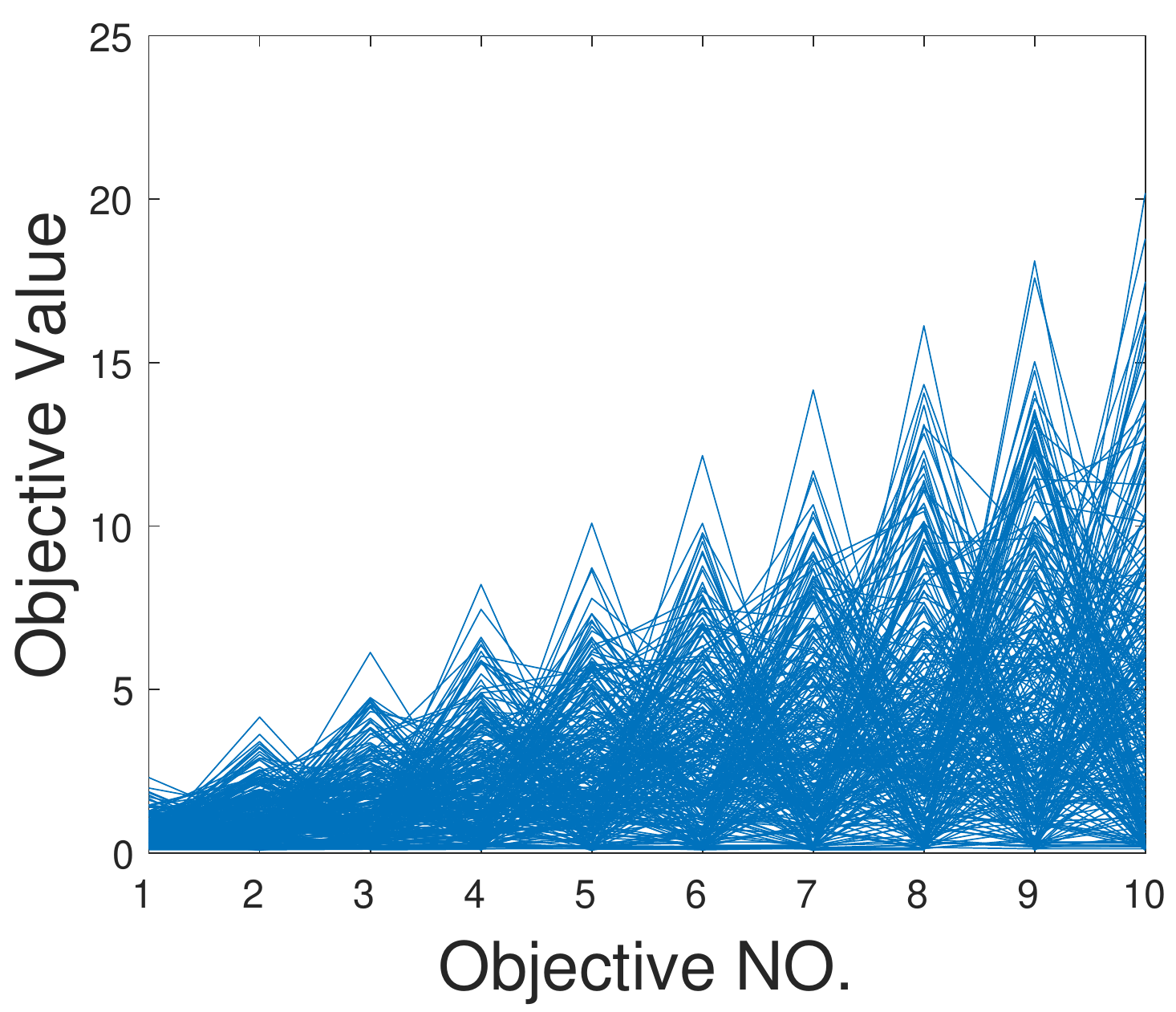}}
      \subfigure[AnD]{\includegraphics[width=4cm]{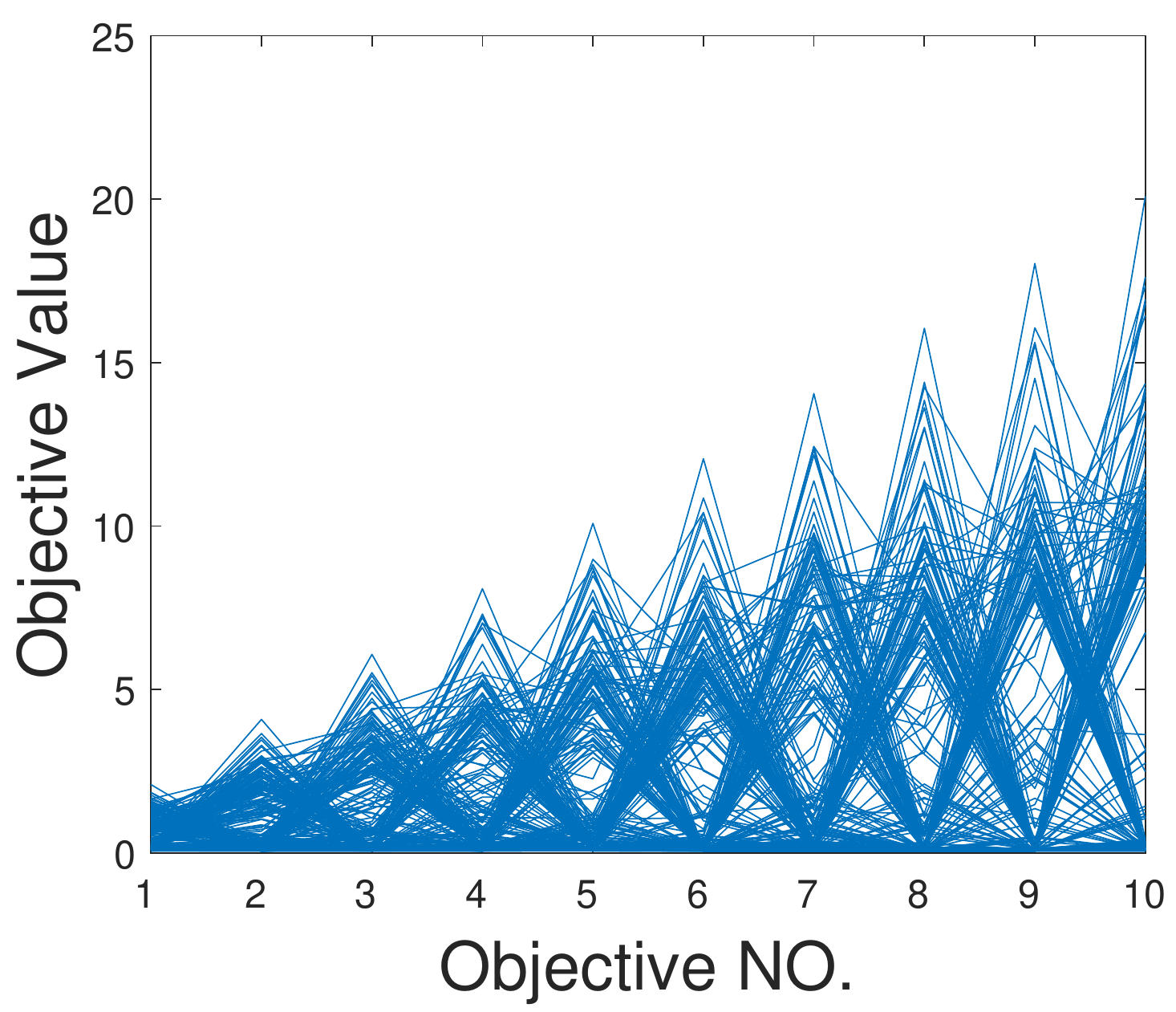}}
       \caption{The final solution sets of the eight compared algorithms on WFG4 with ten objectives by parallel coordinates.}\label{fig:wfgf}
    \end{center}
\end{figure*}

\subsubsection{WFG Test Suite}

The first observation from Tables~\ref{table:IGDstateWFG} and~\ref{table:HVstateWFG} is that AnS attains the best overall performance in terms of both IGD and HV. Next, we give the detailed discussions.

\begin{figure*} [!t]
    \begin{center}
      \subfigure[RVEA]{\includegraphics[width=4cm]{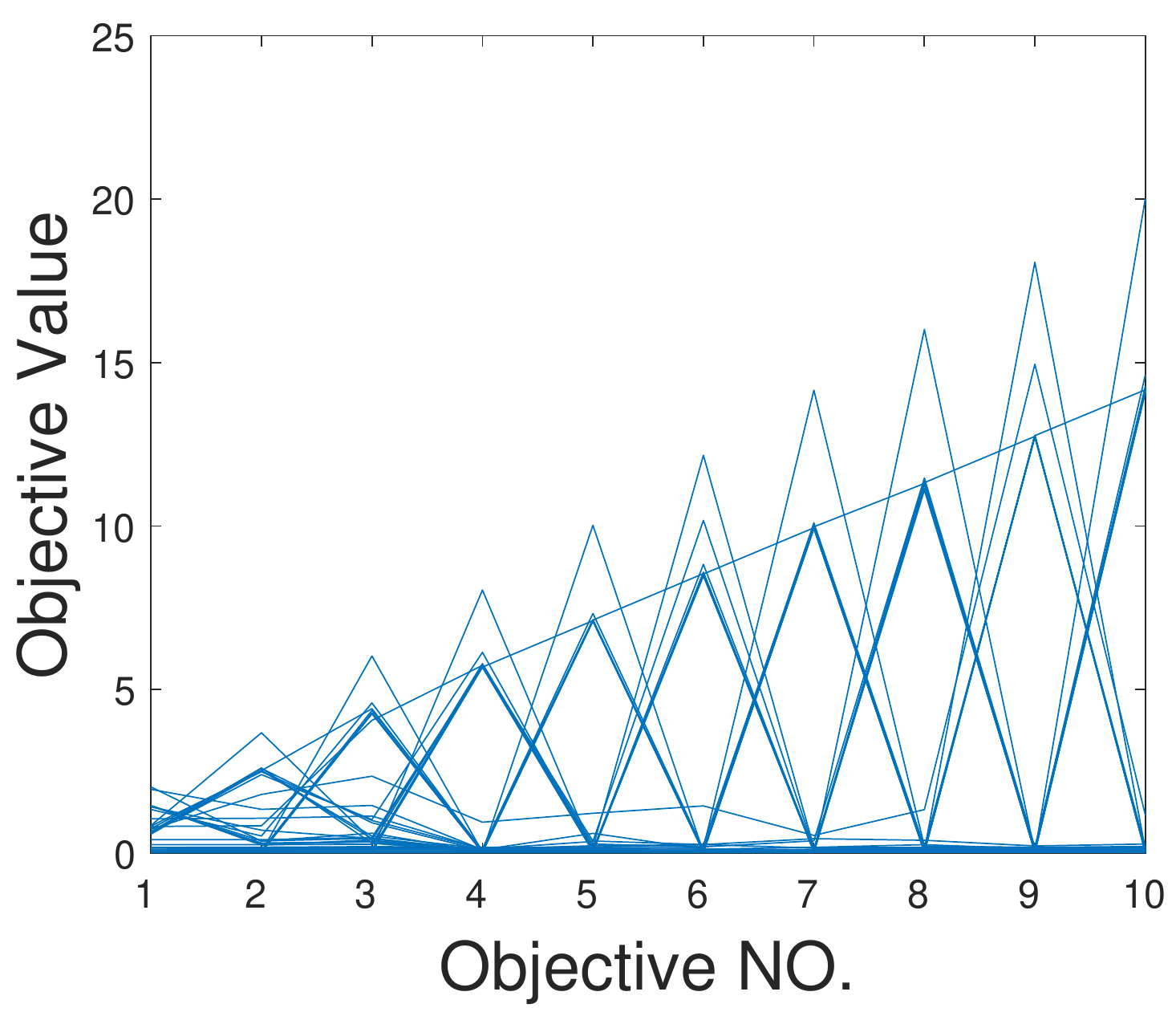}}
      \subfigure[SPEA2+SDE]{\includegraphics[width=4cm]{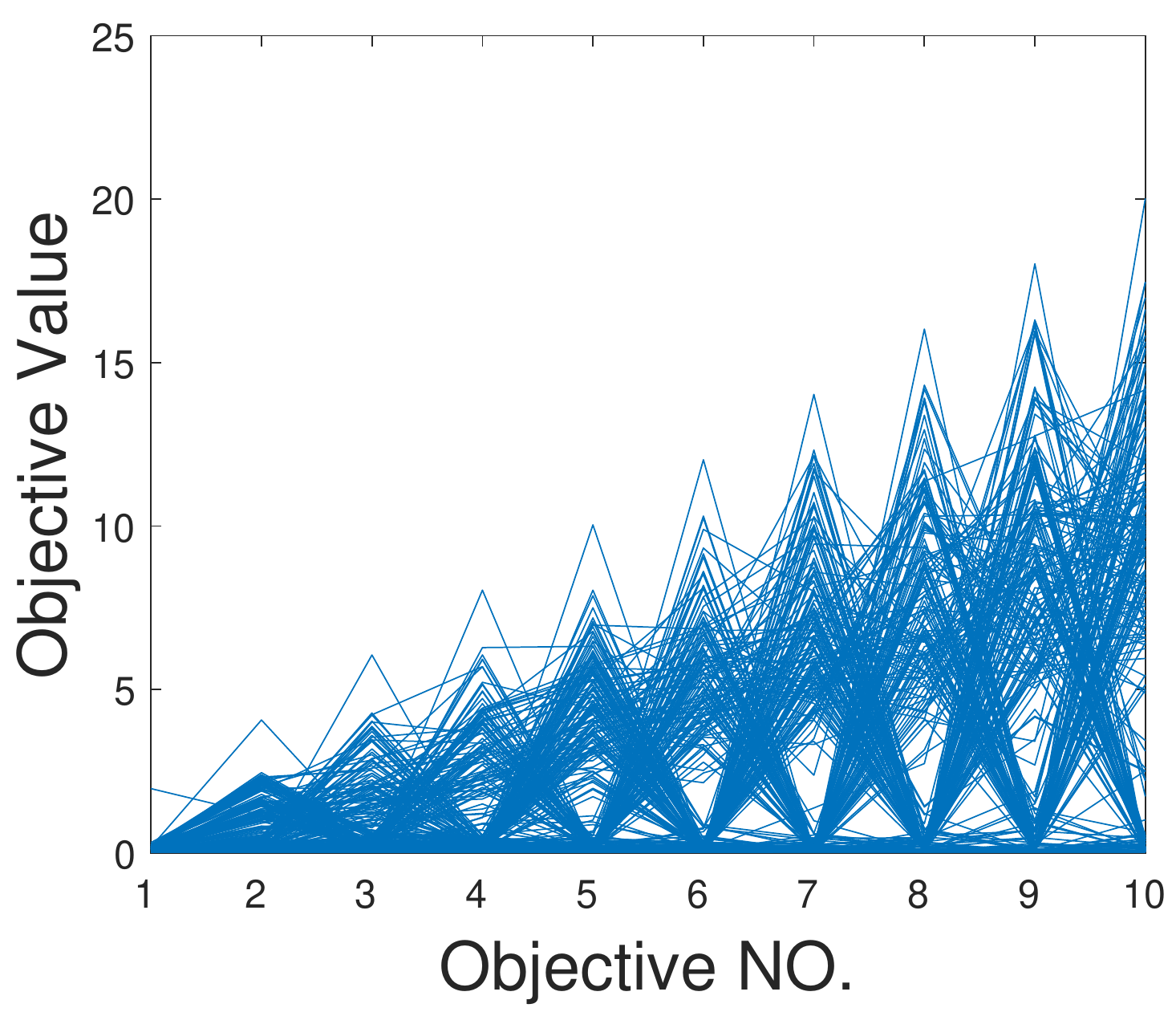}}
      \subfigure[MOEA/D]{\includegraphics[width=4cm]{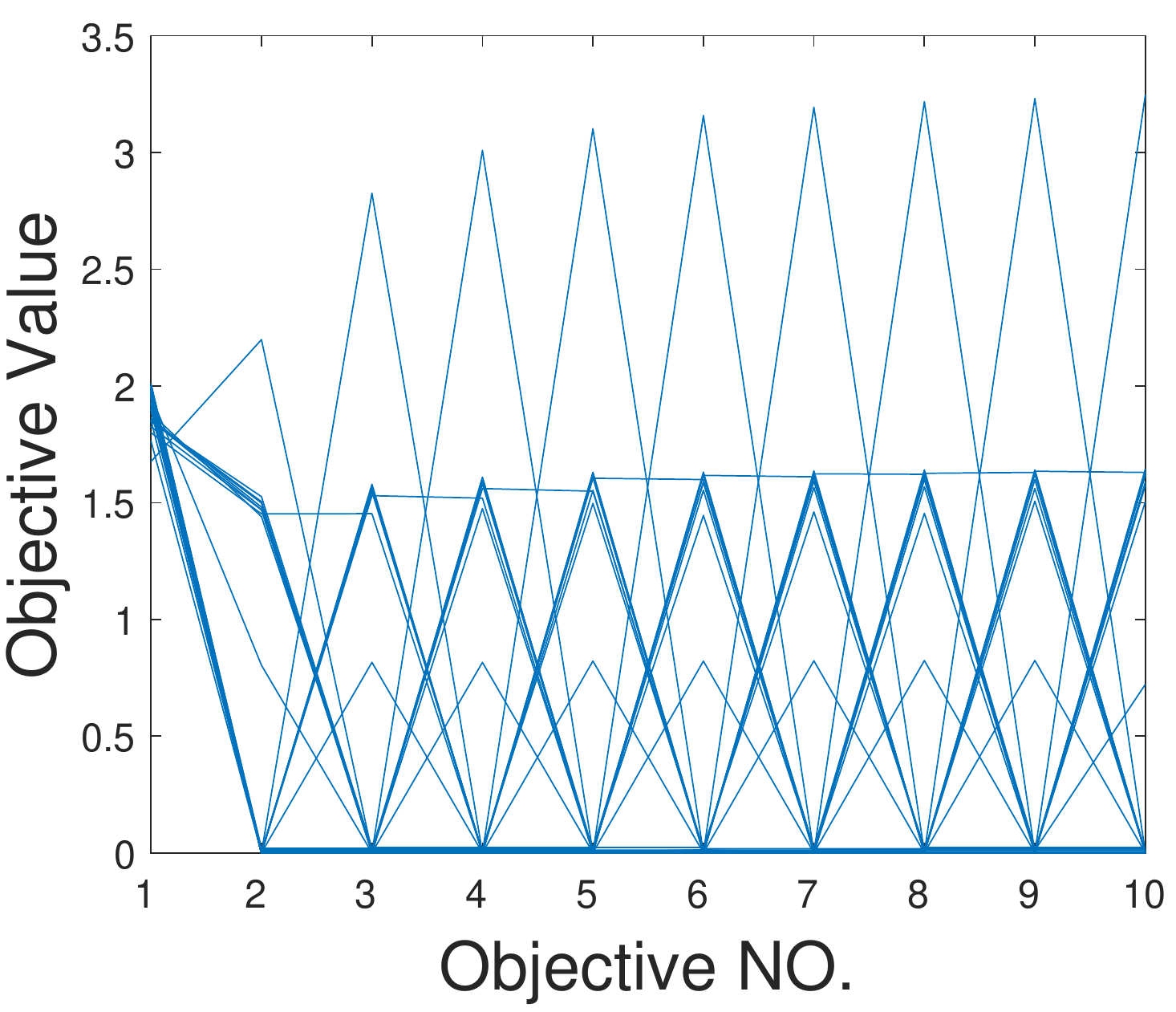}}
      \subfigure[NSGA-III]{\includegraphics[width=4cm]{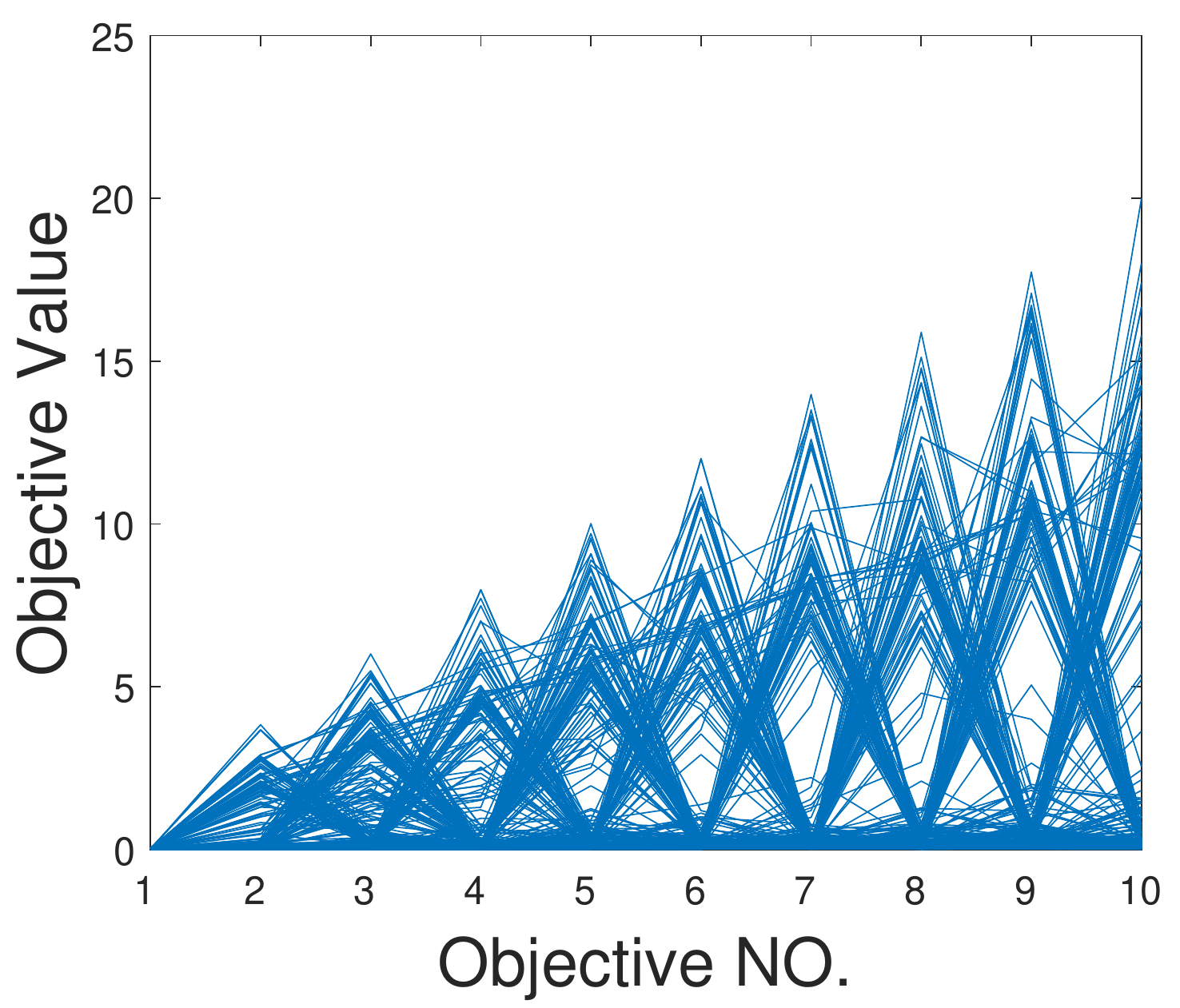}}
      \subfigure[MOMBI-II]{\includegraphics[width=4cm]{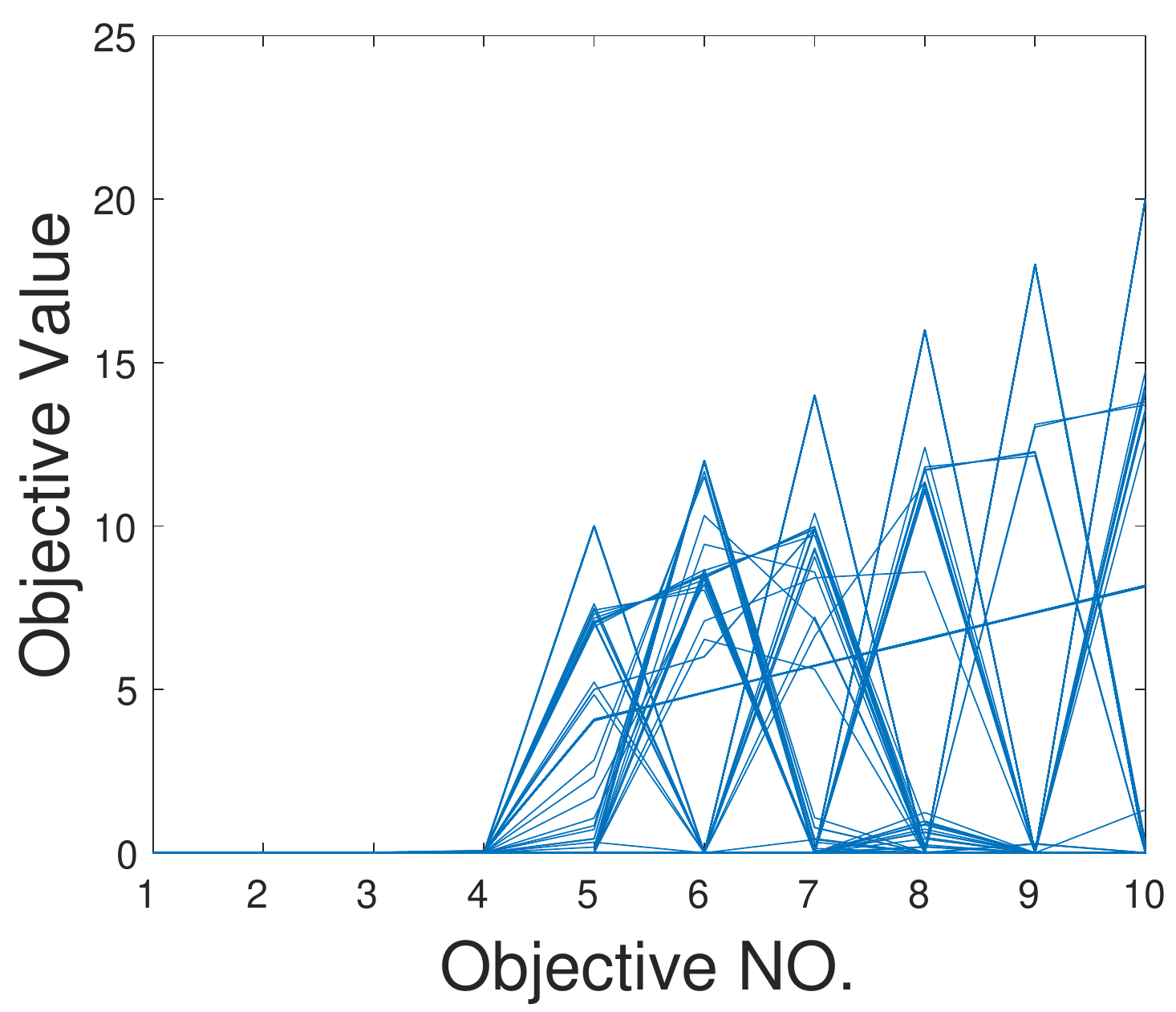}}
      \subfigure[MOEA/DD]{\includegraphics[width=4cm]{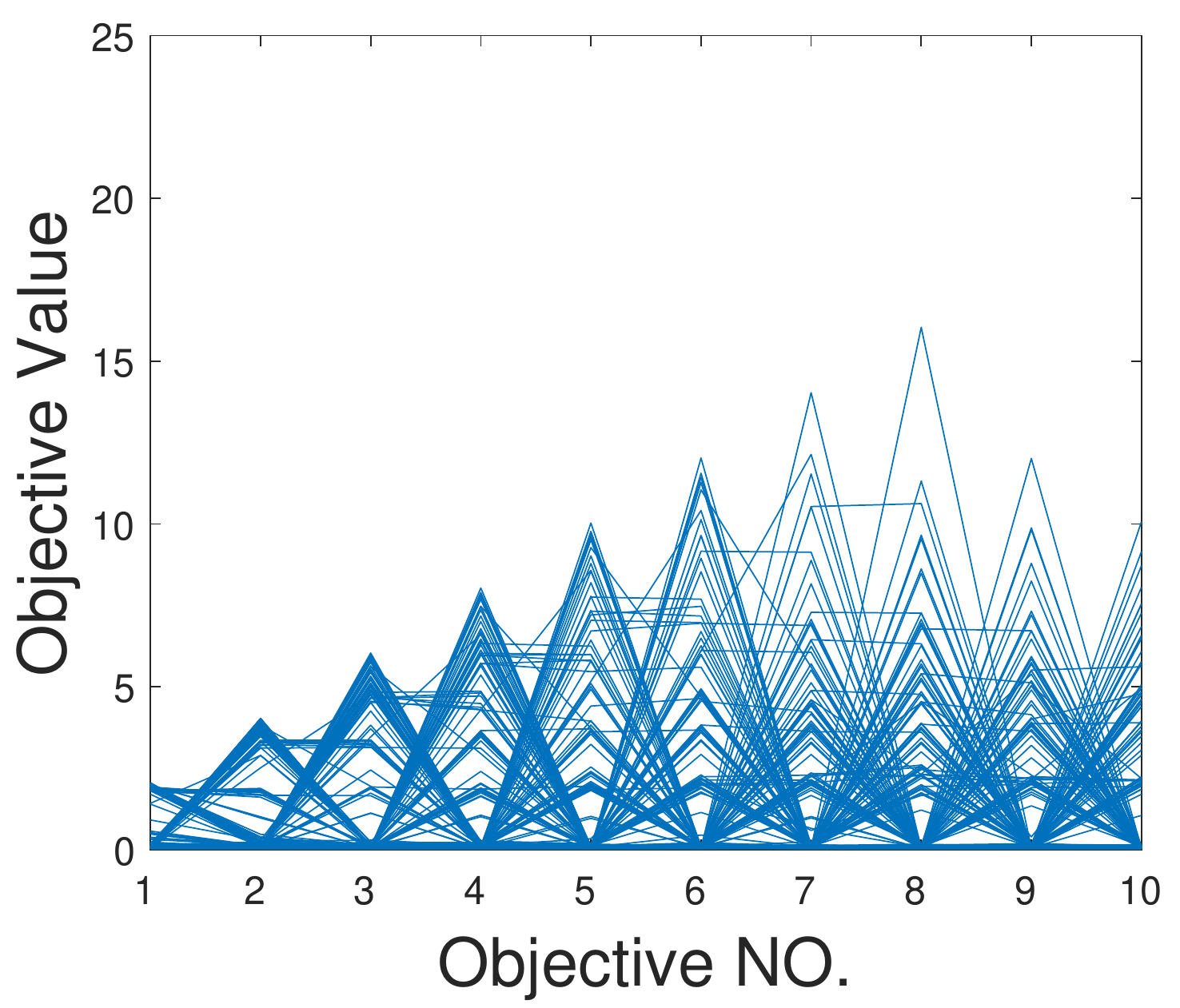}}
      \subfigure[Two\_Arch2]{\includegraphics[width=4cm]{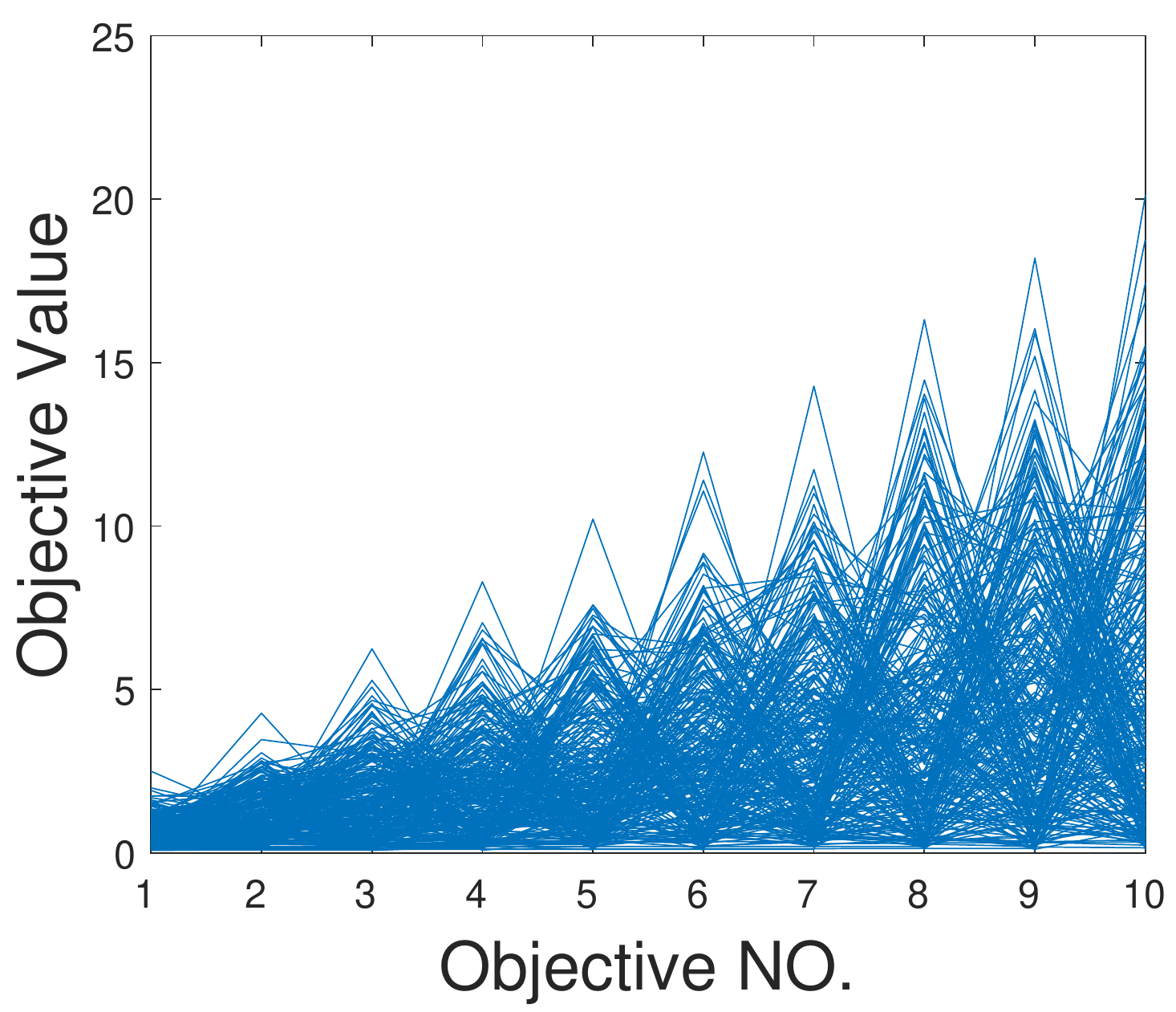}}
      \subfigure[AnD]{\includegraphics[width=4cm]{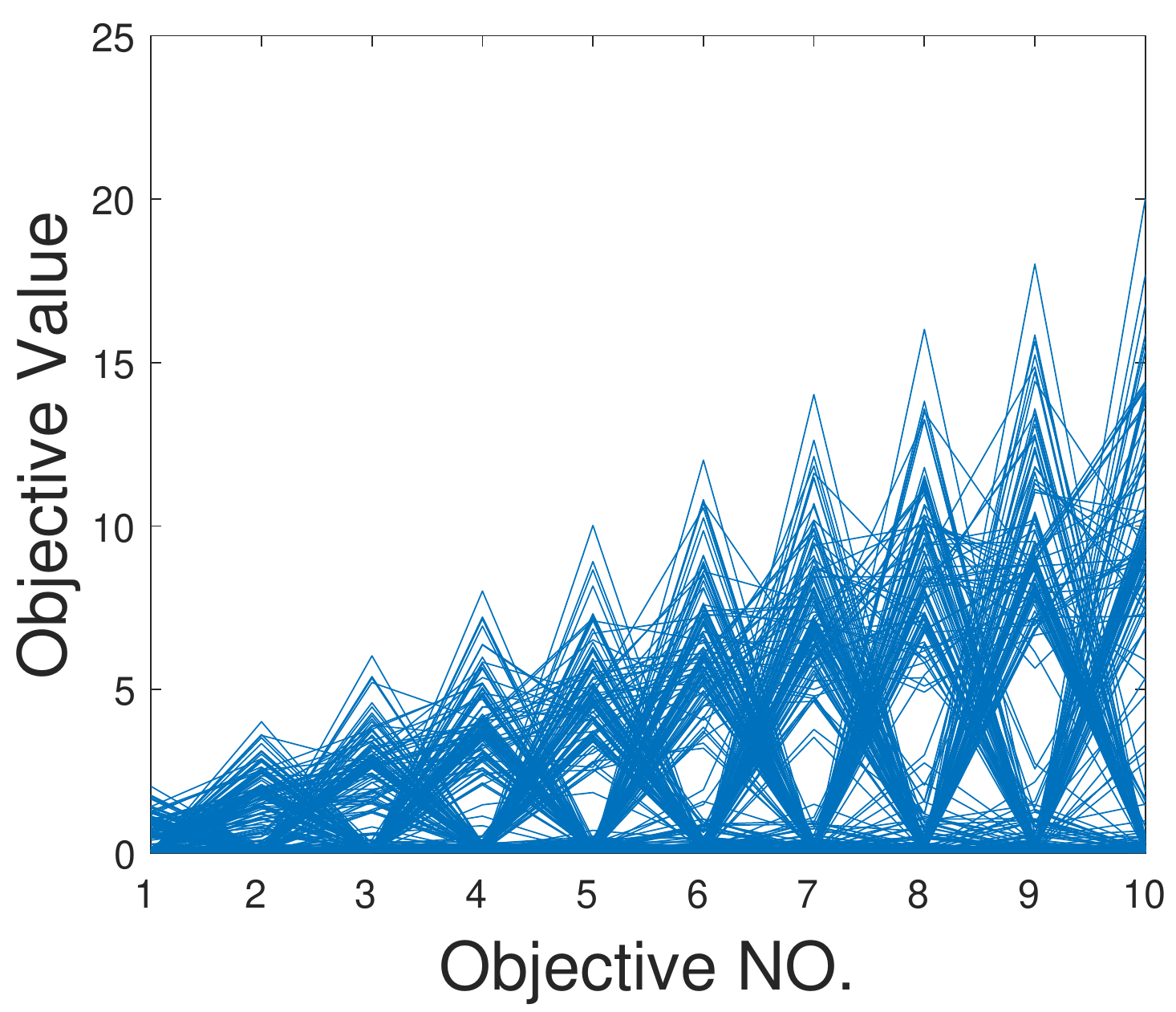}}
       \caption{The final solution sets of the eight compared algorithms on WFG7 with ten objectives by parallel coordinates.}\label{fig:wfgs}
    \end{center}
\end{figure*}

\begin{figure*} [!t]
    \begin{center}
      \subfigure[IGD]{\label{fig:bijiaoIGD}\includegraphics[width=6cm]{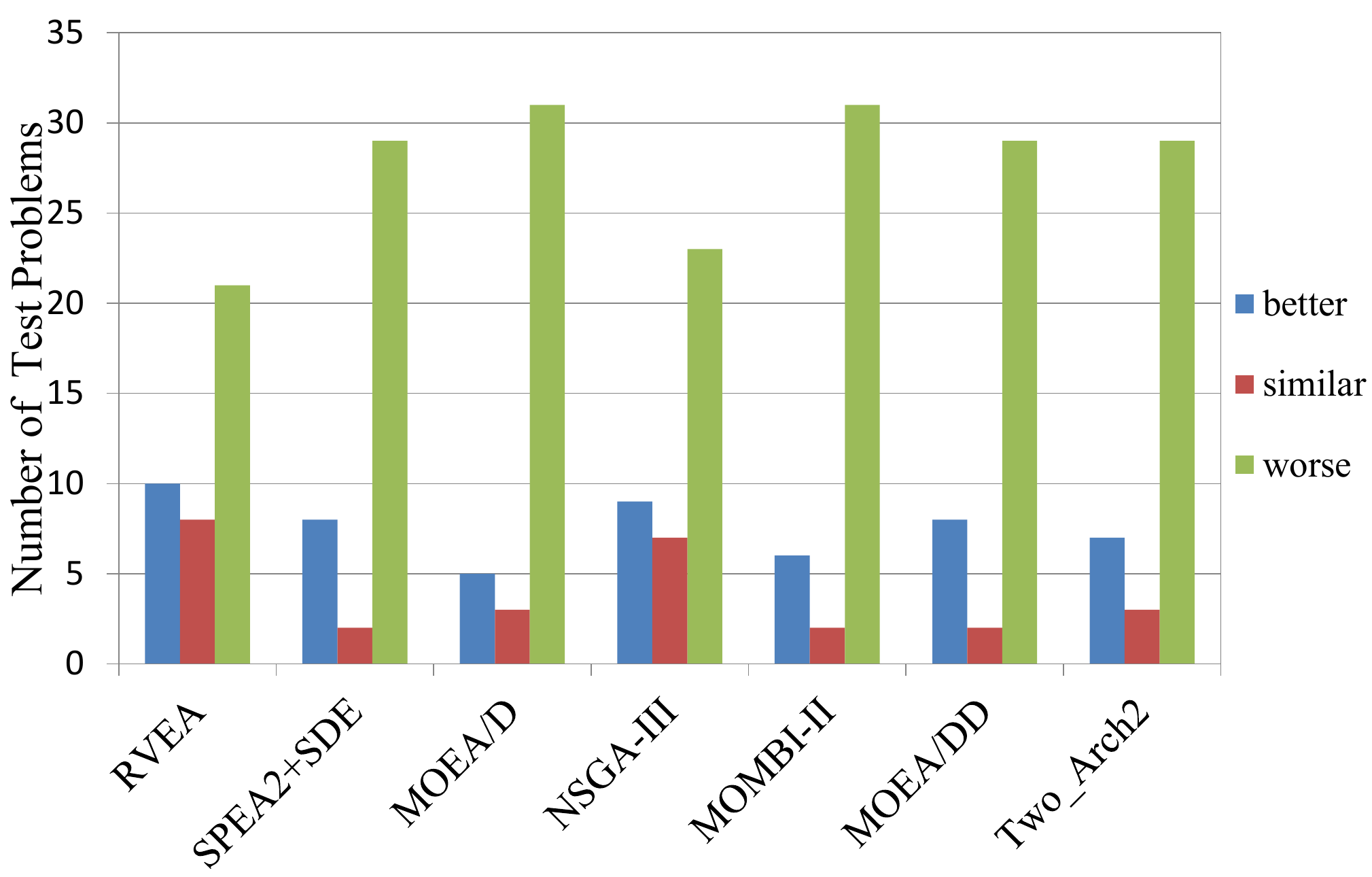}}
        \subfigure[HV]{\label{fig:bijiaoHV}\includegraphics[width=6cm]{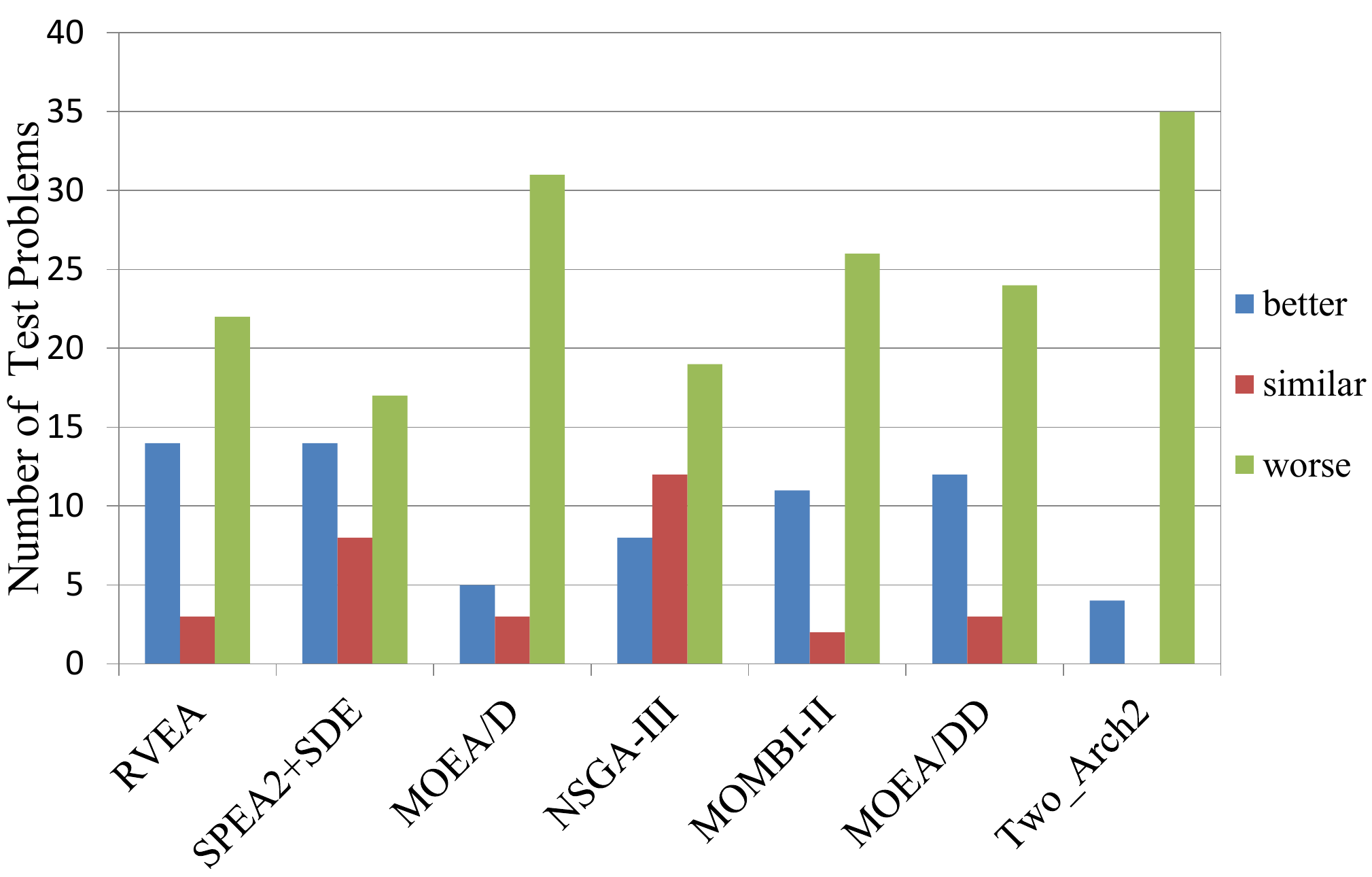}}
       \caption{Wilcoxon rank sum test between AnD and its seven competitors (i.e., RVEA, SPEA2+SDE, MOEA/D, NSGA-III, MOMBI-II, MOEA/DD, and Two\_Arch2) on all the test problems (including DTLZ  and WFG test suites) in terms of IGD and HV. ``better``, ``similar`` and ``worse`` mean that a competitor performs better than, similar to, and worse than AnD, respectively. }\label{fig:bijiao}
    \end{center}
\end{figure*}

\begin{figure*} [!t]
    \begin{center}
      \subfigure[IGD]{\includegraphics[width=5.3cm]{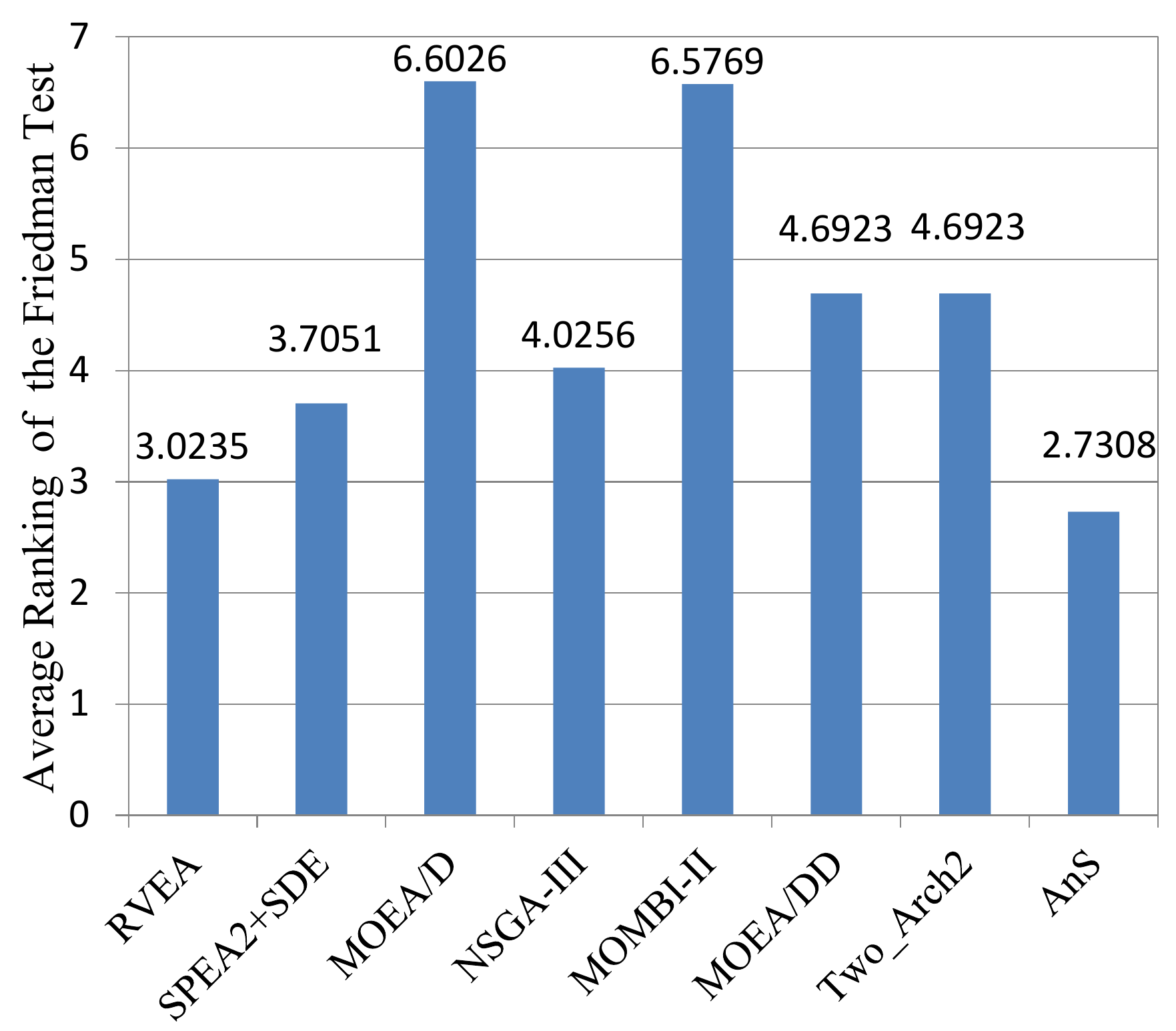}}
        \subfigure[HV]{\includegraphics[width=5.3cm]{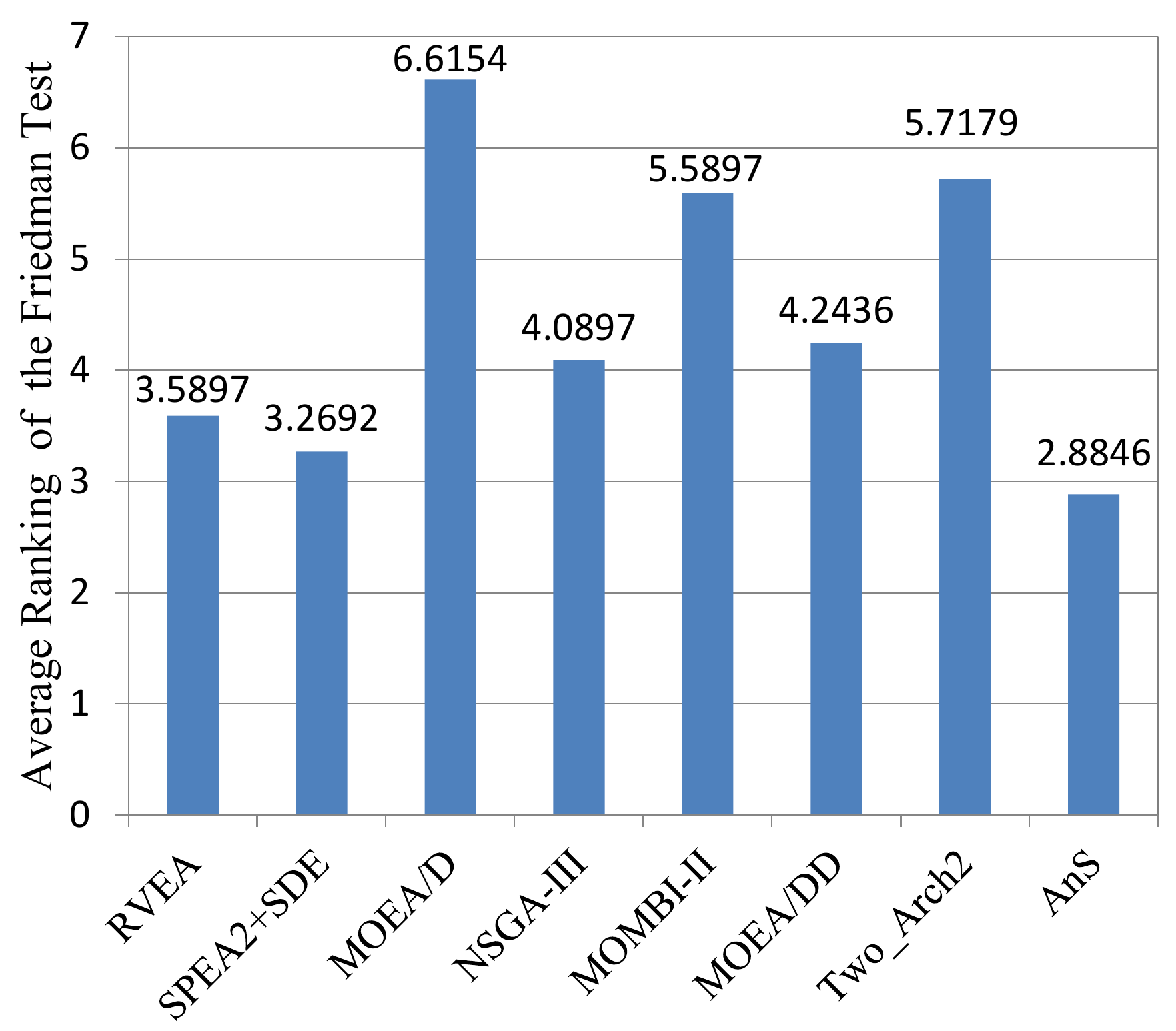}}
       \caption{Friedman test on AnD and its seven competitors (i.e., RVEA, SPEA2+SDE, MOEA/D, NSGA-III, MOMBI-II, MOEA/DD, and Two\_Arch2) on all the test problems (including DTLZ and WFG test suites) in terms of IGD and HV. The smaller the ranking, the better the performance of an algorithm. }\label{fig:paixu}
    \end{center}
\end{figure*}

Table~\ref{table:IGDstateWFG} shows the IGD values resulting from the eight compared algorithms. Clearly, AnS and RVEA are the two top algorithms and they have a clear advantage over the other six algorithms on the majority of test problems. Actually, AnS provides the best and second best IGD values on 12 and six out of 27 test problems, respectively. As for RVEA, it generates six best results and nine second best results. In addition, SPEA2+SDE obtains three best results and two second best results, Two\_Arch2 produces five best results, NSGA-III shows one best result and seven second best results, and MOEA/DD and MOMBI-II reach one second best result. One interesting phenomenon we have observed is that the methods based on weight vectors or reference points (i.e., MOEA/D, MOEA/DD, NSGA-III, and MOMBI-II) seem to lose their effectiveness on this test suite. This can be attributed to the fact that the PFs of WFG test suite are irregular, discontinued or mixed, and scaled with different ranges in each objective. Therefore, a well-distributed weight vectors/reference points can not guarantee a good distribution of obtained solutions. Note, however, that RVEA, which also uses the reference vectors to guide the search, performs better than MOEA/D, MOEA/DD, NSGA-III, and MOMBI-II. It is perhaps because the usage of angle information helps RVEA to alleviate this issue to some extent.

The HV values are given in Table~\ref{table:HVstateWFG}. From Table~\ref{table:HVstateWFG}, AnS and SPEA2+SDE achieve the best and second best overall performance, respectively. Specifically, AnS produces 14 best results and three second best results out of 27 test problems, and SPEA2+SDE shows three best results and eight second best results. It is also observed that AnS reaches the best performance on WFG2, WFG4, WFG6, and WFG7. For MOMBI-II, Two\_Arch2, RVEA, and SPEA2+SDE, they exhibit the best overall performance on WFG1, WFG3, WFG5, and WFG8, respectively. With regard to WFG9, AnS and SPEA2+SDE are the two best algorithms for solving it.

With the aim of revealing more details of the eight compared algorithms, their experimental results on both WFG4 and WFG7 with ten objectives are presented by parallel coordinates in Figs.~\ref{fig:wfgf} and~\ref{fig:wfgs}, respectively. From Fig.~\ref{fig:wfgf}, one can see that MOEA/D and MOEA/DD have relatively poor distributions. It might be because they are lack of a normalization procedure before the evaluation of an individual. As for RVEA and MOMBI-II, the former fails to cover the seventh objective well, while the latter is unable to cover the first four objectives well. In terms of SPEA2+SDE, NSGA-III, Two\_Arch2, and AnD, all of them can cover the whole PF. The difference between them is that the results derived from SPEA2+SDE, NSGA-III, and Two\_Arch2 concentrate mainly on the boundary or the middle parts of the PF, while in AnD, the obtained results can spread out on the whole PF very well. The similar phenomena can also be observed in Fig.~\ref{fig:wfgs}. AnD still has the best distribution. Note that NSGA-III fails to cover the first objective well.

\subsubsection{Discussion}


To analyze the overall performance on both DTLZ and WFG test suites, the Wilcoxon rank sum test was implemented between AnD and the other seven MaOEAs in terms of both IGD and HV metrics. The statistical test results are presented in Fig.~\ref{fig:bijiao}. Fig.~\ref{fig:bijiaoIGD} gives the comparison results in terms of IGD. From Fig.~\ref{fig:bijiaoIGD}, we can find that AnD outperforms RVEA, SPEA2+SDE, MOEA/D, NSGA-III, MOMBI-II, MOEA/DD, and Two\_Arch2 on 21, 29, 31, 23, 31, 29, and 29 test problems, respectively, while it loses on 10, eight, five, nine, six, eight, and seven test problems, accordingly. The comparison results for HV is shown in Fig.~\ref{fig:bijiaoHV}. As shown in Fig.~\ref{fig:bijiaoHV}, AnD performs better than RAVE, SPEA2+SDE, MOEA/D, NSGA-III, MOMBI-II, MOEA/DD, and Two\_Arch2 on 22, 17, 31, 19, 26, 24, and 35 test problems, respectively, while performs worse on 14, 14, five, eight, 11, 12, and four test problems, accordingly. Thus, we can conclude that AnD is able to obtain the better overall performance compared with the seven competitors in terms of both IGD and HV.

Further, the Friedman test was also implemented on all the test problems in terms of both IGD and HV. In the Friedman test, the smaller the ranking, the better the performance of an algorithm. From Fig.~\ref{fig:paixu}, it is evident that AnD has the smallest ranking in terms of both IGD and HV, followed by RVEA and SPEA2+SDE. For RVEA, it ranks the third best and the second best in terms of IGD and HV, respectively. For SPEA2+SDE, it works the second best and the third best in terms of IGD and HV, respectively. The above results indicate that the algorithm with either the shift-based density estimation (i.e., SPEA2+SDE) or angle information (i.e., RVEA) is suitable for solving MaOPs. Moreover, the algorithm with these two elements (i.e., AnD) achieves the best performance, which verifies the main motivation of this paper.

\begin{table}[!]
\caption{Experimental Results (Mean and Standard Deviation) of the IGD and HV Values on C1-DTLZ1, C2-DTLZ2, and C3-DTLZ4. The Better Result between C-AnD and C-NSGA-III on Each Test Problem is Highlighted in Gray.}
\label{tableIGDHVconstriant}
\centering
\resizebox{9cm}{!}{

\begin{tabular}{ccc c}
\toprule
IGD&$m$    &  C-NSGA-III &  C-AnD\\
\midrule
C1-DTLZ1 &   5 & $5.8093e-2$ ($5.60e-3$) &\cellcolor{gray} $5.4026e-2$ ($5.46e-3$) \\
C1-DTLZ1 &  10 & $1.2436e-1$ ($4.98e-3$) &\cellcolor{gray} $1.1237e-1$ ($3.26e-3$)  \\
C1-DTLZ1 &  15 & $2.0774e-1$ ($1.13e-2$) &\cellcolor{gray} $1.7216e-1$ ($2.01e-3$)  \\
\multicolumn{4}{ c }{\hdashrule[0.5ex]{9cm}{0.5pt}{0.8mm} } \\
C2-DTLZ2 &   5 & $1.8916e-1$ ($7.44e-2$) &\cellcolor{gray} $1.5010e-1$ ($6.73e-3$) \\
C2-DTLZ2 &  10 & $3.7888e-1$ ($1.20e-1$) &\cellcolor{gray} $2.6826e-1$ ($5.60e-2$)  \\
C2-DTLZ2 &  15 & $7.2215e-1$ ($1.56e-1$) &\cellcolor{gray}$2.6633e-1$ ($1.62e-1$)  \\
\multicolumn{4}{ c }{\hdashrule[0.5ex]{9cm}{0.5pt}{0.8mm} } \\
C3-DTLZ4 &   5 &\cellcolor{gray} $2.8139e-1$ ($2.04e-2$) & $2.8605e-1$ ($1.40e-2$) \\
C3-DTLZ4 &  10 & $5.7678e-1$ ($6.60e-2$) &\cellcolor{gray} $5.4039e-1$ ($1.34e-2$)  \\
C3-DTLZ4 &  15 & $1.1089e+0$ ($2.72e-1$) &\cellcolor{gray} $7.3981e-1$ ($2.88e-3$)  \\
\bottomrule
\toprule
HV&$m$       &C-NSGA-III &  C-AnD\\
\midrule
C1-DTLZ1 &   5 & $9.7255e-1$ ($2.73e-3$) & \cellcolor{gray}$9.7770e-1$ ($1.90e-3$) \\
C1-DTLZ1 &  10 & $9.8800e-1$ ($1.05e-2$) & \cellcolor{gray}$9.8803e-1$ ($1.66e-2$)  \\
C1-DTLZ1 &  15 & $9.5938e-1$ ($3.76e-2$) & \cellcolor{gray}$9.9054e-1$ ($1.19e-2$)  \\
\multicolumn{4}{ c }{\hdashrule[0.5ex]{9cm}{0.5pt}{0.8mm} } \\
C2-DTLZ2 &   5 & $7.3902e-1$ ($4.53e-2$) &\cellcolor{gray} $7.4493e-1$ ($3.45e-3$)  \\
C2-DTLZ2 &  10 & $8.3045e-1$ ($7.04e-2$) &\cellcolor{gray} $8.7759e-1$ ($1.46e-2$)  \\
C2-DTLZ2 &  15 & $5.9223e-1$ ($2.35e-1$) & \cellcolor{gray}$9.0674e-1$ ($1.65e-1$)  \\
\multicolumn{4}{ c }{\hdashrule[0.5ex]{9cm}{0.5pt}{0.8mm} } \\
C3-DTLZ4 &   5 & $9.4462e-1$ ($2.68e-2$) &\cellcolor{gray} $9.5475e-1$ ($1.61e-3$)  \\
C3-DTLZ4 &  10 & $9.9877e-1$ ($2.51e-3$) & \cellcolor{gray}$9.9909e-1$ ($1.08e-4$)  \\
C3-DTLZ4 &  15 & $9.5658e-1$ ($3.70e-2$) &\cellcolor{gray} $9.9991e-1$ ($1.84e-5$)  \\
\bottomrule
\end{tabular}}
\end{table}

\subsection{Constrained MaOPs}

One may be interested in whether AnD can be applied to solve constrained MaOPs, which are frequently encountered in the real-world applications. To answer this question, AnD was extended to cope with this kind of optimization problem and the resultant algorithm is called C-AnD.

The constraint-handling technique of C-AnD is inspired by the feasibility rule~\cite{Deb2000Anefficient}, which is a well-known constraint-handling technique for constrained single-objective optimization problems. Firstly, we compute the degree of constraint violation for each individual:
\begin{equation}\label{eqn:constraint}
CV(\textbf{x})=\sum_{j=1}^{J}max\{0,g_{j}(\textbf{x})\}+\sum_{k=1}^{K}|h_{k}(\textbf{x})|
\end{equation}
where $g_{j}\geq 0$ and $h_{j} = 0$ denote the $j$th inequality constraint and the $j$th equality constraint, respectively, and $J$ and $K$ are the number of  inequality constraints and equality constraint, respectively. Subsequently, the number of feasible solutions in the union population $\mathcal{U}_{t}$ is calculated. If the number of feasible solutions is larger than $N$, then \textbf{Algorithm 2} is triggered to select $N$ feasible solutions from all the feasible solutions into the next generation. Otherwise, we sort the individuals in $\mathcal{U}_{t}$ according to their degree of constraint violations, and then pick out $N$ individuals with the smallest degree of constraint violations into the next generation.

Overall, the implementation of C-AnD is simple. The performance of C-AnD was compared with C-NSGA-III~\cite{jain2014constraint}, which is the constrained version of NSGA-III, on three representative constrained MaOPs, namely C1-DTLZ1, C2-DTLZ2, and C3-DTLZ4 with five, 10, and 15 objectives. Both C-AnD and C-NSGA-III were run 20 times independently for each test problem. In each run, the maximum number of FEs was set to 180,000 for C1-DTLZ1, and 90,000 for C2-DTLZ2 and C3-DTLZ4. The experimental results are summarized in Table~\ref{tableIGDHVconstriant}.

From Table~\ref{tableIGDHVconstriant}, it can be seen that C-AnD beats C-NSGA-III on all the test problems except C3-DTLZ4 with five objectives in terms of IGD. Therefore, C-AnD is also a simple and effective algorithm for constrained many-objective optimization. It is worth noting that there are no reference points in C-AnD, thus C-AnD does not experience the degeneration of reference points as in constrained decomposition-based approaches.

\section{Conclusion}\label{conclusion}

In this paper, a novel algorithm for dealing with MaOPs, named AnD, was proposed. AnD not only has a simple structure, but also is free from the usage of the Pareto-dominance relation, weight vectors or reference points, and  indicators. The main characteristic of AnD is making use of two strategies (i.e., the angle-based selection and the shift-based density estimation) to delete the poor individuals one by one in the environmental selection.

The aim of the angle-based selection is to maintain the diversity of the search directions. It identifies a pair of individuals with the minimum vector angle, which means that these two individuals search in the most similar direction. Subsequently, the shift-based density estimation is conducted to differentiate them by considering both the diversity and convergence, and to remove the inferior one. We validated that these two strategies play very important roles and are indispensable in AnD. In addition, we compared AnD with seven state-of-the-art MaOEAs for solving MaOPs with up to 15 objective in DTLZ and WFG test suites. The experimental results indicated that, overall, AnD achieves the best performance in terms of both IGD and HV. Due to the fact that MaOPs in real world  often include constraints, AnD was further extended to solve constrained MaOPs and the experimental results verified its effectiveness.

In the future, we will apply AnD to solve some MaOPs in the fields of engineering such as automotive lightweight design and adaptive walking of humanoid robots. Another promising research direction is to combine AnD with other kinds of MaOEAs such as Pareto-based, decomposition-based and indicator-based approaches.





\begin{thebibliography}{10}

\bibitem{zhou2011multiobjective}
A.~Zhou, B.-Y. Qu, H.~Li, S.-Z. Zhao, P.~N. Suganthan, and Q.~Zhang,
  ``Multiobjective evolutionary algorithms: A survey of the state of the art,''
  \emph{Swarm and Evolutionary Computation}, vol.~1, no.~1, pp. 32--49, 2011.

\bibitem{ishibuchi2011behavior}
H.~Ishibuchi, N.~Akedo, H.~Ohyanagi, and Y.~Nojima, ``Behavior of {EMO}
  algorithms on many-objective optimization problems with correlated
  objectives,'' in \emph{2011 IEEE Congress on Evolutionary Computation (CEC
  2011)}.\hskip 1em plus 0.5em minus 0.4em\relax IEEE, 2011, pp. 1465--1472.

\bibitem{ishibuchi2015behavior}
H.~Ishibuchi, N.~Akedo, and Y.~Nojima, ``Behavior of multiobjective
  evolutionary algorithms on many-objective knapsack problems,'' \emph{IEEE
  Transactions on Evolutionary Computation}, vol.~19, no.~2, pp. 264--283,
  2015.

\bibitem{deb2002fast}
K.~Deb, A.~Pratap, S.~Agarwal, and T.~Meyarivan, ``A fast and elitist
  multiobjective genetic algorithm: {NSGA-II},'' \emph{IEEE Transactions on
  Evolutionary Computation}, vol.~6, no.~2, pp. 182--197, 2002.

\bibitem{zitzler2001spea2}
E.~Zitzler, M.~Laumanns, and L.~Thiele, ``{SPEA2}: Improving the strength
  pareto evolutionary algorithm for multiobjective optimization,''
  \emph{Proceedings of Evolutionary Methods for Design, Optimization and
  Control with Applications to Industrial Problems, EUROGEN'2001}, pp. 95--100,
  2001.

\bibitem{zhang2007moea}
Q.~Zhang and H.~Li, ``{MOEA/D}: A multiobjective evolutionary algorithm based
  on decomposition,'' \emph{IEEE Transactions on Evolutionary Computation},
  vol.~11, no.~6, pp. 712--731, 2007.

\bibitem{zhang2008multiobjective}
Q.~Zhang, A.~Zhou, S.~Zhao, P.~N. Suganthan, W.~Liu, and S.~Tiwari,
  ``Multiobjective optimization test instances for the {CEC} 2009 special
  session and competition,'' \emph{University of Essex, Colchester, UK and
  Nanyang technological University, Singapore, special session on performance
  assessment of multi-objective optimization algorithms, technical report},
  vol. 264, 2008.

\bibitem{bader2011hype}
J.~Bader and E.~Zitzler, ``{HypE}: An algorithm for fast hypervolume-based
  many-objective optimization,'' \emph{Evolutionary Computation}, vol.~19,
  no.~1, pp. 45--76, 2011.

\bibitem{Wagner2013A}
M.~Wagner and F.~Neumann, ``A fast approximation-guided evolutionary
  multi-objective algorithm,'' in \emph{Conference on Genetic and Evolutionary
  Computation}, 2013, pp. 687--694.

\bibitem{li2017multi}
M.~Li, C.~Grosan, S.~Yang, X.~Liu, and X.~Yao, ``Multi-line distance
  minimization: A visualized many-objective test problem suite,'' \emph{IEEE
  Transactions on Evolutionary Computation}, 2017, in press. DOI:
  10.1109/TEVC.2017.2655451.

\bibitem{bhattacharjee2017bridging}
K.~Bhattacharjee, H.~Singh, M.~Ryan, and T.~Ray, ``Bridging the gap:
  Many-objective optimization and informed decision-making,'' \emph{IEEE
  Transactions on Evolutionary Computation}, 2017, in press, DOI:
  10.1109/TEVC.2017.2687320.

\bibitem{laumanns2002combining}
M.~Laumanns, L.~Thiele, K.~Deb, and E.~Zitzler, ``Combining convergence and
  diversity in evolutionary multiobjective optimization,'' \emph{Evolutionary
  Computation}, vol.~10, no.~3, pp. 263--282, 2002.

\bibitem{zou2008new}
X.~Zou, Y.~Chen, M.~Liu, and L.~Kang, ``A new evolutionary algorithm for
  solving many-objective optimization problems,'' \emph{IEEE Transactions on
  Systems, Man, and Cybernetics, Part B (Cybernetics)}, vol.~38, no.~5, pp.
  1402--1412, 2008.

\bibitem{wang2007fuzzy}
G.~Wang and H.~Jiang, ``Fuzzy-dominance and its application in evolutionary
  many objective optimization,'' in \emph{International Conference on
  Computational Intelligence and Security Workshops (CISW 2007)}.\hskip 1em
  plus 0.5em minus 0.4em\relax IEEE, 2007, pp. 195--198.

\bibitem{yang2013grid}
S.~Yang, M.~Li, X.~Liu, and J.~Zheng, ``A grid-based evolutionary algorithm for
  many-objective optimization,'' \emph{IEEE Transactions on Evolutionary
  Computation}, vol.~17, no.~5, pp. 721--736, 2013.

\bibitem{adra2011diversity}
S.~F. Adra and P.~J. Fleming, ``Diversity management in evolutionary
  many-objective optimization,'' \emph{IEEE Transactions on Evolutionary
  Computation}, vol.~15, no.~2, pp. 183--195, 2011.

\bibitem{li2014shift}
M.~Li, S.~Yang, and X.~Liu, ``Shift-based density estimation for pareto-based
  algorithms in many-objective optimization,'' \emph{IEEE Transactions on
  Evolutionary Computation}, vol.~18, no.~3, pp. 348--365, 2014.

\bibitem{zhang2015knee}
X.~Zhang, Y.~Tian, and Y.~Jin, ``A knee point-driven evolutionary algorithm for
  many-objective optimization,'' \emph{IEEE Transactions on Evolutionary
  Computation}, vol.~19, no.~6, pp. 761--776, 2015.

\bibitem{trivedi2017survey}
A.~Trivedi, D.~Srinivasan, K.~Sanyal, and A.~Ghosh, ``A survey of
  multiobjective evolutionary algorithms based on decomposition,'' \emph{IEEE
  Transactions on Evolutionary Computation}, vol.~21, no.~3, pp. 440--462,
  2017.

\bibitem{Liu2017Adaptively}
H.~L. Liu, L.~Chen, Q.~Zhang, and K.~Deb, ``Adaptively allocating search effort
  in challenging many-objective optimization problems,'' \emph{IEEE
  Transactions on Evolutionary Computation}, 2017, in press, DOI:
  10.1109/TEVC.2017.2725902.

\bibitem{Yuan2016Balancing}
Y.~Yuan, H.~Xu, B.~Wang, B.~Zhang, and X.~Yao, ``Balancing convergence and
  diversity in decomposition-based many-objective optimizers,'' \emph{IEEE
  Transactions on Evolutionary Computation}, vol.~20, no.~2, pp. 180--198,
  2016.

\bibitem{Wang2016Decomposition}
R.~Wang, Q.~Zhang, and T.~Zhang, ``Decomposition-based algorithms using pareto
  adaptive scalarizing methods,'' \emph{IEEE Transactions on Evolutionary
  Computation}, vol.~20, no.~6, pp. 821--837, 2016.

\bibitem{liu2014decomposition}
H.-L. Liu, F.~Gu, and Q.~Zhang, ``Decomposition of a multiobjective
  optimization problem into a number of simple multiobjective subproblems,''
  \emph{IEEE Transactions on Evolutionary Computation}, vol.~18, no.~3, pp.
  450--455, 2014.

\bibitem{deb2014evolutionary}
K.~Deb and H.~Jain, ``An evolutionary many-objective optimization algorithm
  using reference-point-based nondominated sorting approach, {Part I}: Solving
  problems with box constraints.'' \emph{IEEE Transactions on Evolutionary
  Computation}, vol.~18, no.~4, pp. 577--601, 2014.

\bibitem{ishibuchi2017performance}
H.~Ishibuchi, Y.~Setoguchi, H.~Masuda, and Y.~Nojima, ``Performance of
  decomposition-based many-objective algorithms strongly depends on pareto
  front shapes,'' \emph{IEEE Transactions on Evolutionary Computation},
  vol.~21, no.~2, pp. 169--190, 2017.

\bibitem{hughes2007msops}
E.~J. Hughes, ``{MSOPS-II}: A general-purpose many-objective optimiser,'' in
  \emph{IEEE Congress on Evolutionary Computation (CEC 2007)}.\hskip 1em plus
  0.5em minus 0.4em\relax IEEE, 2007, pp. 3944--3951.

\bibitem{zitzler1998multiobjective}
E.~Zitzler and L.~Thiele, ``Multiobjective optimization using evolutionary
  algorithms -- {A} comparative case study,'' in \emph{International Conference
  on Parallel Problem Solving from Nature}.\hskip 1em plus 0.5em minus
  0.4em\relax Springer, 1998, pp. 292--301.

\bibitem{zitzler2004indicator}
E.~Zitzler and S.~K{\"u}nzli, ``Indicator-based selection in multiobjective
  search,'' in \emph{International Conference on Parallel Problem Solving from
  Nature}.\hskip 1em plus 0.5em minus 0.4em\relax Springer, 2004, pp. 832--842.

\bibitem{trautmann2013r2}
H.~Trautmann, T.~Wagner, and D.~Brockhoff, ``{R2-EMOA}: Focused multiobjective
  search using {R}2-indicator-based selection,'' in \emph{International
  Conference on Learning and Intelligent Optimization}.\hskip 1em plus 0.5em
  minus 0.4em\relax Springer, 2013, pp. 70--74.

\bibitem{li2016stochastic}
B.~Li, K.~Tang, J.~Li, and X.~Yao, ``Stochastic ranking algorithm for
  many-objective optimization based on multiple indicators,'' \emph{IEEE
  Transactions on Evolutionary Computation}, vol.~20, no.~6, pp. 924--938,
  2016.

\bibitem{wang2013preference}
R.~Wang, R.~C. Purshouse, and P.~J. Fleming, ``Preference-inspired
  coevolutionary algorithms for many-objective optimization,'' \emph{IEEE
  Transactions on Evolutionary Computation}, vol.~17, no.~4, pp. 474--494,
  2013.

\bibitem{wang2015ipicea}
R.~Wang, R.~C. Purshouse, I.~Giagkiozis, and P.~J. Fleming, ``The {iPICEA-g}: a
  new hybrid evolutionary multi-criteria decision making approach using the
  brushing technique,'' \emph{European Journal of Operational Research}, vol.
  243, no.~2, pp. 442--453, 2015.

\bibitem{singh2011pareto}
H.~K. Singh, A.~Isaacs, and T.~Ray, ``A pareto corner search evolutionary
  algorithm and dimensionality reduction in many-objective optimization
  problems,'' \emph{IEEE Transactions on Evolutionary Computation}, vol.~15,
  no.~4, pp. 539--556, 2011.

\bibitem{bandyopadhyay2015algorithm}
S.~Bandyopadhyay and A.~Mukherjee, ``An algorithm for many-objective
  optimization with reduced objective computations: A study in differential
  evolution,'' \emph{IEEE Transactions on Evolutionary Computation}, vol.~19,
  no.~3, pp. 400--413, 2015.

\bibitem{yuan2017objective}
Y.~Yuan, Y.-S. Ong, A.~Gupta, and H.~Xu, ``Objective reduction in
  many-objective optimization: Evolutionary multiobjective approaches and
  comprehensive analysis,'' \emph{IEEE Transactions on Evolutionary
  Computation}, 2017, in press, DOI: 10.1109/TEVC.2017.2672668.

\bibitem{li2015evolutionary}
K.~Li, K.~Deb, Q.~Zhang, and S.~Kwong, ``An evolutionary many-objective
  optimization algorithm based on dominance and decomposition,'' \emph{IEEE
  Transactions on Evolutionary Computation}, vol.~19, no.~5, pp. 694--716,
  2015.

\bibitem{wang2015two_arch2}
H.~Wang, L.~Jiao, and X.~Yao, ``{Two\_Arch2}: An improved two-archive algorithm
  for many-objective optimization,'' \emph{IEEE Transactions on Evolutionary
  Computation}, vol.~19, no.~4, pp. 524--541, 2015.

\bibitem{ishibuchi2008evolutionary}
H.~Ishibuchi, N.~Tsukamoto, and Y.~Nojima, ``Evolutionary many-objective
  optimization: A short review,'' in \emph{2008 IEEE Congress on Evolutionary
  Computation (CEC)}.\hskip 1em plus 0.5em minus 0.4em\relax IEEE, 2008, pp.
  2419--2426.

\bibitem{li2015many}
B.~Li, J.~Li, K.~Tang, and X.~Yao, ``Many-objective evolutionary algorithms: A
  survey,'' \emph{ACM Computing Surveys (CSUR)}, vol.~48, no.~1, p.~13, 2015.

\bibitem{cheng2016reference}
R.~Cheng, Y.~Jin, M.~Olhofer, and B.~Sendhoff, ``A reference vector guided
  evolutionary algorithm for many-objective optimization,'' \emph{IEEE
  Transactions on Evolutionary Computation}, vol.~20, no.~5, pp. 773--791,
  2016.

\bibitem{xiang2017vector}
Y.~Xiang, Y.~Zhou, M.~Li, and Z.~Chen, ``A vector angle-based evolutionary
  algorithm for unconstrained many-objective optimization,'' \emph{IEEE
  Transactions on Evolutionary Computation}, vol.~21, no.~1, pp. 131--152,
  2017.

\bibitem{wang2016cooperative}
J.~Wang, W.~Zhang, and J.~Zhang, ``Cooperative differential evolution with
  multiple populations for multiobjective optimization,'' \emph{IEEE
  Transactions on Cybernetics}, vol.~46, no.~12, pp. 2848--2861, 2016.

\bibitem{coello2007evolutionary}
C.~A. Coello~Coello, G.~B. Lamont, D.~A. Van~Veldhuizen \emph{et~al.},
  \emph{Evolutionary {A}lgorithms for {S}olving {M}ulti-objective
  {P}roblems}.\hskip 1em plus 0.5em minus 0.4em\relax Springer, 2007, vol.~5.

\bibitem{wang2016localized}
R.~Wang, Z.~Zhou, H.~Ishibuchi, T.~Liao, and T.~Zhang, ``Localized weighted sum
  method for many-objective optimization,'' \emph{IEEE Transactions on
  Evolutionary Computation}, 2017, in press, DOI: 10.1109/TEVC.2016.2611642.

\bibitem{cai2017decomposition}
X.~Cai, Z.~Yang, Z.~Fan, and Q.~Zhang, ``Decomposition-based-sorting and
  angle-based-selection for evolutionary multiobjective and many-objective
  optimization,'' \emph{IEEE Transactions on Cybernetics}, vol.~47, no.~9, pp.
  2824--2837, 2017.

\bibitem{Xiang2017An}
Y.~Xiang, J.~Peng, Y.~Zhou, M.~Li, and Z.~Chen, ``An angle based constrained
  many-objective evolutionary algorithm,'' \emph{Applied Intelligence},
  vol.~47, no.~3, pp. 705--720, 2017.

\bibitem{he2017many}
Z.~He and G.~G. Yen, ``Many-objective evolutionary algorithms based on
  coordinated selection strategy,'' \emph{IEEE Transactions on Evolutionary
  Computation}, vol.~21, no.~2, pp. 220--233, 2017.

\bibitem{PESA}
D.~W. Corne and J.~D. Knowles, ``Techniques for highly multiobjective
  optimisation: Some nondominated points are better than others,'' in
  \emph{Proceedings of the 9th Annual Conference on Genetic and Evolutionary
  Computation}, 2007, pp. 773--780.

\bibitem{Jiang2016A}
S.~Jiang and S.~Yang, ``A strength pareto evolutionary algorithm based on
  reference direction for multi-objective and many-objective optimization,''
  \emph{IEEE Transactions on Evolutionary Computation}, vol.~21, no.~3, pp.
  329--346, 2017.

\bibitem{denysiuk2017multiobjective}
R.~Denysiuk and A.~Gaspar-Cunha, ``Multiobjective evolutionary algorithm based
  on vector angle neighborhood,'' \emph{Swarm and Evolutionary Computation},
  2017, in press, DOI: 10.1016/j.swevo.2017.05.005.

\bibitem{deb2005scalable}
K.~Deb, L.~Thiele, M.~Laumanns, and E.~Zitzler, ``Scalable test problems for
  evolutionary multiobjective optimization,'' \emph{Evolutionary Multiobjective
  Optimization. Theoretical Advances and Applications}, pp. 105--145, 2005.

\bibitem{huband2006review}
S.~Huband, P.~Hingston, L.~Barone, and L.~While, ``A review of multiobjective
  test problems and a scalable test problem toolkit,'' \emph{IEEE Transactions
  on Evolutionary Computation}, vol.~10, no.~5, pp. 477--506, 2006.

\bibitem{emmerich2005emo}
M.~Emmerich, N.~Beume, and B.~Naujoks, ``An {EMO} algorithm using the
  hypervolume measure as selection criterion.'' in \emph{EMO}, vol. 3410.\hskip
  1em plus 0.5em minus 0.4em\relax Springer, 2005, pp. 62--76.

\bibitem{hernandez2015improved}
R.~Hern{\'a}ndez~G{\'o}mez and C.~A. Coello~Coello, ``Improved metaheuristic
  based on the {R2} indicator for many-objective optimization,'' in
  \emph{Proceedings of the 2015 Annual Conference on Genetic and Evolutionary
  Computation}.\hskip 1em plus 0.5em minus 0.4em\relax ACM, 2015, pp. 679--686.

\bibitem{TianAn}
Y.~Tian, R.~Cheng, X.~Zhang, F.~Cheng, and Y.~Jin, ``An indicator based
  multi-objective evolutionary algorithm with reference point adaptation for
  better versatility,'' \emph{IEEE Transactions on Evolutionary Computation},
  2017, in press, DOI: 10.1109/TEVC.2017.2749619.

\bibitem{Tian2017PlatEMO}
Y.~Tian, R.~Cheng, X.~Zhang, and Y.~Jin, ``{PlatEMO}: A matlab platform for
  evolutionary multi-objective optimization,'' \emph{IEEE Computational
  Intelligence Magzine}, 2017, in press.

\bibitem{ishibuchi2014review}
H.~Ishibuchi, H.~Masuda, Y.~Tanigaki, and Y.~Nojima, ``Review of coevolutionary
  developments of evolutionary multi-objective and many-objective algorithms
  and test problems,'' in \emph{2014 IEEE Symposium on Computational
  Intelligence in Multi-Criteria Decision-Making}.\hskip 1em plus 0.5em minus
  0.4em\relax IEEE, 2014, pp. 178--184.

\bibitem{Deb2000Anefficient}
K.~Deb, ``An efficient constraint handling method for genetic algorithms,''
  \emph{Computer Methods in Applied Mechanics and Engineering}, vol. 186,
  no.~2, pp. 311--338, 2000.

\bibitem{jain2014constraint}
H.~Jain and K.~Deb, ``An evolutionary many-objective optimization algorithm
  using reference-point based nondominated sorting approach, {Part II}:
  Handling constraints and extending to an adaptive approach,'' \emph{IEEE
  Transactions on Evolutionary Computation}, vol.~18, no.~4, pp. 602--622,
  2014.

\end{thebibliography}
\end{document}